\documentclass[journal]{IEEEtran}

\usepackage[utf8]{inputenc}  
\usepackage{amssymb}         
\usepackage{amsmath}         
\usepackage{graphicx}        
\usepackage{subcaption}      
\usepackage{cite}            
\usepackage{booktabs}        
\usepackage{multirow}        
\usepackage{multicol}        
\usepackage{algorithm}       
\usepackage{algpseudocode}   
\usepackage{xcolor}          
\usepackage{enumerate}
\usepackage{tabularx}        
\usepackage{booktabs} 
\usepackage{xcolor}  
\usepackage{array}  
\usepackage{booktabs} 
\usepackage{boldline} 
\usepackage{adjustbox}

\ifCLASSINFOpdf
\else
\fi
\begin{document}
%
\title{Learning Where to Focus: Density-Driven Guidance for Detecting Dense Tiny Objects}
%
%
%

\author{Zhicheng Zhao, Xuanang Fan, Lingma Sun, Chenglong Li and Jin Tang
        
\thanks{This work was supported in part by the National Natural Science Foundation of China(No. 62306005, 62006002, and 62076003), and in part by the Natural Science Foundation of Anhui Higher Education Institution (No. 2022AH040014). (Corresponding author:Lingma Sun).}

\thanks{Zhicheng Zhao and Chenglong Li are with Key Laboratory of Intelligent Computing \& Signal Processing (Anhui University), Ministry of Education, Anhui Provincial Key Laboratory of Multimodal Cognitive Computation, School of Artificial Intelligence, Anhui University, Hefei 230601, China. Zhicheng Zhao is also with the 38th Research Institute, China Electronics Technology Group Corporation, Hefei 230088, China. (Email: zhaozhicheng@ahu.edu.cn, lcl1314@foxmail.com).}
\thanks{Lingma Sun is with Collaborative Innovation Laboratory for Computer Vision and Pattern Recognition, School of Artificial Intelligence and Big Data, Hefei University, Hefei, 230601, China. (Email: sunlm@hfuu.edu.cn).}
\thanks{Xuanang Fan and Jin Tang are with Anhui Provincial Key Laboratory of Multimodal Cognitive Computation, School of Computer Science and Technology, Anhui University, Hefei 230601, China. (Email: tangjin@ahu.edu.cn, e23301203@stu.ahu.edu.cn).}
}

\markboth{Journal of \LaTeX\ Class Files,~Vol.~13, No.~9, September~2014}%
{Shell \MakeLowercase{\textit{et al.}}: Bare Demo of IEEEtran.cls for Journals}
%



\maketitle

\begin{abstract}
High-resolution remote sensing imagery increasingly contains dense clusters of tiny objects, the detection of which is extremely challenging due to severe mutual occlusion and limited pixel footprints. Existing detection methods typically allocate computational resources uniformly, failing to adaptively focus on these density-concentrated regions, which hinders feature learning effectiveness. To address these limitations, we propose the Dense Region Mining Network (DRMNet), which leverages density maps as explicit spatial priors to guide adaptive feature learning. First, we design a Density Generation Branch (DGB) to model object distribution patterns, providing quantifiable priors that guide the network toward dense regions. Second, to address the computational bottleneck of global attention, our Dense Area Focusing Module (DAFM) uses these density maps to identify and focus on dense areas, enabling efficient local-global feature interaction. Finally, to mitigate feature degradation during hierarchical extraction, we introduce a Dual Filter Fusion Module (DFFM). It disentangles multi-scale features into high- and low-frequency components using a discrete cosine transform and then performs density-guided cross-attention to enhance complementarity while suppressing background interference. Extensive experiments on the AI-TOD and DTOD datasets demonstrate that DRMNet surpasses state-of-the-art methods, particularly in complex scenarios with high object density and severe occlusion.

\end{abstract}


%
\IEEEpeerreviewmaketitle
\section{INTRODUCTION}
\IEEEPARstart{I}{n} recent years, remote sensing object detection has emerged as an important research direction in computer vision. Its primary goal is to automatically locate and classify valuable objects, such as aircraft, ships, and vehicles, from remote sensing images. As a fundamental task, remote sensing object detection demonstrates broad application prospects in various fields including military reconnaissance, agricultural management, and urban planning \cite{beijing-1, beijing-2, beijing-3}. With the development of remote sensing imaging technology, high-resolution images have been widely adopted, providing rich object information. However, high-resolution images contain numerous tiny objects that are often densely packed with mutual occlusion. Therefore, detecting tiny objects remains a major challenge in remote sensing object detection.

\begin{figure}
    \centering
    \includegraphics[width=1\linewidth]{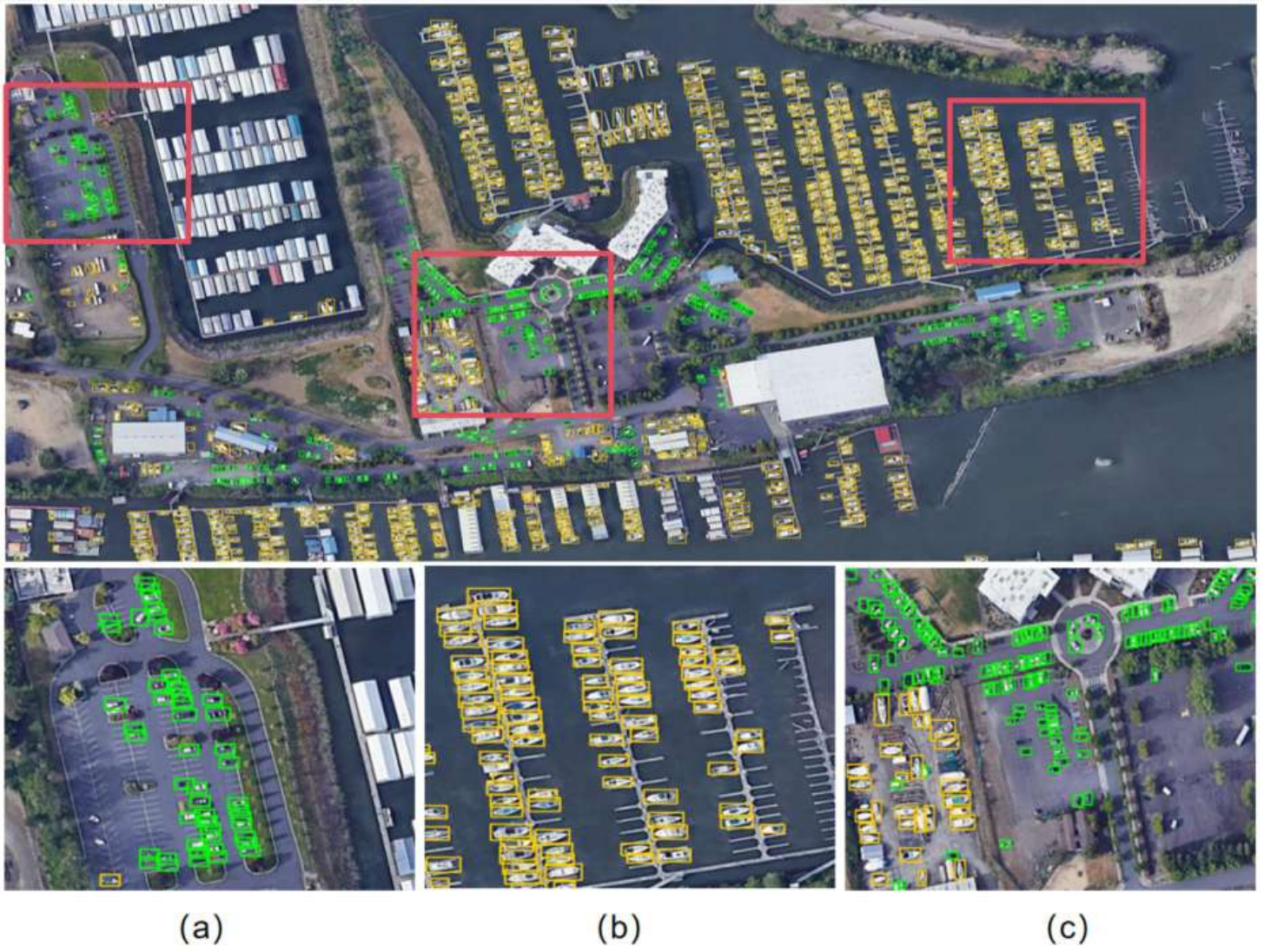}
    \caption{Visualization of real-world scenarios with densely arranged tiny objects in remote sensing imagery. (a) Closely arranged cars in a parking lot; (b) Ships navigating along the seashore; (c) Densely packed objects of different categories in a scene. Tiny objects in remote sensing imagery are often densely packed, appearing as scattered clusters.
}
    \label{fig:enter-label}
\end{figure}

With breakthroughs in both technical paradigms and performance metrics, significant progress has been achieved in general object detection. UniDetector proposed by Li et al. \cite{unidetector} employs image-text alignment and heterogeneous label space training, combined with decoupled training methods and category-agnostic localization networks, to achieve the capability of detecting numerous categories in the open world, greatly improving model generalization. Zhang et al. proposed CP-DETR \cite{cpdetr}, which introduces a hybrid encoder incorporating concept cues and visual cues and enhances the interaction of image and text information through multi-scale fusion, performing well in zero-shot detection tasks. YOLOv10 designed by Wang et al. \cite{yolov10_1} employs a consistent dual allocation strategy to eliminate NMS dependency and improves detection performance while ensuring real-time capability through efficiency-accuracy driven component optimization. Although these methods have achieved significant progress, they still cannot accurately detect tiny objects. As shown in Fig. \ref{fig:enter-label}, tiny objects in remote sensing images are usually densely distributed and randomly distributed. General object detection methods cannot be biased toward objects of interest and inevitably capture only weak and insufficient features for tiny objects.

 Researchers have proposed various methods \cite{ffca, 2025frequency, cabdet} to address the challenges of tiny objects by enhancing the feature representation or interaction of regions containing tiny objects. For instance, Zhang et al. \cite{ffca} proposed a spatial context awareness module to enhance local perception of tiny objects and strengthen global spatial correlation. Zhao et al. \cite{2025frequency} designed a gradient perception frequency attention module to directly highlight the high-frequency information of tiny objects by utilizing the characteristics of infrared images. He et al. \cite{cabdet} designed a feature extraction and fusion optimization module to address the problems of insufficient feature extraction in the backbone network, misalignment of features in the neck network, and information loss of tiny objects. Although these methods have achieved notable results in feature representation and object region determination, two key limitations remain. First, the prior information of object locations is not fully explored and consistently integrated into the entire feature learning process, resulting in inaccurate computation of interest regions. Second, the fusion of frequency domain and spatial domain features lacks explicit object guidance, making it difficult to extract the detailed information and semantic associations of tiny objects. Consequently, feature interaction remains insufficient and the application scenarios are relatively limited.

\begin{figure}
    \centering
    \includegraphics[width=1\linewidth]{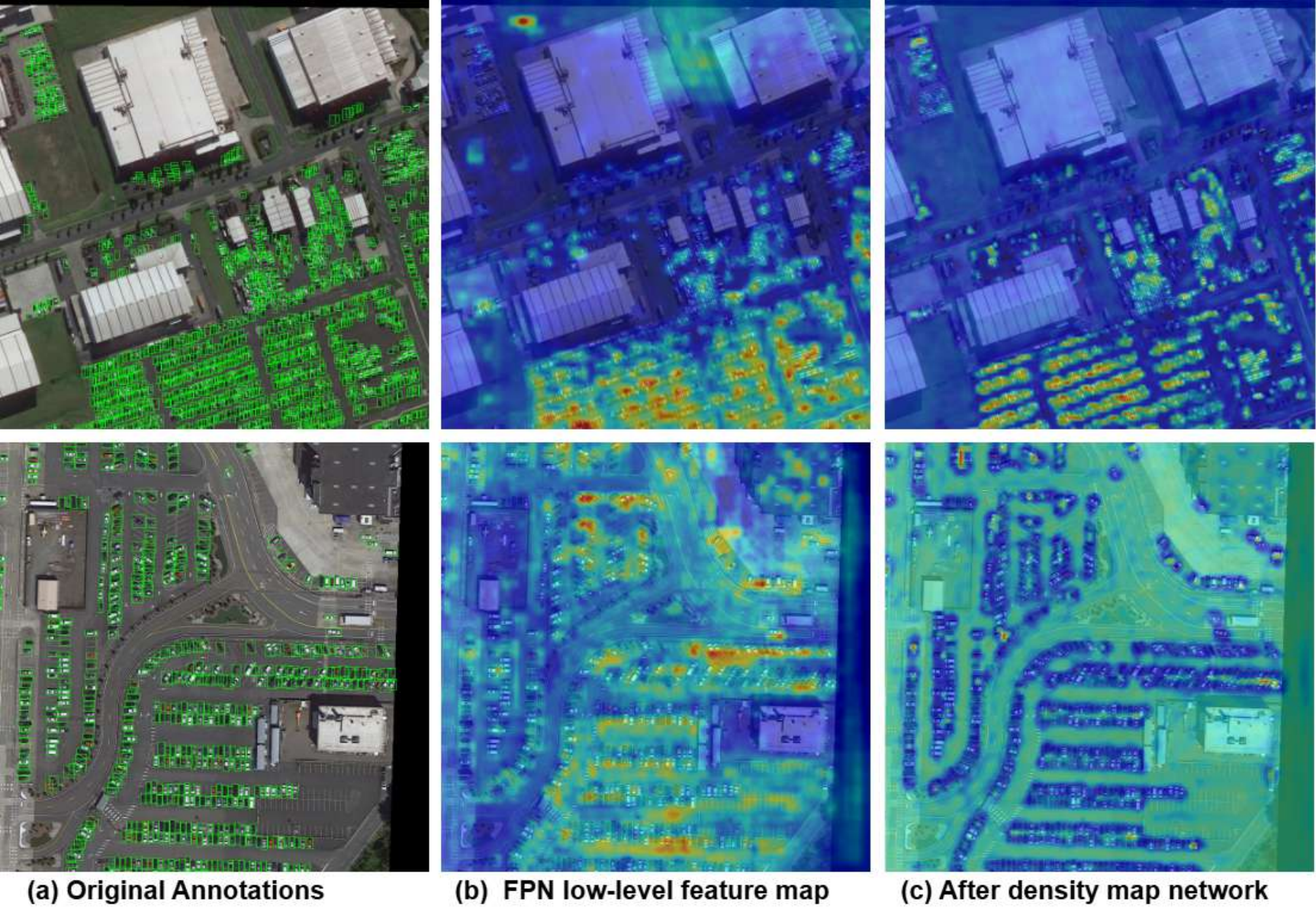}
    \caption{Visualization of feature perception of the foreground: The density maps generated by our Density Generation Branch (DGB) exhibit stronger foreground perception capabilities compared to the model's underlying features and can more clearly distinguish object boundary information.
}
    \label{fig:enter-label_2}
\end{figure}

The fundamental challenge in detecting dense tiny objects lies in the lack of explicit spatial prior information that can effectively match resource allocation with object distribution. Since tiny objects cluster in dispersed dense areas but remain sparse overall, uniform computation wastes resources in background-dominated regions and fails to enhance the features of dense areas. As illustrated in Fig. \ref{fig:enter-label_2}, the density maps produced by our density generation branch provide markedly superior foreground perception compared with conventional low-level feature maps. They clearly distinguish the boundaries of dense objects, demonstrating that density maps can serve as reliable guidance. Moreover, as network depth increases, background interference becomes progressively more pronounced, and tiny objects are highly sensitive to boundaries. Therefore, assigning more attention to foreground regions based on reliable density priors is critical for improving detection performance on dense tiny objects.


In this paper, we propose a novel Dense Region Mining Network (DRMNet), a density-driven network that adaptively focuses on dense tiny objects to enhance feature discrimination capabilities in complex remote sensing scenarios. DRMNet is motivated by leveraging the density prior of tiny objects to guide feature learning. To this end, we first introduce a Density Generation Branch (DGB) to effectively capture the spatial distribution of tiny objects. Typically, tiny objects exhibit local clustering yet global sparsity. To exploit this property, we develop a Dense Area Focusing Module (DAFM), which uses the density map generated by DGB to identify regions where tiny objects concentrate. DAFM extracts features from dense regions to guide subsequent attention operations. Furthermore, to mitigate the problem of high-frequency information being masked by background clutter or low-frequency semantic information during network propagation, we propose a Dual Filter Fusion Module (DFFM). This module applies Discrete Cosine Transform (DCT) to decompose multi-level features into high-frequency and low-frequency components. Then, a density map-guided cross-attention mechanism is designed to adaptively fuse these components, thereby enhancing foreground perception and suppressing background noise. The main contributions of this paper are summarized as follows.

\begin{itemize}
    \item We propose a novel collaborative framework called DRMNet, consisting of a main detection network and an auxiliary density generation branch. The generated density map guides the main network to concentrate computational resources on densely packed object regions.
    \item We introduce a DAFM, which utilizes the density map as spatial prior information to intelligently identify and highlight key regions. These regions are extracted as surrogate representations to facilitate efficient local-to-global information interaction, avoiding the computational burden of global attention mechanisms.
    \item We design DFFM to address the severe degradation of tiny object features in deep networks. DFFM exploits the density map to fuse the multi-scale decomposed high-frequency and low-frequency features based on DCT. The density-guided cross-enhancement mechanism effectively suppresses background noise.
    \item Extensive experiments on two challenging tiny object detection benchmark datasets, AI-TOD and DTOD, validate the effectiveness of the proposed DRMNet. The results demonstrate that the proposed method achieves excellent performance, especially in scenarios with densely packed objects.

\end{itemize}

\section{RELATED WORK}

\subsection{General Object Detection}
With the development of Transformer architectures, object detection technology based on CNN frameworks has undergone multiple generations of innovation, evolving from traditional methods to Transformer-based architectures. Early detectors \cite{maskrcnn, focaloss, efficientdet} were initially effective but had limitations in complex scenes and diverse object appearances. Deep learning based detectors, leveraging CNNs to learn robust features from large-scale datasets, have significantly improved detection accuracy. R-CNN, proposed by Redmon et al. \cite{r-cnn}, laid the foundation for deep learning in object detection by introducing a two-stage detection paradigm that first obtains candidate regions and then screens boxes. YOLO, a one-stage detector introduced by R et al. \cite{yolo}, transformed detection into a regression problem, greatly boosting model efficiency. Recently, the introduction of Transformer architectures has revolutionized detection paradigms. Ni et al. \cite{detr} first applied Transformers to end-to-end detection with DETR, replacing NMS with bipartite matching and simplifying the detection process. Multimodal and open-world detection have become recent research hotspots, with models like CLIP \cite{clip} and SAM \cite{sam} enhancing localization capabilities by integrating vast amounts of information. 

Although significant progress has been achieved, these methods still face challenges in detecting dense tiny objects, resulting in limited generalization ability and poor performance in these scenarios.


\subsection{Tiny Object Detection}
In recent years, advanced feature fusion techniques have been widely explored to enhance multi-scale object detection in remote sensing images. Liu et al. \cite{ffca} simultaneously enhanced the network's local perception capability, multi-scale feature fusion efficiency, and global correlation across channels and spaces by itegrating a feature enhancement module, a feature fusion module, and a spatial context awareness module. Based on these strategies, Wang et al. \cite{b-33} extracted multi-scale local features using a CNN backbone and achieved global feature fusion with the selective scanning mechanism of Mamba. Li et al. \cite{B-4} proposed CFENet to address the issue of small object features being easily overwhelmed, which incorporates a feature suppression module to filter background and redundant information, and combined it with an improved Gauss-Wasserstein distance loss (IGWD) and an enhanced detection head. In addition, Zhang et al. \cite{B-5} improved fine-grained feature extraction by integrating a lightweight backbone network called RepViT-TD with a lightweight detection head equipped with deformable convolution to balance performance and computational cost.

Despite recent progress, most existing methods still lack dedicated mechanisms for accurately locating and characterizing tiny objects, which typically occupy only a few pixels in an image, leading to extremely sparse feature representations and degraded detection performance.

\subsection{Density Map Assistance} 
Density maps play a crucial role in computer vision and have been widely applied in object detection\cite{densedetection}, crowd counting\cite{densecrowd} , and object segmentation\cite{densesegement}. By highlighting both the spatial locations of objects and their corresponding density weights, density maps significantly facilitate object localization. In crowd detection, for example, density maps are integrated into the query vectors of the Transformer decoder\cite{redensecq} to achieve dynamic matching between the query and the object distributions. Duan et al.\cite{redense2021coarse} segment the image into object-dense sub-regions, avoiding background redundancy and object truncation in uniform cropping, but such density cues are not involved in the feature learning process. DMNet \cite{dmnet} adopts a pixel-intensity filtering strategy to crop the density map as a form of data augmentation at the input stage. YOLC  \cite{yolc} utilizes CenterNet heatmaps as an indirect representation of density, which only enlarges and crops the identified clustered regions to avoid calculations on the entire layer.

Although density maps have proven highly effective, most existing approaches employ them merely as auxiliary cues and do not incorporate them into deeper feature fusion or attention interaction stages. 

\subsection{Frequency-Domain Feature Learning}
Frequency domain analysis has become a key approach to addressing the feature attenuation of tiny objects in remote sensing imagery, as discriminative details of tiny objects are mainly concentrated in high-frequency components, whereas background noise is often manifested in low-frequency components. Recent studies have explored various frequency domain enhancement strategies to improve tiny object representation. Zhang et al. \cite{relfrequency} used Discrete Fourier Transform (DFT) to transform the features extracted by the pre-trained visual transformer to the frequency domain, and then dynamically adjusted the intensity of high-frequency details and low-frequency semantics according to the importance of the frequency components. In the field of multimodal remote sensing object detection, a frequency-filtering expert module was designed to decompose RGB and infrared multimodal features into high-frequency/low-frequency components \cite{Low-Rank}, suppress amplitude noise, and fuse high- and low-frequency components.  Dynamic snake convolution (DSConv) \cite{yolouav} was developed to enhance the ability to capture high-frequency details of tiny objects in UAV aerial images and strengthen the interaction between frequency and spatial domain features.

However, most methods employ fixed frequency adjustment strategies that cannot adapt to the dynamic distribution of objects. Furthermore, frequency decomposition is typically limited to a single scale, failing to fully capture multi-scale frequency information.

\section{METHODOLOGY}
The overall framework and designed modules are presented in this section. We first provide a comprehensive overview of the proposed DRMNet, and then elaborate the DGB, DAFM and DFFM in subsequent subsections.


\begin{figure*}
\centering 
\includegraphics[width=1\textwidth]{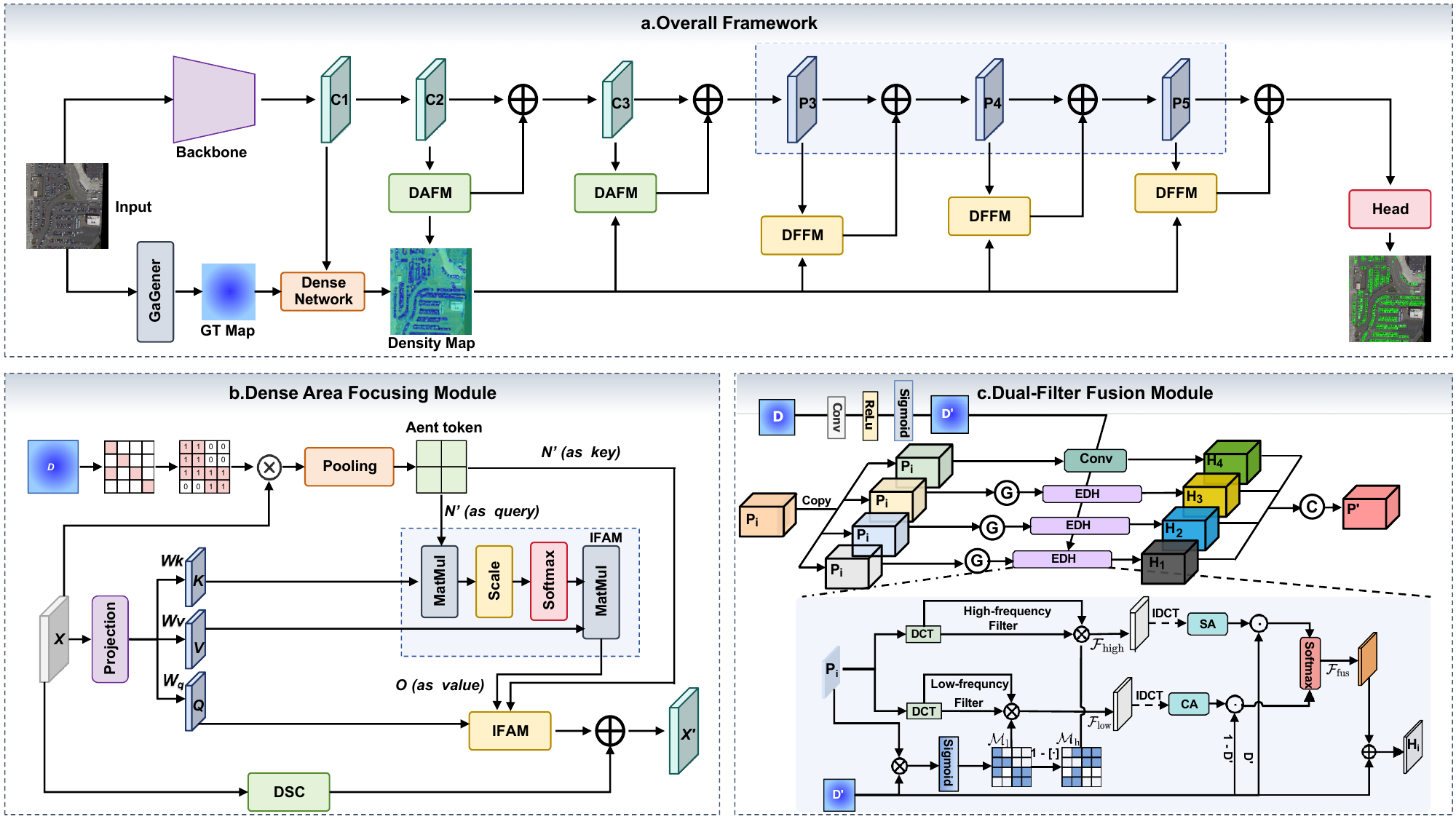} 
\caption{The architecture of proposed DRMNet. In our network, the Dense Area Focusing Module (DAFM) acquires layers C2 and C3 of the feature pyramid. Based on the density map information corresponding to the object, it obtains the corresponding object feature aggregation regions. This module has relatively low computational cost, yet it achieves sufficient information interaction between focused features and global features. The Dual Filter Fusion Module (DFFM) uses a density-guided cross-enhancement mechanism to fuse these components, effectively suppressing background noise while integrating key details and contextual information.}
\label{Fig.3}
\end{figure*}

\subsection{Overall Framework}

The overall architecture of DRMNet is depicted in Fig. 3. The proposed framework is composed of three innovative components: a Density Generation Branch (DGB) for explicit spatial prior modeling, a Dense Area Focusing Module (DAFM) for adaptive region-aware feature interaction, and a Dual Filter Fusion Module (DFFM) for frequency-aware feature enhancement.

Specifically, the backbone features are first fed into DGB, where a branch network generates a density map at the corresponding feature levels. This branch is optimized during training to continuously refine the quality of the predicted density maps. In the feature interaction stage, the DAFM employs a region segmentation strategy to partition the feature space into regions of interest based on the density distribution. These regions are then re-weighted through an attention mechanism to emphasize dense-object areas. In the detection stage, the DFFM module further exploits the density map as guidance to enhance feature fusion. Multi-scale features are decomposed into high- and low-frequency components via DCT, followed by a cross-frequency interaction mechanism guided by the density map.

\subsection{Density Generation Branch}
Tiny objects occupy only a limited proportion of remote sensing images, making their feature representations easily overlooked. Moreover, tiny objects are usually densely arranged in scattered areas of the image. Such a distribution pattern not only increases the difficulty of object detection, but also severely disrupts the efficient allocation of computational resources. As a result, conventional detection models struggle to effectively detect these scattered yet densely distributed tiny objects under constrained computational budgets.

To address this challenge, we employ the DGB to explicitly assign density weights to each pixel within the feature map. The overall structure of the DGB consists of three parts: an encoder, a decoder, and a regressor. The encoder utilizes a pre-trained ResNet-18 backbone to extract new features from the input features. The decoder progressively upsamples the encoded features through three stages of BasicBlock modules\cite{resnet18}. Finally, a regression head consisting of a $3 × 3$ convolution followed by a ReLU activation maps the decoded features to the object resolution, producing the corresponding density map.


To supervise and evaluate the quality of density map generation during training time, we follow the commonly adopted density map construction strategy in crowd counting tasks\cite{clip-ebc}. Specifically, a Gaussian kernel is applied to each annotated object location to produce the ground truth density map, as defined in Equations \ref{MakeGT} and \ref{Gusskernael}.

\begin{equation}
D_i(x, y) = \frac{1}{2\pi\gamma_i^2} \exp\left(-\frac{(x - \mu_{x})^2 + (y - \mu_{y})^2}{2\gamma_i^2}\right)
\label{MakeGT}
\end{equation}

\begin{equation}
\gamma_i = \frac{1}{2} \sqrt{H_i^2 + W_i^2}
\label{Gusskernael}
\end{equation}

For object  \(i\), the bounding box center \((\mu_{x}, \mu_{y})\) is applied as the excitation point for density map generation. To accommodate substantial variations in object size and shape, the standard deviation \(\gamma\) of the Gaussian kernel is adaptively determined based on the height \(H_i\) and width \(W_i\) of the object. This adaptive ground-truth construction based on object-specific scale enables scale-consistent density representation, effectively overcoming the limitations of fixed-parameter methods when dealing with cross-scale objects and improving the fidelity of the generated density map.
Furthermore, a dual-supervision strategy is introduced by integrating density map supervision (depicted in Equation \ref{Ldense}) with the original detection loss.

\begin{equation}
L_{\text{dense}} = \frac{1}{M \times N} \sum_{x=1}^{M} \sum_{y=1}^{N} \left( D_{\text{gt}}(x,y) - D_{\text{pred}}(x,y) \right)^2
\label{Ldense}
\end{equation}

\begin{equation}
L= L_{\text{reg}} + L_{\text{cls}} + L_{\text{dense}}
\label{Lsum}
\end{equation}
where \(M\) and \(N\) represent the ranges of the horizontal and vertical dimensions of the image, respectively. \(D_{\text{pred}}(\cdot)\) is the predicted density value at the specified coordinates. The pixel-wise discrepancy between the generated density map and the ground-truth density map is measured using the mean squared error. Subsequently, the auxiliary supervision loss \(L_{\text{dense}}\) is jointly optimized with the classification loss \(L_{\text{cls}}\) and regression loss \(L_{\text{reg}}\) of the main detection framework. This joint optimization strategy not only preserves the detection accuracy of the primary task but also strengthens feature representation through auxiliary supervision of the density map.

During training, all parameters are jointly optimized in an end-to-end manner. Through backpropagation, gradient updates simultaneously refine both the density map generation branch and the backbone network. This collaborative optimization enables the density branch to produce more reliable spatial priors, which improves subsequent detection performance.

\subsection{Dense Area Focusing Module}
Due to the locally dense yet globally sparse distribution characteristics of tiny objects, traditional global attention mechanisms need to perform uniform calculations on the entire image, which not only generates substantial redundant computation consumption, but also easily loses key features of dense areas due to background noise interference. Therefore, we designed the DAFM to guide the model to focus on the clustered areas of tiny objects, thereby enhancing the capability to represent and locate dense tiny objects.

Specifically, DAFM first uses shallow features (such as the C2 layer) and the density map generated by DGB as input. The size of the density map is adjusted to the corresponding size using bilinear interpolation, and the overall processing is shown below.

\begin{equation}
X’ = \text{DAFM}(X_{\text{C2}}, \mathcal{I}(D))
\end{equation}
where $\mathcal{I}$ represents the bilinear interpolation operation, $X$ denotes the feature (C2 in this formulation), and $X’$ is the output.

Next, we leverage the density map to effectively quantify spatial clustering patterns, where regions with densely clustered objects exhibit higher activation responses. Based on this spatial correlation, we design an adaptive region filtering mechanism. The initial data screening process is formulated as follows.

\begin{equation}
M[i,j]=\begin{cases}
1, & D[i,j]\geq\tau\\
0, & \text{otherwise}
\end{cases}\label{M}
\end{equation}
where $\text{D}$ represents the density map. $\tau$ represents a predefined adaptive threshold representing the minimum proportion of locations whose values exceed the threshold. This operation generates the initial mask map $M$, which is subsequently employed to identify regions of clustered object features.

However, the retained points are spatially discrete, requiring clustering to establish coherent spatial relationships. Given the distribution characteristics of dense tiny objects in remote sensing images, we employ the K-Means algorithm to perform binary clustering on the initial mask $M$, using the Euclidean distance between points as the similarity metric. Due to discrete points inevitably resulting in partial foreground information loss, we merge the minimum bounding rectangles corresponding to the two clusters to form a unified set of rectangles. The initial binary mask $M$ is then updated based on this rectangle set to generate the refined clustered region mask $M'$. The complete implementation process is summarized in Algorithm \ref{ag1}.

\begin{algorithm}
\caption{Region Selection from Density Map}
\begin{algorithmic}[1]
\Require Binary density map $M \in \{0,1\}^{h \times w}$
\Ensure Refined binary map $M'$
\State $P \gets \{(i,j) \mid M[i,j] = 1\}$ \Comment{Active points set}
\If{$P = \emptyset$} \Return $\mathbf{0}^{h \times w}$ \EndIf
\State $\mathbf{c}_1 \gets (1,1)$, $\mathbf{c}_2 \gets (h,w)$
\State $L \gets \text{KMeans}(P, \{\mathbf{c}_1, \mathbf{c}_2\})$

\For{$k \in \{1,2\}$}  
    \State $r_{\min}^k \gets \min\limits_{(i,j) \in P} \{i \mid L(i,j)=k\}$
    \State $r_{\max}^k \gets \max\limits_{(i,j) \in P} \{i \mid L(i,j)=k\}$
    \State $c_{\min}^k \gets \min\limits_{(i,j) \in P} \{j \mid L(i,j)=k\}$
    \State $c_{\max}^k \gets \max\limits_{(i,j) \in P} \{j \mid L(i,j)=k\}$
\EndFor

\State $M' \gets \mathbf{0}^{h \times w}$
\For{$k \in \{1,2\}$}
    \For{$x \gets r_{\min}^k$ \textbf{to} $r_{\max}^k$, $y \gets c_{\min}^k$ \textbf{to} $c_{\max}^k$}
        \State $M'[x,y] \gets 1$
    \EndFor
\EndFor
\State \Return $M'$
\end{algorithmic}
\label{alg:density_map_refinement}
\label{ag1}
\end{algorithm}

In the above process, directly applying the $\otimes$ operation on the input $X$ still retains a relatively large number of pixels. To further improve computational efficiency, we adopt $f^{7\times7}_{\text{pool}}(\cdot)$ to compress the filtered features. 

\begin{equation}
\begin{aligned}
N &= f^{1 \times 1}_{\text{conv}} \left( f^{7 \times 7}_{\text{pool}}(M'\otimes X) \right) \\
\end{aligned}
\label{denseBank}
\end{equation}

In object detection, attention mechanisms enhance feature representation by strengthening responses related to the object while suppressing background interference. However, conventional self-attention mechanisms calculate pairwise similarities across all spatial locations in the feature map, leading to computational complexity that grows quadratically with spatial resolution. To overcome this limitation, inspired by the agent-based attention mechanism \cite{agent-trans}, we design a dual-stream focused attention called Interactive Feature Attention Module (IFAM), which decomposes global dense attention into two lightweight stages. Specifically, the input features $X$ are first projected into vectors $Q_{X}$, $K_{X}$, and $V_{X}$. The focused knowledge base $N$ serves as a query $Q$. The simplified computational process is formalized in Equation \ref{stage_1qkv}.


\begin{equation}
\begin{aligned}
\text{O}_{A} &= sigmoid\left( \frac{N K_X^\top}{\sqrt{d}} + b_{A \to X} \right) 
\end{aligned}
\label{stage_1qkv}
\end{equation}
where $b_{A \to X}$ is a learnable bias term that enables the model to adaptively modulate the attention according to the input features, thereby enhancing flexibility in modeling the associations between queries and the knowledge base. ${O}_{A}$ is the aggregated attention features, which simultaneously encode the global information and the holistic representation of the attention features. To further preserve original feature information and enable deeper contextual integration, a second-stage attention computation is introduced:

\begin{equation}
\begin{aligned}
\text{Y} &= sigmoid\left( \frac{ Q_X\text{N}^\top}{\sqrt{d}} + b_{A \to X} \right)·\text{O}_{A}
\end{aligned}
\label{Stage_2qkv}
\end{equation}







In this stage, attention features are used as keys to form a key-value vector $K$. ${O}_{A}$ serves as the value, which, through interaction with the original query $Q_X$, can selectively retrieve the most relevant global context information for the object region. Further application of the bias term $b_{A \to X}$ refines the attention weights, achieving precise alignment between the global context and the object features. By computing with the original query vector $Q_X$, the final two-stage attention computation result $Y$ is obtained, enriching the input features with corresponding global information. Overall, this two-stage mechanism effectively decouples dense interactions, significantly reducing computational complexity while modeling long-distance dependencies between the object region and the global context.

The dual-stream attention mechanism effectively captures long-range dependencies among densely distributed objects and enhances the model’s focus on object-relevant features. However, it struggles to preserve the original feature structure. To overcome this limitation, we introduce a depth-separable convolution to extract local feature information. The operation is formulated as follows:

\begin{equation}
\begin{aligned}
X'=\mathbf{Y}+DWConv(X)
\end{aligned}
\label{Fineout}
\end{equation}

Here, DWConv(·) represents a depthwise separable convolution operation, which decomposes the original features with low computational cost and retains the local structural information within the channels. It supplements the lost spatial details at the bottom layer denoted by $Y$.


\subsection{Dual Filter Fusion Module}
Multi-scale features tend to lose high-frequency details and suffer from foreground-background confusion. During downsampling, critical information such as the edges and textures of tiny objects (high-frequency components) is often overshadowed by low-frequency semantic information from the background. Furthermore, traditional multi-scale fusion typically integrates features through simple weighted summation, which cannot dynamically emphasize informative regions according to the object density distribution, thereby inadvertently amplifying background noise. To address these issues, we propose a Dual-Filter Fusion Module (DFFM), which fuses multi-level features (P2–P4) to enhance the utilization of global information. The overall architecture is shown in Fig. \ref{Fig.3}(c).

As previously mentioned, the information provided by tiny objects is extremely limited. A further critical factor contributing to suboptimal detection performance is insufficient suppression of background noise \cite{zaosheng}, which often results in unclear boundaries of predicted boxes in regions with aggregated objects. To address this challenge, inspired by \cite{denseyindao}, our DFFM module leverages the interaction between features extracted from the backbone network and those from the density map generation branch. This enables the acquisition of features with clearer boundaries and enhanced foreground information.

Taking the feature map $\mathbf{P}_3$ as an example, the feature tensor $\mathbf{P}_3 \in \mathbb{R}^{B \times C \times H \times W}$ is first duplicated into four identical copies. One copy retains its original resolution, while the other three are processed using pooling kernels of sizes $3 \times 3$, $6 \times 6$, and $9 \times 9$, respectively, to extract multi-granularity contextual information. The operation can be formulated as follows.

\begin{equation}
H_i  = 
\begin{cases} 
\mathrm{EDH}[\mathrm{AVG}_{k \times k}(P_3), D'] & k \in \{3,6,9\}, \, i=1,2,3 \\
\mathrm{Conv}_{3 \times 3}(P_3) & i=4 
\end{cases}
\end{equation}
\begin{equation}
\begin{aligned}
\mathbf{D'} = \sigma(\mathrm{Conv}_{1 \times 1}(\mathrm{ReLU}(\mathrm{Conv}_{3 \times 3}(\mathbf{D}))))
\end{aligned}
\label{bianjie-3}
\end{equation}
where, $\mathbf{D} \in \mathbb{R}^{B \times 1 \times H \times W}$ represents the density map obtained from the DGB, which approximately indicates the spatial concentration of objects at each location. Unlike traditional object detection tasks, boundary features and tiny object features in tiny object detection tasks primarily represent high-frequency information \cite{gaoping}. $\mathbf{D'}$ is the result of probabilistic calibration on the density map $\mathbf{D}$, which enhances the contrast between foreground and background features. 

Each copy feature will be transformed to the frequency domain by a dedicated DCT filter. The density map $D$ is also utilized to generate a low-frequency mask $M_{\text{low}}$ and a high-frequency mask $M_{\text{high}}$. Then, the high- and low-frequency components $\mathcal{F}_{\text{high}}$ and $\mathcal{F}_{\text{low}}$ are obtained through pairwise interactions. The process is described as follows:
\begin{align}
\mathcal{F}_{\text{low}} &= \text{DCT}(F) \odot M_{\text{low}} \\
\mathcal{F}_{\text{high}} &= \text{DCT}(F) \odot M_{\text{high}}
\end{align}
\begin{equation}
\mathbf{M}_{\text{low}} = \sigma\left(\mathbf{W}_{\text{low}} \cdot (\mathbf{P} \odot \mathbf{D}) \right)
\end{equation}
\begin{equation}
\mathbf{M}_{\text{high}} = 1 - \mathbf{M}_{\text{low}}
\end{equation}

The low-frequency components provide structural information, while the high-frequency components capture critical edge and texture information. However, traditional methods for processing high- and low-frequency information rely on fixed rules, which lack dynamic adaptability and sufficient interaction between high and low frequencies. Since tiny objects are highly sensitive to boundary information, this limitation may adversely affect detection performance. Therefore, to explore the complementarity of information at different frequencies and enhance the model's ability to represent complex scenes and intricate details, the density map is still used to guide the model in constructing attention weights to suppress background noise and enhance foreground representation. We design a density map guided frequency-aware feature fusion module, termed EDH. This module explores the complementarity of information across different frequency bands, enhancing the model's ability to represent complex scenes and fine details. Specifically, density maps are utilized to guide the construction of attention weights, suppress background noise, and enhance foreground representation. The implementation is formulated as follows:
\begin{equation}
\begin{aligned}
\mathbf{F}_{l} = \mathcal{CA}(\text{IDCT}(\mathcal{F}_{\text{low}}))
\end{aligned}
\end{equation}
\begin{equation}
\begin{aligned}
\mathbf{F}_{h} = \mathcal{SA}(\text{IDCT}(\mathcal{F}_{\text{high}}))
\end{aligned}
\end{equation}
\begin{equation}
\mathbf{A} = \text{Softmax}\left( (\mathbf{W}_h \mathbf{F}_h \odot \mathbf{D'})^\top (\mathbf{W}_l \mathbf{F}_l \odot ({1-\mathbf{D'}})) \right)
\end{equation}
\begin{equation}
\mathbf{H}_{i} = \left(\mathbf{A}\cdot{\mathbf{P}}_{{i}} \right)    \oplus \mathbf{D'}
\end{equation}
where, $\text{IDCT}(\cdot)$ represents the inverse discrete cosine transform, which converts information from the frequency domain to the spatial domain. $\mathcal{CA}$ and $\mathcal{SA}$ denote the channel attention and spatial attention operations, respectively. Notably, we incorporate the guidance of the density map into the correlation computation of high- and low-frequency information, thereby enabling more accurate subsequent weight calculations. Finally, the integrated feature $H_{i}$ is obtained.

While foreground information is constructed across feature dimensions, a significant challenge is that multi-pooling operations inevitably lead to the loss of scale-related information. To address this, we aggregate and integrate the feature outputs $H_1-H_4$ obtained through multi-level processing. By leveraging the mutual compensation of attention information across layers, we mitigate the inherent discrepancies between features at different levels, thereby ensuring a more consistent and robust representation. This approach effectively compensates for scale information loss and preserves comprehensive scale features, including tiny objects and their surrounding contexts.

\ifCLASSOPTIONcaptionsoff
  \newpage
\fi

\section{EXPERIMENTS}
\subsection{Datasets and Evaluation Metrics}
\noindent{\textbf{AI-TOD}}. 
AI-TOD \cite{aitod} is a highly influential and authoritative dataset for tiny object detection, playing a pivotal role in advancing research on aerial imagery analysis. The average object size in AI-TOD is merely 12.8 pixels, and this extremely small size makes detection much more difficult and challenging than general object detection in remote sensing. The dataset consists of 28,036 aerial images containing 700,621 object instances across eight classes, including aircraft, bridges, and oil tanks. There are 11,214 images used for training, 2,804 images for validation, and 14,018 images for testing, providing a comprehensive benchmark for evaluating detector performance.

\noindent{\textbf{DTOD}}. 
DTOD \cite{shijie} is another large-scale benchmark dataset for dense tiny object detection, which contains 11,600 images (9,280 used for training and 2,320 for testing) and 1,019,800 instances. On average, each image contains 88 objects, with a maximum of 11,112 objects in a single image. The average object size is less than or equal to 13 pixels. Among them, extremely tiny objects (less than 12 pixels) account for 48.2\%, and relatively tiny objects ($12 < \text{area} \leq 20$) account for 40.4\%, meaning over 88\% of objects fall into tiny-object categories. The dataset covers diverse scenes such as parking lots and ports, and includes interfering conditions such as dense fog. DTOD is primarily divided into two categories: vehicles and ships. Image resolutions of DTOD exhibit significant variability, with the dimensions of some images reaching up to $4700 \times 2700$ pixels, which poses higher demands on the model's ability to detect tiny objects.

\noindent{\textbf{Evaluation Metrics}}.
In our experiments, the evaluation strictly follows the COCO computing protocol\cite{cocojisuan}, where the Average Precision ($AP$) is calculated to quantify the detection performance of the model. Following the evaluation protocol of the AI-TOD dataset, we define the evaluation basis based on the Intersection over Union (IoU) matching criteria: True Positive (TP) refers to correctly detected positive samples, while False Positive (FP) and False Negative (FN) denote misidentified positive samples and missed negative samples, respectively. We adopt the $AP$ series metrics to evaluate the performance of various detectors. Specifically, $AP$ represents the average value of 10 $AP$ scores ($AP_{50}$ to $AP_{95}$) with IoU thresholds ranging from 0.5 to 0.95 in steps of 0.05. $AP_{50}$ and $AP_{75}$ correspond to the average precision when the IoU threshold is set to 0.5 and 0.75, respectively. Additionally, to distinguish performance across different object scales, we introduce four metrics: $AP_{vt}$, $AP_{t}$, $AP_{s}$, and $AP_{m}$. $AP_{vt}$ is designed for objects smaller than $8 \times 8$ pixels, $AP_{t}$ for objects sized between $8 \times 8$ and $16 \times 16$ pixels, $AP_{s}$ for objects in the range of $16 \times 16$ to $32 \times 32$ pixels, and $AP_{m}$ for objects larger than $32 \times 32$ pixels.

\subsection{Implementation Details}
Our experiments were implemented based on PyTorch and the MMDetection toolkit, with all experiments conducted using PyTorch on an RTX A40 GPU equipped with 48GB of memory. The main detection framework of DRMNet adopts the YOLOv8 network architecture. During the model training phase, the image size requirements vary across different datasets. For the AI-TOD dataset, the image size is uniformly set to 800 pixels during training. In contrast, for the DTOD dataset, which contains images with varying resolutions and imposes higher requirements, the image size is fixed at 1024 pixels during both training and inference. Images are not cropped before training, and the overall model follows a single-stage structure. The model optimization is performed using the Stochastic Gradient Descent (SGD) optimizer, with an initial learning rate of 0.01 and a momentum of 0.9. All methods based on the MMDetection framework are trained for 24 epochs to ensure a fair comparison. In contrast, methods based on the YOLO framework are trained for 100 epochs with a batch size of 8. To ensure fairness, all compared methods in the experiments are fine-tuned to achieve the best results. Additionally, considering that the DTOD dataset contains images with a large number of objects, the maximum number of objects allowed per image during detection is set to 1500.

\begin{table*}
\caption{On the AI-TOD dataset, we compare our proposed method with state-of-the-art object detection methods. $AP_{50}$ calculates the average accuracy for all classes at IoU = 0.50. $AP_{75}$ calculates the average accuracy for all classes at IoU = 0.75. $AP_{50}$, $AP_{vt}$, $AP_{t}$, $AP_{s}$ and $AP_{m}$ calculate the average accuracy for objects of different sizes at IoU = 0.50.}
\scalebox{0.9}{
\begin{tabular}{c|c|cccccccc|c|cc|cccc}
\hline
Method&Backbone&AI&BR&ST&SH&SP&VE&PE&WM&$AP$&$AP_{50}$&$AP_{75}$&$AP_{vt}$&$AP_{t}$&$AP_{s}$&$AP_{m}$\\\hline
Faster R-CNN \cite{can-1}&ResNet-50&19.6&1.4&17.3&17.4&7.9&10.4&3.6&0.0&9.7&22.1&6.8&0.0&5.2&21.7&32.5\\
Mask R-CNN \cite{can-2}&ResNet-50&20.0&1.9&17.4&18.6&8.0&10.9&3.8&0.0&10.1&22.7&7.6&0.0&5.8&22.1&32.6\\
PAFPN \cite{can-3}&ResNet-50&19.8&1.4&17.0&17.5&7.7&10.4&3.7&0.0&9.7&22.2&6.8&0.0&5.4&21.6&32.5\\
YOLOv3 \cite{can-4}&DarkNet-53&19.1&9.7&21.7&20.9&4.7&14.5&5.2&2.7&12.3&35.3&5.6&3.2&12.0&18.5&24.2\\
TOOD \cite{can-5}&ResNet-50&22.2&5.3&28.9&27.9&9.1&18.6&7.0&0.5&14.9&34.7&10.6&3.3&12.8&21.7&33.1\\
ATSS \cite{can-6}&ResNet-50&0.5&9.8&16.8&26.4&0.6&12.0&4.0&1.2&8.9&22.5&5.3&2.1&10.1&11.4&15.0\\
AutoAssign \cite{can-7}&ResNet-50&14.7&10.7&25.5&26.4&4.1&16.5&6.0&3.0&13.4&36.2&6.9&3.5&13.5&17.0&24.9\\
FCOS \cite{can-8}&ResNet-50&17.2&1.6&21.2&19.8&0.8&13.3&4.9&0.0&9.8&24.1&6.2&1.2&7.8&16.1&27.2\\
RepPoints \cite{can-9}&ResNet-50&0.0&0.0&12.7&16.1&0.0&10.6&2.4&0.0&5.2&15.2&2.2&1.3&5.4&7.5&13.7\\
FoveaBox \cite{can-10}&ResNet-50&14.7&0.0&19.2&17.7&0.0&12.4&4.2&0.0&8.5&20.6&5.7&0.9&6.1&14.1&27.2\\
CenterNet \cite{can-11}&ResNet-18&12.8&2.0&15.3&15.9&1.7&11.7&4.6&1.5&8.2&24.3&3.4&1.3&6.9&12.9&22.3\\
KLDNet \cite{can-12}&ResNet-50&13.6&18.7&35.7&42.3&5.2&24.9&9.3&6.7&19.6&-&-&8.4&20.6&22.7&26.4\\
M-CenterNet \cite{can-13}&DLA-34&18.6&10.6&27.6&22.3&7.5&18.6&9.2&2.0&14.5&40.7&6.4&6.1&15.0&19.4&20.4\\
FSANet \cite{can-14}&Swin-T&30.9&15.1&35.0&40.3&19.8&24.9&8.9&5.6&22.6&52.8&15.6&7.4&21.6&29.1&38.5\\
DetectorRS w/RFLA \cite{can-15}&ResNet-50&27.6&17.1&36.8&44.8&15.5&24.7&11.3&5.8&23.0&52.8&16.4&8.8&23.1&28.3&36.9\\
MENet \cite{can-16}&Swin-T&27.5&16.4&37.4&42.1&18.8&24.9&10.2&8.2&23.2&56.2&15.0&9.7&23.9&25.3&34.4\\
DNTR \cite{can-17}&ResNet-50&-&-&-&-&-&-&-&-&26.2&56.7&20.2&12.8&26.4&31.0&37.0\\
CAFENet-M \cite{can-18}&ResNet-50&39.7&22.8&43.7&50.6&24.6&34.9&17.1&8.9&30.2&63.7&25.0&12.8&30.3&36.7&41.9\\
LTDNet \cite{ltdnet} &ResNet-50&15.1&16.9&29.4&36.0&10.8&23.5&7.9&4.6&18.0&46.3&10.2&6.1&18.8&21.9&26.1\\
BAFNet \cite{can-19}&ResNet-50&34.5&22.7&39.4&68.1&29.3&28.1&14.9&6.9&30.5&59.8&26.6&16.6&31.3&35.1&40.5\\\hline
YOLOv8-M  \cite{can-20} & DarkNet-53 & 32.8 & 22.0 & 38.5 & 62.7 & 20.5 & 23.0 & 12.5 & 4.7 & 30.1 & 62.1 & 24.3 & 10.6 & 28.4 & 40.2 & 46.5 \\
YOLOv11-M \cite{can-21} & DarkNet-53 & 33.0 & 23.5 & 35.4 & 63.2 & 21.4 & 22.3 & 12.1 & 6.7 & 30.5 & 63.5 & 25.0 & 10.3 & 29.7 & 40.5 & 45.5 \\
YOLOv12-M \cite{can-22} & DarkNet-53 & 28.4 & 17.2 & 36.5 & 43.2 & 17.4 & 25.2 & 11.4 & 8.0 & 26.3 & 56.4 & 20.4 & 7.9 & 24.6 & 35.2 & 40.5 \\
DRMNet                  & DarkNet-53 & 38.2 & 23.6 & 44.0 & 52.3 & 25.7 & 33.2 & 16.5 & 7.2 & 31.9 & 65.0 & 26.4 & 13.3 & 32.0 & 39.8 & 41.6 \\\hline
\end{tabular}
}
\label{big-1b}
\end{table*}

\begin{table*}[htbp]
\centering
\caption{Comparison with state-of-the-art object detection methods on the DTOD dataset.$AP_{50}$ denotes the average precision at an IoU of 0.50, and $AP_{75}$ at an IoU of 0.75. The metrics $AP_{50}^{e s}$, $AP_{50}^{r s}$, and $AP_{50}^{N}$ specify the average precisionfor objects defined as extremely small (es), relatively small (rs), and Normal (N), respectively, all at an IoU of 0.50.}
\label{tab:dtod_sota_comparison}
\begin{tabular*}{\linewidth}{@{\extracolsep{\fill}} l c c c c c c c c @{}}
\toprule
Method   & Venue & Backbone   & Param. (M) & $AP_{50}$ & $AP_{75}$ & $AP_{50}^{e s}$ & $AP_{50}^{r s}$ & $AP_{50}^{N}$ \\
\midrule
Faster R-CNN \cite{can-1}      & NeurIPS'2015     & ResNet-50           & 41.13               & 11.6               & 2.1                & 14.8                 & 9.0                  & 22.5                 \\
RetinaNet \cite{retinanet}         & ICCV'2017        & ResNet-50           & 31.47               & 5.2                & 0.7                & 8.9                  & 4.8                  & 8.7                  \\
Cascade R-CNN \cite{cascade}     & CVPR'2018        & ResNet-50           & 68.93               & 9.9                & 1.7                & 12.9                 & 8.0                  & 20.7                 \\
YOLOV3 \cite{can-4}            & ArXiv'2018       & DarkNet-53          & 61.53               & 4.4                & 0.1                & 5.6                  & 4.2                  & 14.1                 \\
FCOS \cite{can-8}             & ICCV'2019        & ResNet-50           & 31.91               & 7.0                & 1.2                & 9.3                  & 5.6                  & 10.7                 \\
RepPoints \cite{repponit}         & ICCV'2019        & ResNet-50           & 31.95               & 5.0                & 0.8                & 7.2                  & 5.1                  & 10.8                 \\
TridentNet \cite{TridenNet}       & ICCV'2019        & ResNet-50           & 32.77               & 2.3                & 1.0                & 3.3                  & 3.4                  & 10.2                 \\
ATSS \cite{atss}              & CVPR'2020        & ResNet-50           & 31.96               & 14.6               & 2.2                & 14.0                 & 11.9                 & 29.7                 \\
YOLOv5 \cite{jocher2021ultralytics}            & CVPR'2020        & CSPDarkNet-s        & 26.80               & 13.8               & 2.4                & 17.6                 & 10.3                 & 29.2                 \\
FoveaBox \cite{foveabox}         & TIP'2020         & ResNet-50           & 36.08               & 12.0               & 1.8                & 12.7                 & 10.2                 & 28.0                 \\
TPH-YOLOV5 \cite{tph}       & ICCV'2021        & CSPDarkNet-s        & 27.60               & 10.1               & 1.9                & 14.8                 & 7.0                  & 24.8                 \\
Dyhead \cite{dyhead}           & CVPR'2021        & ResNet-50           & 38.73               & 10.5               & 1.6                & 12.1                 & 8.2                  & 24.5                 \\
YOLOV8 \cite{can-20}            & TGRS'2022        & CSPDarkNet-s        & 12.00               & 10.0               & 1.7                & 12.2                 & 7.5                  & 23.1                 \\
RT-DETR \cite{rt-detr}          & 2022           & HGNetV2             & 31.29               & 7.6                & 0.6                & 9.4                  & 5.4                  & 18.7                 \\
DDOD \cite{chen2023ddod}              & TMM'2023         & ResNet-50           & 32.04               & 14.5               & 2.2                & 14.9                 & 11.9                 & 28.6                 \\
DINO \cite{dino}              & ICLR'2023        & ResNet-50           & 47.54               & 8.5                & 1.1                & 12.2                 & 7.0                  & 12.6                 \\
SCDNet \cite{zhao2024}           & TGRS'2023        & CSPDarkNet-s        & 43.60               & 24.2               & 3.3                & 21.0                 & 18.3                 & 43.7                 \\
KLDNet \cite{kldet_2}            & TGRS'2024        & ResNet-50           & 45.20               & 13.3               & 2.1                & 13.4                 & 11.3                 & 27.9                 \\
YOLOV10 \cite{yolov10}           & 2024           & CSPDarkNet-s        & 7.69                & 11.9               & 2.3                & 13.6                 & 9.2                  & 20.5                 \\
YOLOV11 \cite{can-21}           & 2024           & C3k2Net-s           & 8.99                & 12.2               & 2.2                & 13.9                 & 9.5                  & 23.8                 \\
LTDNet \cite{ltdnet}           & TGRS'2025           & ResNet-50           & 4.85                & 15.3               & 1.7               & 14.5                 & 7.5                  & 27.4                 \\
\midrule
\textbf{DRMNet (Ours)} & -        & CSPDarkNet-s        & 29.80              & 26.5      & 4.0       & 20.4        & 22.3        & 41.2       \\
\bottomrule
\end{tabular*}
\label{table:2}
\end{table*}

\subsection{Comparative Experiments}
We compared DRMNet with other state-of-the-art object detection models on the AI-TOD and DTOD datasets.

\subsubsection{Main Results on AI-TOD}
As shown in Table \ref{big-1b}, the methods or experimental results mentioned in the table are from literature \cite{can-1, can-11, can-22}. From Table \ref{big-1b}, it can be observed that DRMNet performs well in both general object detection metrics $AP$, $AP_{50}$, and $AP_{75}$, as well as in tiny object detection metrics $AP_{vt}$, $AP_{t}$, and $AP_{s}$. To verify the effectiveness of the proposed DRMNet, we conducted comparative experiments with several advanced object detection algorithms (including some representative state-of-the-art methods) on the AI-TOD dataset under similar training configurations. In terms of model complexity, DRMNet achieved a 1.8\% improvement in $AP$ and a 2.9\% improvement in $AP_{50}$ compared to the baseline model YOLOV8; an 8.7\% improvement in $AP$ compared to the advanced two-stage detector MENet; Furthermore, DRMNet consistently outperformed other competitive methods on the AI-TOD dataset, including M-CenterNet \cite{can-13}, DetectoRS w/RFLA \cite{can-15}, DNTR \cite{can-17}, and FSANet \cite{can-14}, with $AP_{50}$ improvements of 24.3\%, 12.2\%, 8.3\%, and 12.2\%, respectively. Compared to the recent state-of-the-art method BAFNet \cite{can-19}, DRMNet achieved a 5.2\% improvement in $AP_{50}$ and a 1.4\% improvement in $AP$. These results indicate that DRMNet achieves new state-of-the-art performance on the AI-TOD dataset, especially in detecting tiny objects, with more significant improvements in tiny object size categories compared to other methods.

\subsubsection{Main Results on DTOD}
To validate its robustness, we evaluated DRMNet on the challenging DTOD dataset, which features a higher density of tiny objects with more severe aggregation than AI-TOD. For a fair comparison with the state-of-the-art method SCDNet, which utilizes a YOLO-S backbone, we adopted the same backbone for DRMNet.
As shown in Table \ref{table:2}, DRMNet establishes a new state-of-the-art on this benchmark. It surpasses the previous best method, SCDNet, by2.3\%in $AP_{50}$. Furthermore, our model demonstrates significant advantages over other strong baselines. Compared to the Transformer-based DINO and the density-aware DDOD, DRMNet improves $AP_{50}$ by 18\% and 12\%, respectively, highlighting its superiority in dense scenes. The model also consistently outperforms recent YOLO variants (YOLOv8, v10, v11) by a significant margin (up to 16.5\%), addressing their known limitations in detecting densely aggregated tiny objects.
\subsection{Ablation Study}
In this section, we conducted a series of ablation experiments on the AI-TOD and DTOD test sets to demonstrate the effectiveness of our components. DRMNet consists of three components: DGB, DAFM, and DFFM. To assess the contribution of each component, we conducted an ablation study by progressively inserting these modules into the baseline model. The experiments were performed on the AI-TOD dataset. Table \ref{tab:dgb_comparison} summarizes the detection performance under different module configurations. 

\subsubsection{Effectiveness of the DGB}
To validate the effectiveness of the DGB module, we conducted an ablation study by removing it from our network. As shown in Table \ref{tab:dgb_comparison}, this ablation not only nullified the model's gains but caused performance to drop below the baseline. This demonstrates that without the crucial density-map guidance from DGB, subsequent components like DAFM and DFFM lack effective spatial priors, leading to suboptimal feature learning. Specifically, with DGB's guidance, the DAFM module boosts $AP_{50}$ by 3.5\% (from 59.4\% to 62.9\%) and $AP_{vt}$ for tiny objects by 1.5\%. Likewise, the DFFM module achieves a 4.0\% increase in $AP_{50}$ (from 60.2\% to 64.2\%) and a 2.4\% gain in $AP_{vt}$. These results confirm that the DGB-generated density map is indispensable for directing the model's focus toward high-quality foreground regions, significantly enhancing the detection of challenging tiny objects.


\subsubsection{Effectiveness of the DAFM}
We then evaluated the contribution of the DAFM module, which is designed to leverage the DGB-generated density map to refine features in object-dense regions. The results show that integrating DAFM improves both $AP_{50}$ and $AP_{75}$ by 0.8\% over the baseline. Interestingly, while the module boosts performance on small ( $AP_{s}$,  +0.3\%) and tiny ($AP_{t}$, +1.0\%) objects, we observed a slight degradation on very tiny objects  $AP_{vt}$). This suggests that while DAFM effectively aggregates contextual information in crowded areas, its spatial attention mechanism might slightly oversmooth the fine-grained features essential for detecting the most isolated, very tiny objects. Nevertheless, the overall positive gains demonstrate that DAFM successfully translates the density prior into enhanced feature representation. This is further corroborated by feature visualization in Fig. \ref{gram}. Compared to the baseline without DAFM, our full model exhibits a more focused and intense activation in object-dense areas. Moreover, these activated regions show sharper boundaries, which is critical for separating and subsequently detecting isolated objects in crowded scenes.
\begin{table}[!t]
\centering
\caption{Ablation studies of DGB, DAFM, and DFFM on the AI-TOD dataset.}
\begin{tabular}{lcc@{\hspace{2pt}}|@{\hspace{2pt}}ccccc} 
\hline
\textbf{DGB} & \textbf{DAFM} & \textbf{DFFM} & $\boldsymbol{AP_{50}}$ & $\boldsymbol{AP_{75}}$ & $\boldsymbol{AP_{vt}}$ & $\boldsymbol{AP_{t}}$ & $\boldsymbol{AP_{s}}$ \\ 
\hline
- & - & - & 62.1 & 24.3 & 10.6 & 28.4 & 40.2 \\ 
\hline
\multirow{3}{*}{W/o} & \checkmark & - & 59.4 & 22.7 & 8.2 & 26.1 & 38.2 \\
& - & \checkmark & 60.2 & 22.9 & 9.1 & 27.0 & 38.4 \\
& \checkmark & \checkmark & 60.4 & 23.0 & 8.5 & 26.5 & 38.5 \\ 
\hline
\multirow{3}{*}{W/} & \checkmark & - & 62.9 & 25.1 & 9.7 & 29.4 & 39.8 \\
& - & \checkmark & 64.2 & 25.8 & 11.5 & 31.2 & 39.1 \\
& \checkmark & \checkmark & 65.0 & 26.4 & 13.3 & 32.0 & 40.5 \\ 
\hline
\end{tabular}
\vspace{2pt}
\label{tab:dgb_comparison}
\end{table}


\begin{figure}[htbp]
\centering
\begin{minipage}[t]{0.45\textwidth}
  \centering 
  \begin{subfigure}[b]{0.48\linewidth} 
    \centering
    \includegraphics[width=\linewidth]{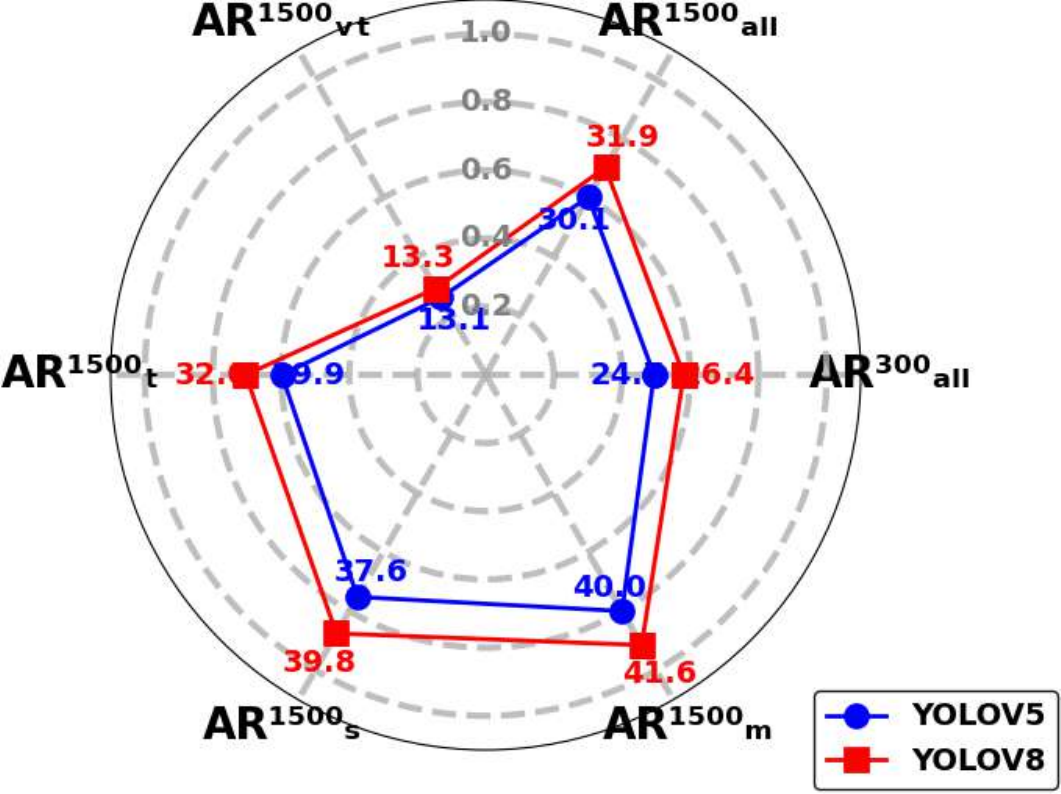}
  \end{subfigure}
  \hfill
  \begin{subfigure}[b]{0.48\linewidth}
    \centering
    \includegraphics[width=\linewidth]{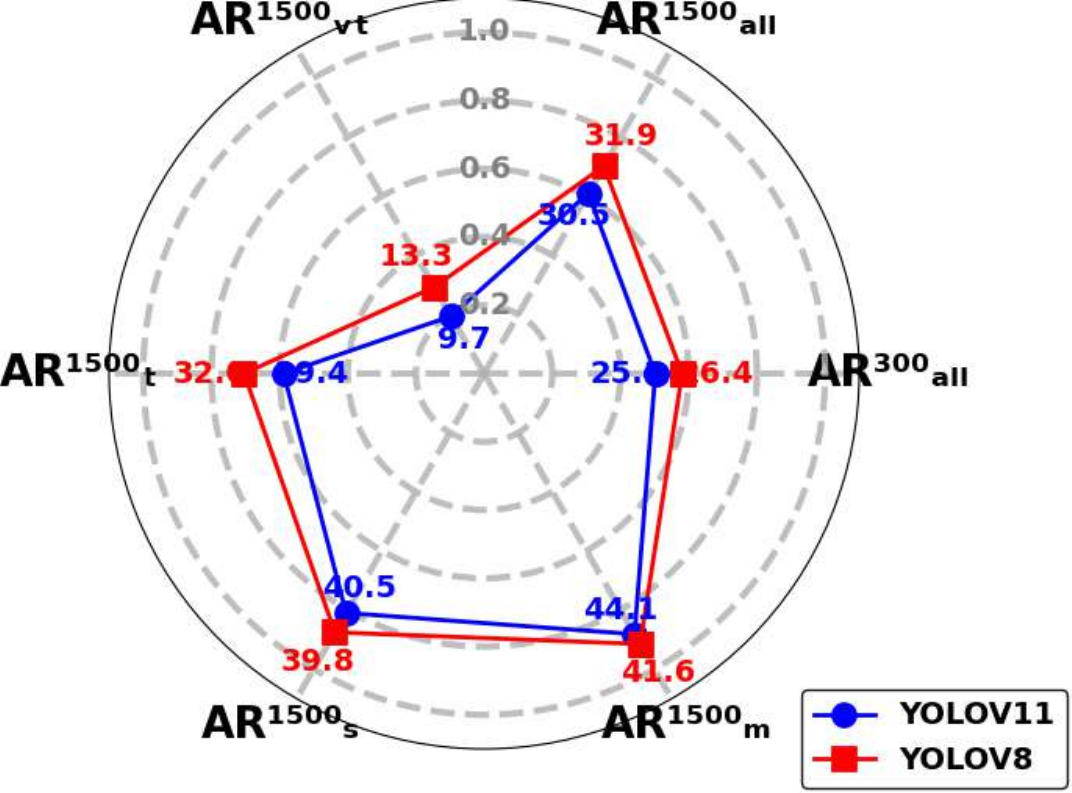}
  \end{subfigure}

  \vspace{10pt}
  \centering
  \begin{subfigure}[b]{0.48\linewidth} 
    \centering
    \includegraphics[width=\linewidth]{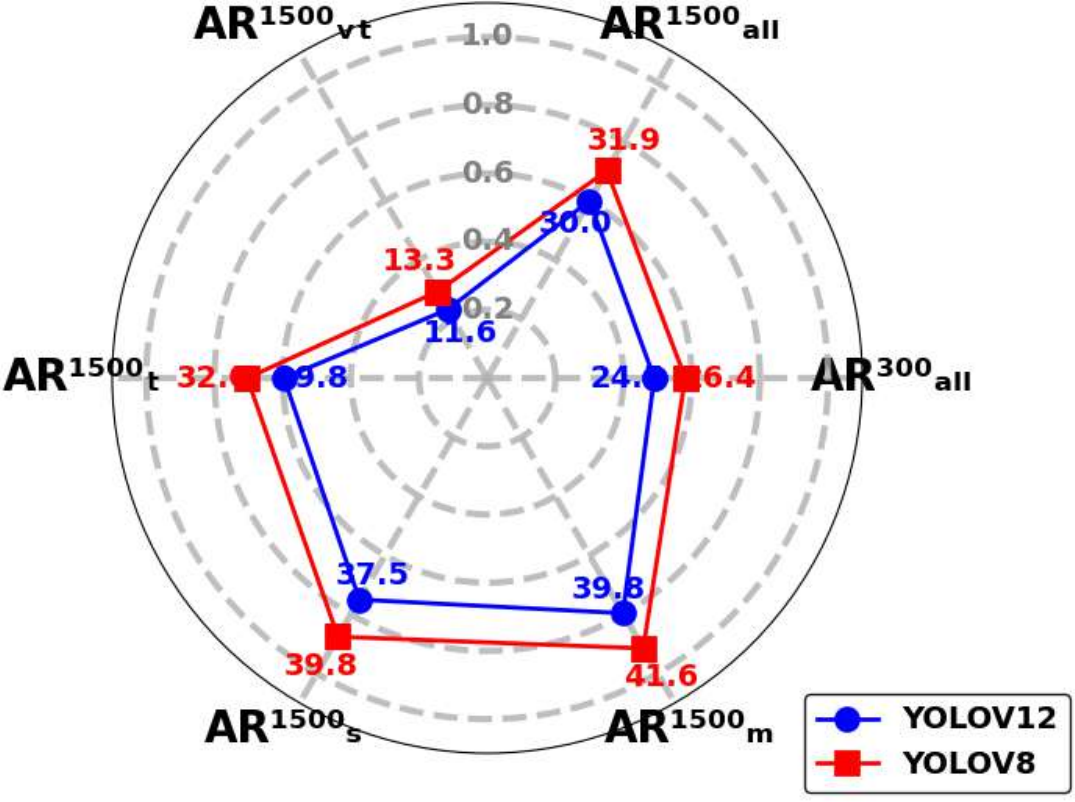}
  \end{subfigure}
\end{minipage}
\hfill 
\caption{AR curve of the AI-TOD test set on different detection backbone networks.}
\label{fig:radar}
\end{figure}

\begin{table}
  \centering
  \caption{The impact on the AI-TOD test set compared to other similar density maps and feature processing methods. Experiments based on DRMNet.}
  \label{tab:fuzadu}
  \begin{tabular}{l|c|c|c|c|c}
    \hline
    Method               & ${AP_{50}}$  & ${AP_{75}}$ & ${AP_{vt}}$ & \#Params & FLOPS  \\
    \hline
    PCF & 62.4 & 24.5 & 11.3  & 0.17M    & 18.7G \\ 
    \hline
    Global-MSA  & 64.8 & \color[HTML]{FE0000} \textbf{27.7} & 8.3  & 0.26M    & 102.4G \\
    \hline
    Agent-Attention                 & \color[HTML]{FE0000} \textbf{65.0} & 26.4 & \color[HTML]{FE0000} \textbf{13.3}  & 0.47M    & 25.36G \\
    \hline
  \end{tabular}
\end{table}

\begin{table}[htbp]
  \centering
  \caption{The impact of different numbers of channels and pooling sizes during high- and low-frequency information extraction. Experiments were conducted on the AI-TOD test set.}
  \begin{tabular*}{\linewidth}{@{\extracolsep{\fill}}c c c c c c@{}}
    \toprule
    Kernel Size & AP & AP$_{50}$ & AP$_{75}$ & AP$_{vt}$ & AP$_{t}$ \\
    \midrule
    (3, 5, 7)        & 29.0 & 60.2 & 24.0 & 12.8 & 28.8 \\
    (5, 7, 9)        & 29.6 & 61.5 & 24.4 & 13.3 & 29.4 \\
    (3, 6, 9)        & \color{red} \textbf{31.9} & \color{red} \textbf{65.0} & \color{red} \textbf{26.4} & \color{red} \textbf{13.3} & \color{red} \textbf{32.0} \\
    (6, 9, 12)       & 29.1 & 60.2 & 23.7 & 11.1 & 28.7 \\
    (6, 7, 8)        & 23.4 & 51.3 & 14.9 & 8.9 & 23.4 \\
    (6, 7, 8, 9)     & 23.8 & 52.3 & 17.9 & 9.3 & 23.8 \\
    (3, 5, 7, 9)     & 29.9 & 62.0 & 24.3 & 13.6 & 29.5 \\
    (3, 6, 9, 12)    & 30.1 & 62.7 & 24.7 & 12.0 & 29.8 \\
    \bottomrule
  \end{tabular*}
  \label{tab:kernel_performance}
\end{table}

\newcommand{\imgwidth}{3.5cm} 
\begin{figure*}[htbp]
    \centering
    \setlength{\tabcolsep}{2pt} 
    \begin{tabular}{>{\centering\arraybackslash}m{\imgwidth}
                    >{\centering\arraybackslash}m{\imgwidth}
                    >{\centering\arraybackslash}m{\imgwidth}
                    >{\centering\arraybackslash}m{\imgwidth}
                    >{\centering\arraybackslash}m{\imgwidth}} 
        \includegraphics[width=\imgwidth]{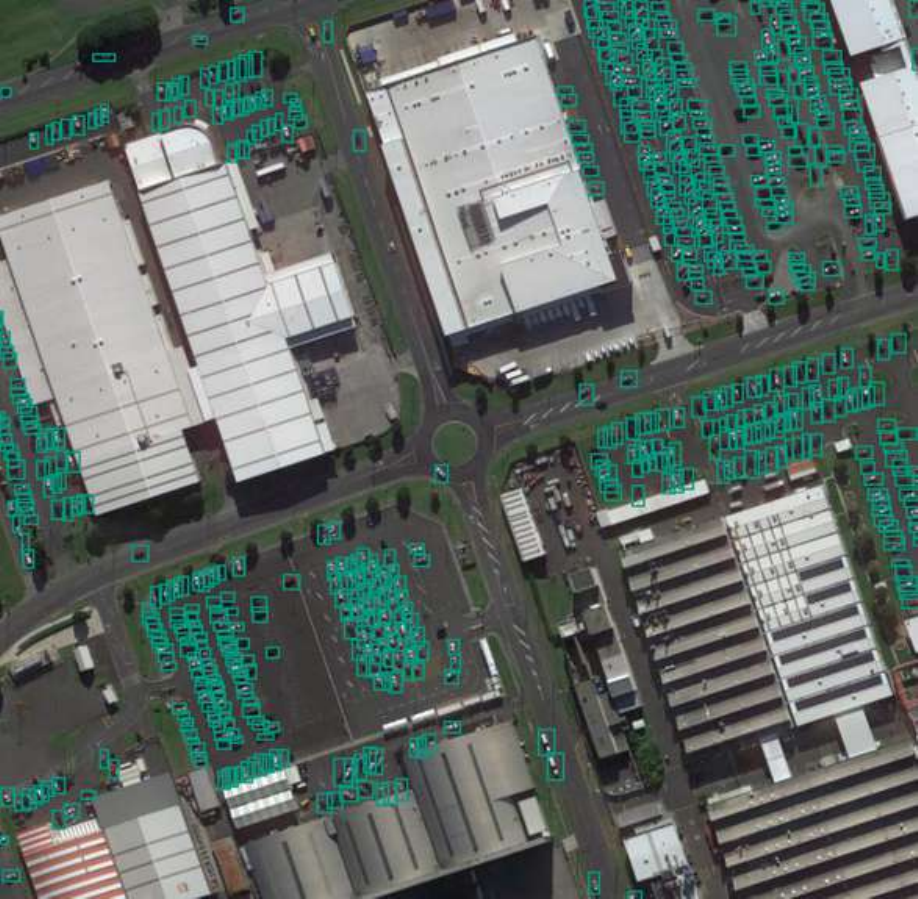} & 
        \includegraphics[width=\imgwidth]{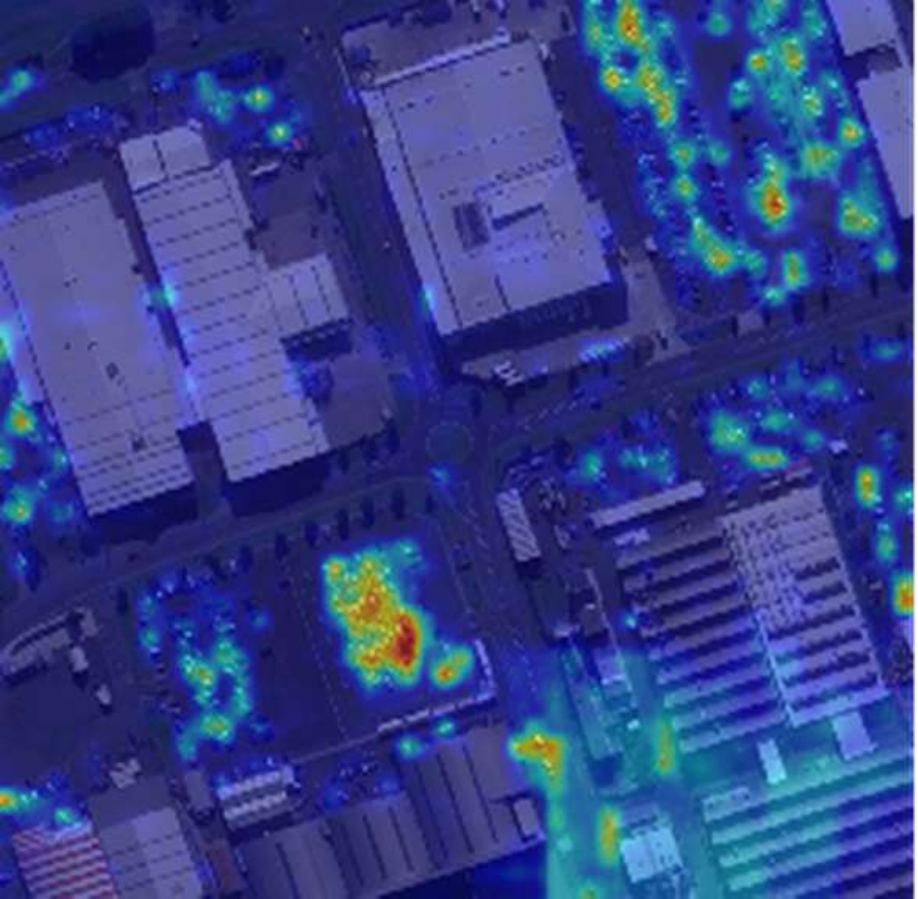} & 
        \includegraphics[width=\imgwidth]{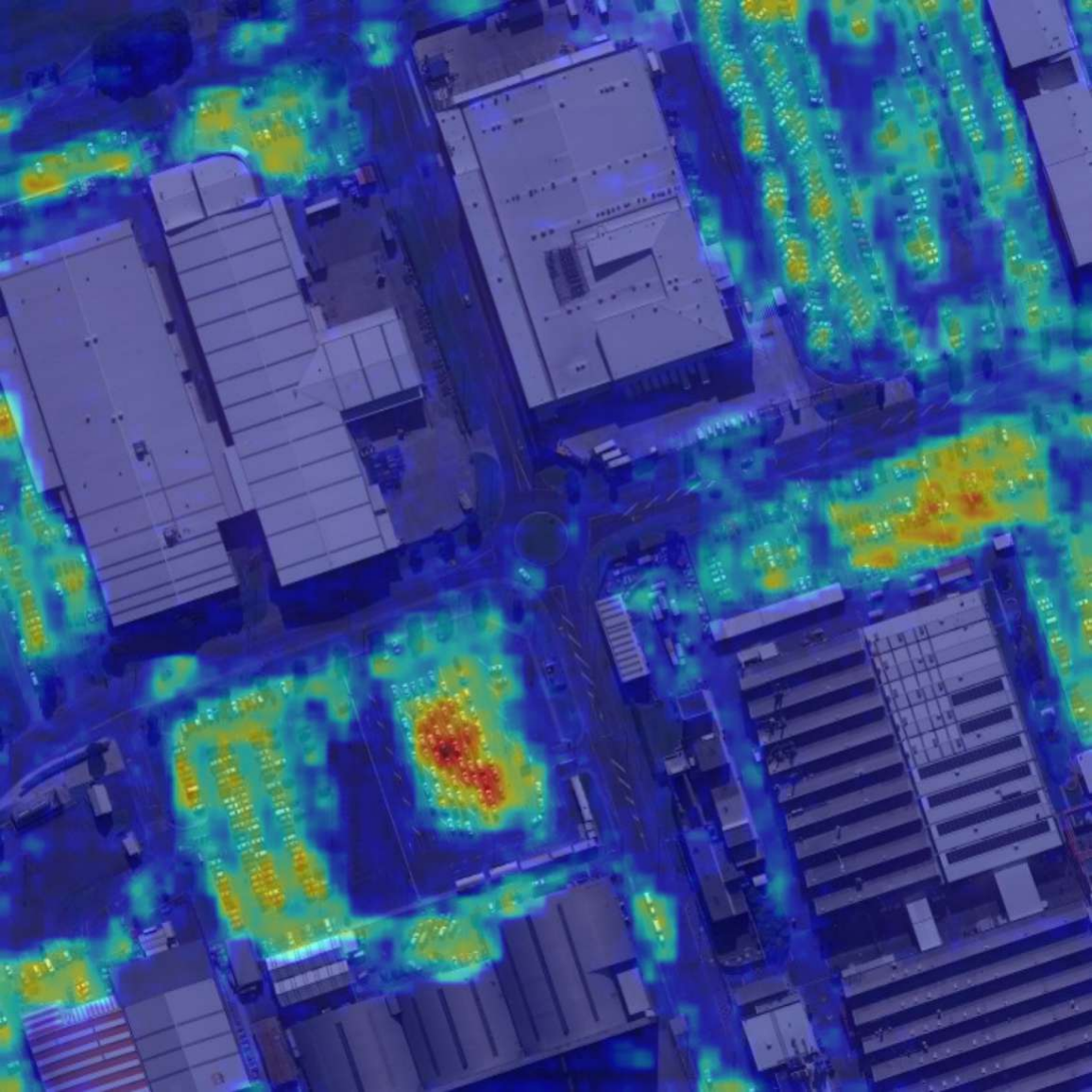} &
        \includegraphics[width=\imgwidth]{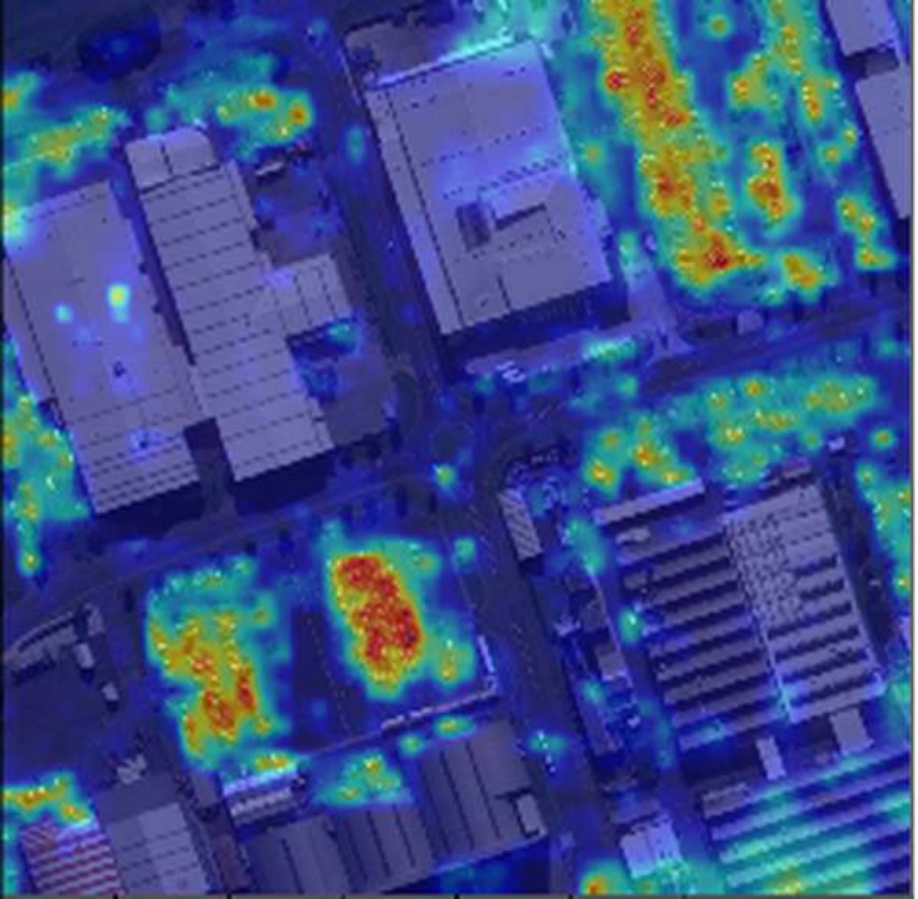} & 
        \includegraphics[width=\imgwidth]{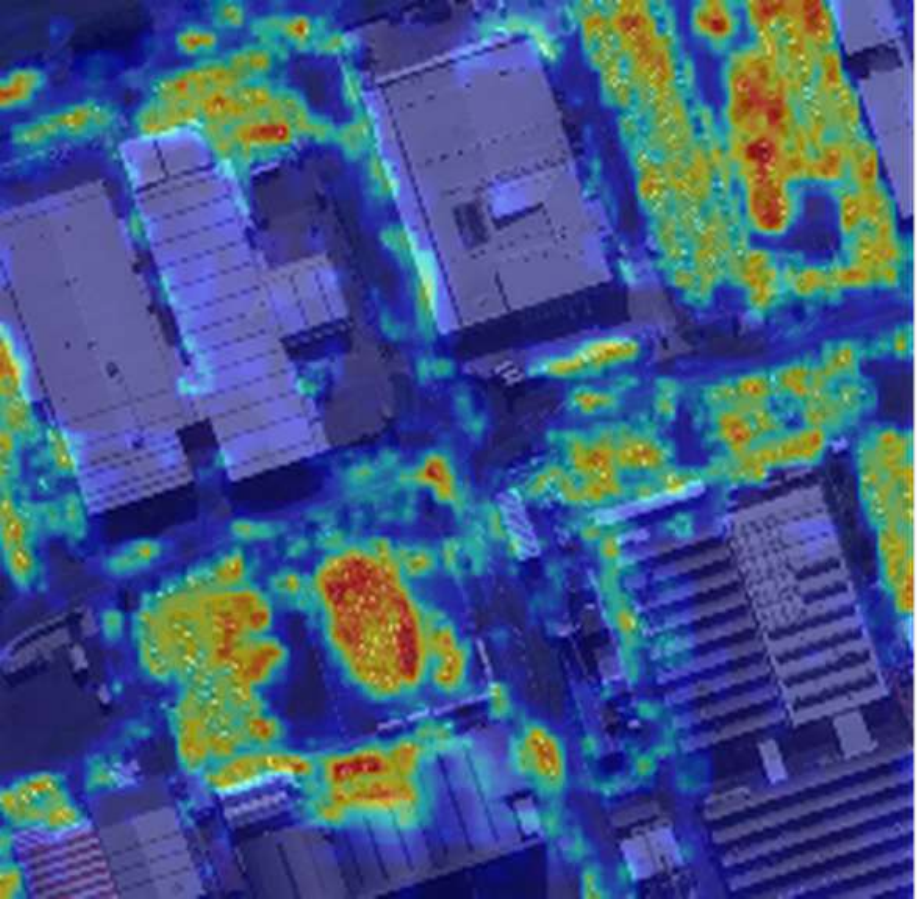} \\[-2pt] 
        
        \includegraphics[width=\imgwidth]{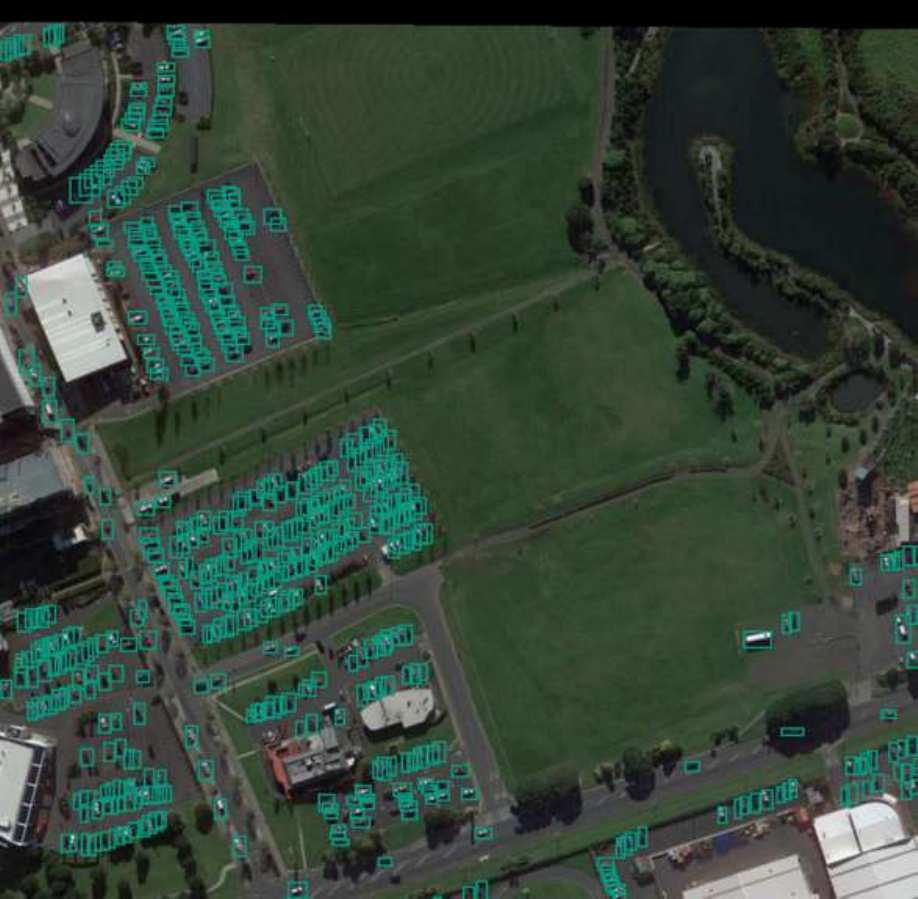} & 
        \includegraphics[width=\imgwidth]{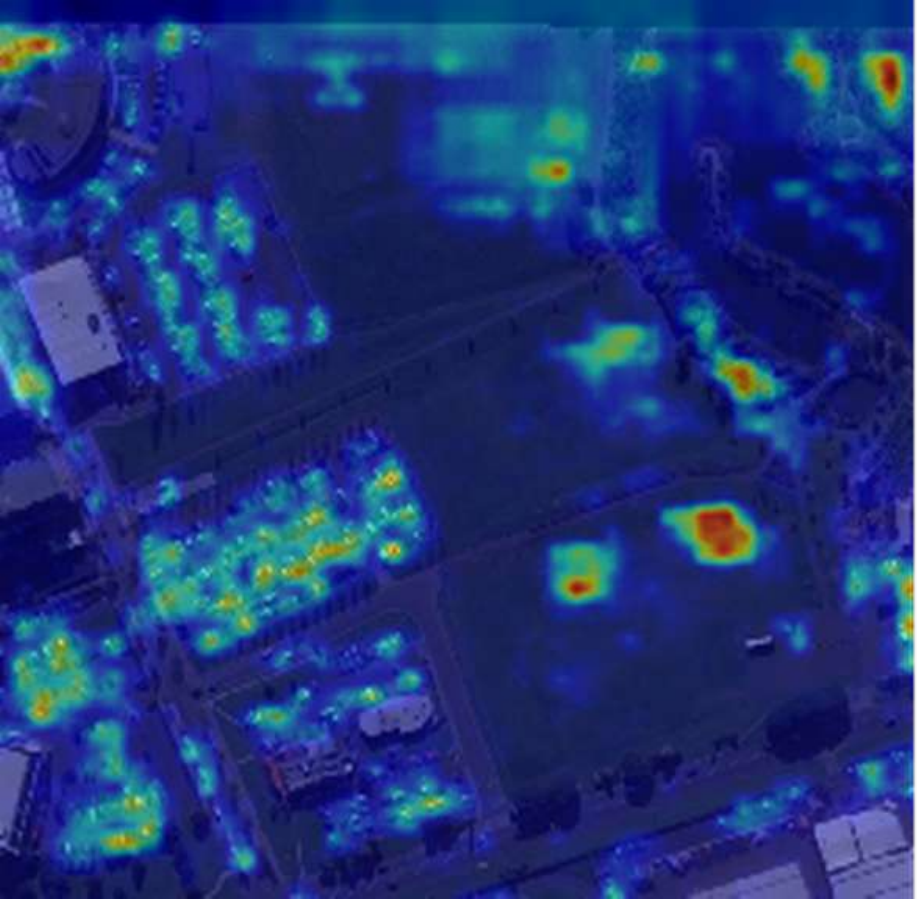} & 
        \includegraphics[width=\imgwidth]{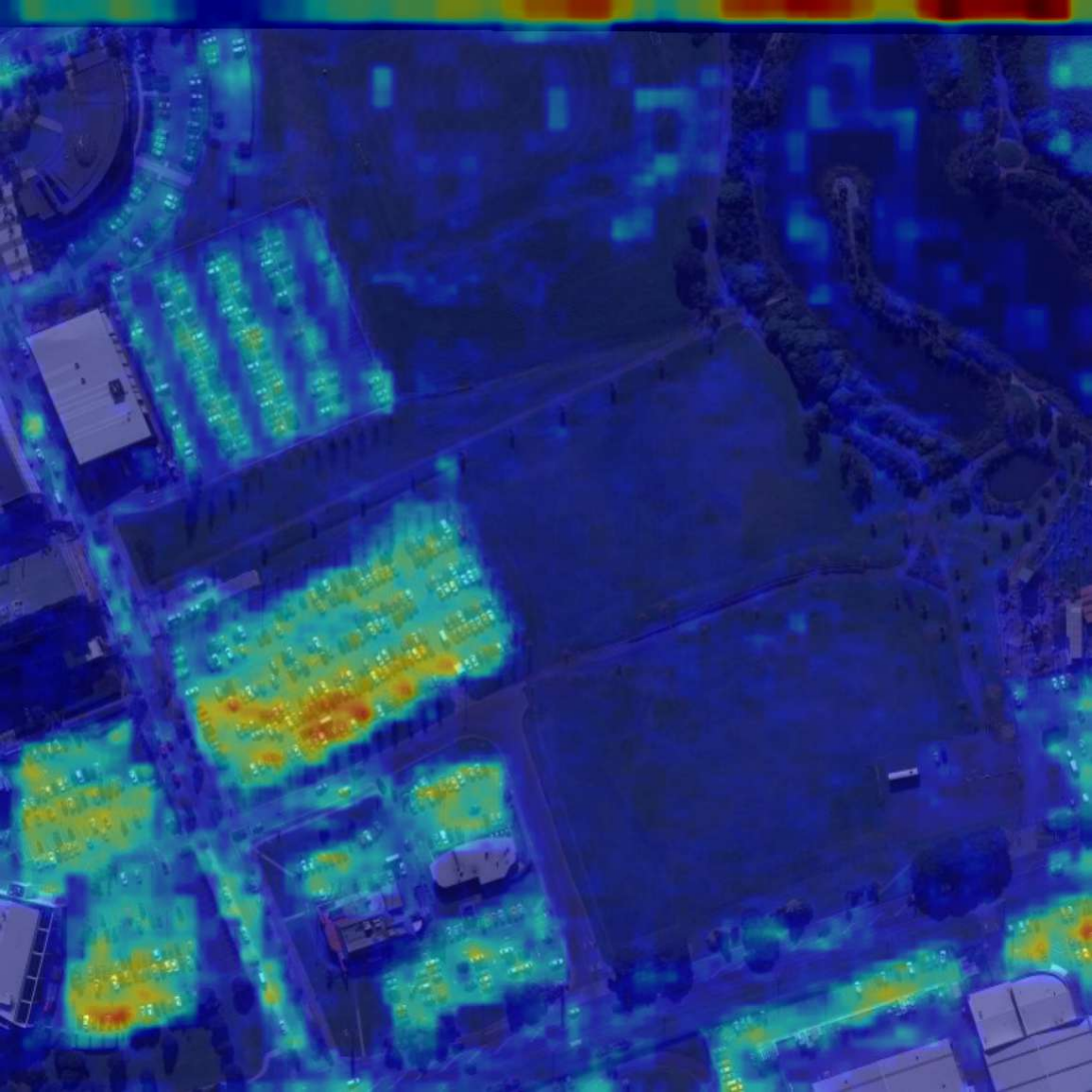} &
        \includegraphics[width=\imgwidth]{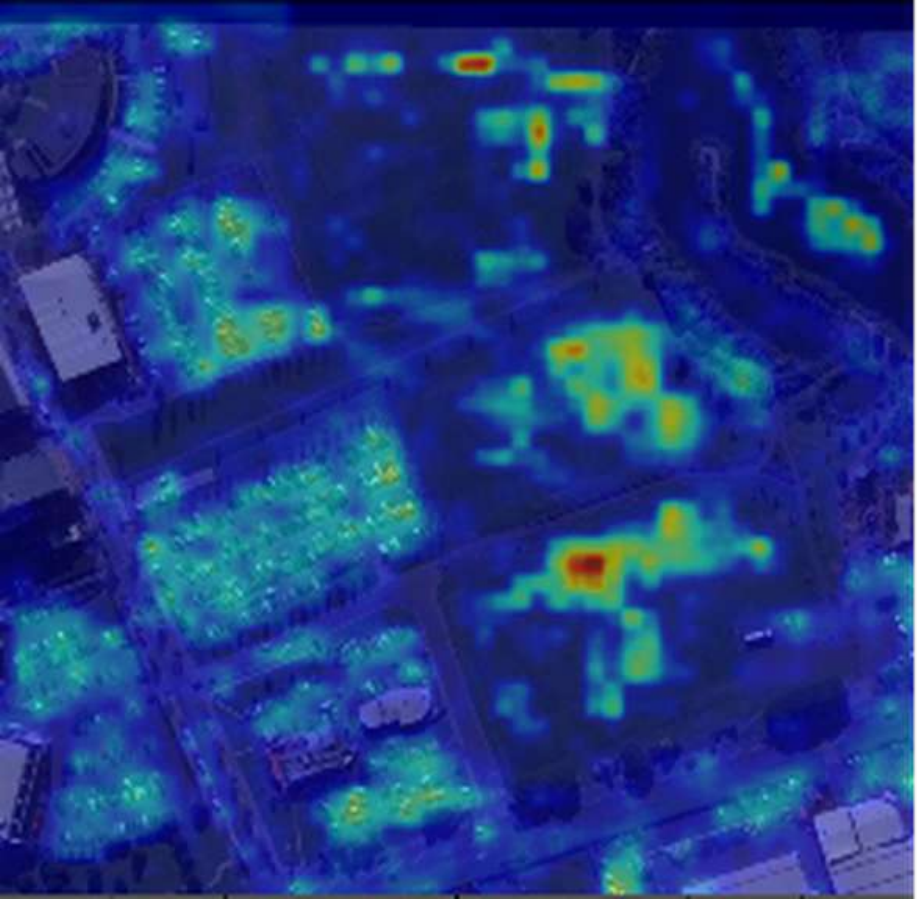} & 
        \includegraphics[width=\imgwidth]{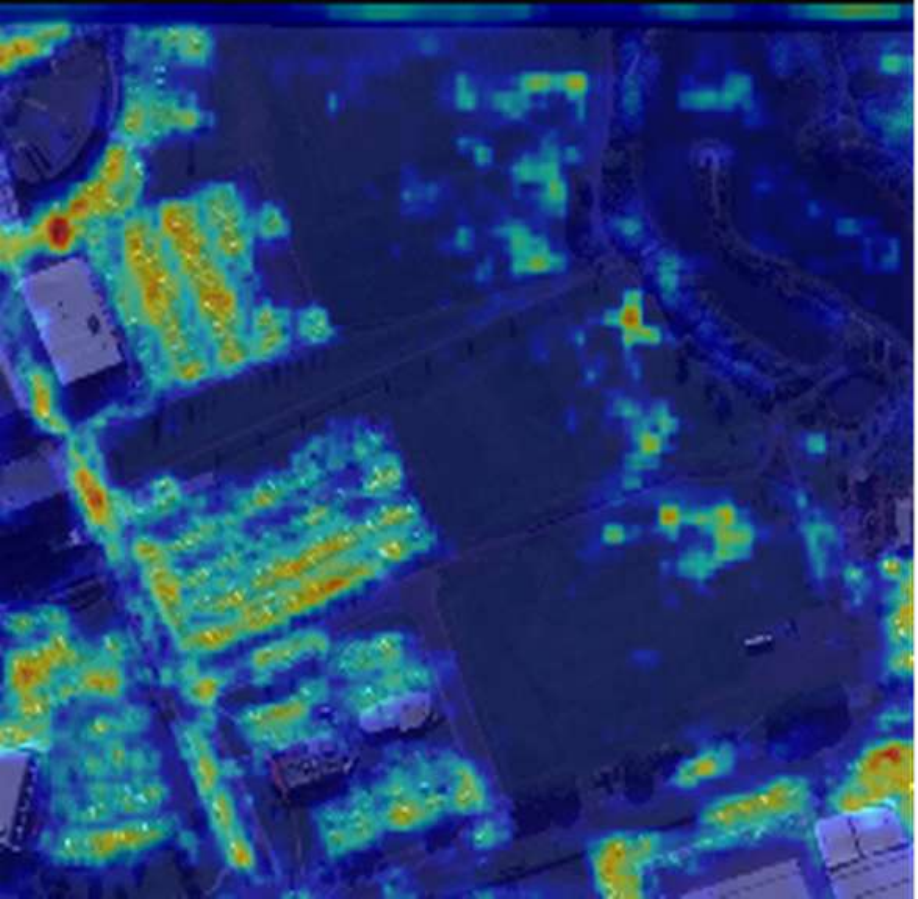} \\[-2pt]
        
        \includegraphics[width=\imgwidth]{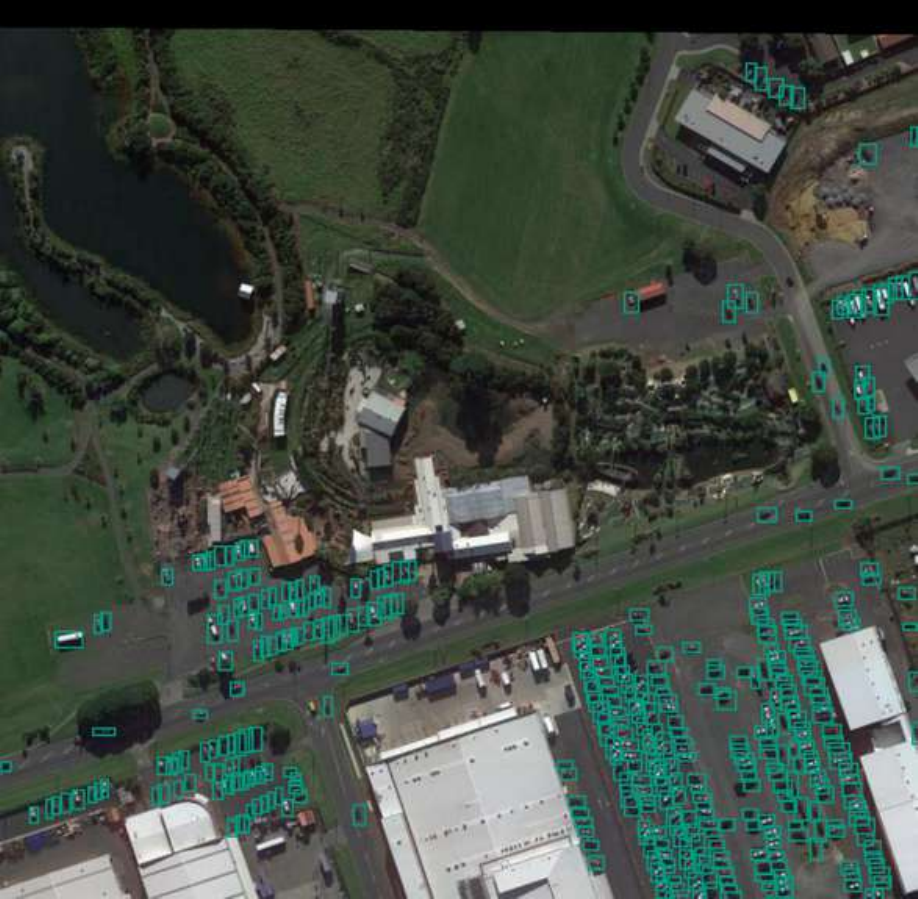} & 
        \includegraphics[width=\imgwidth]{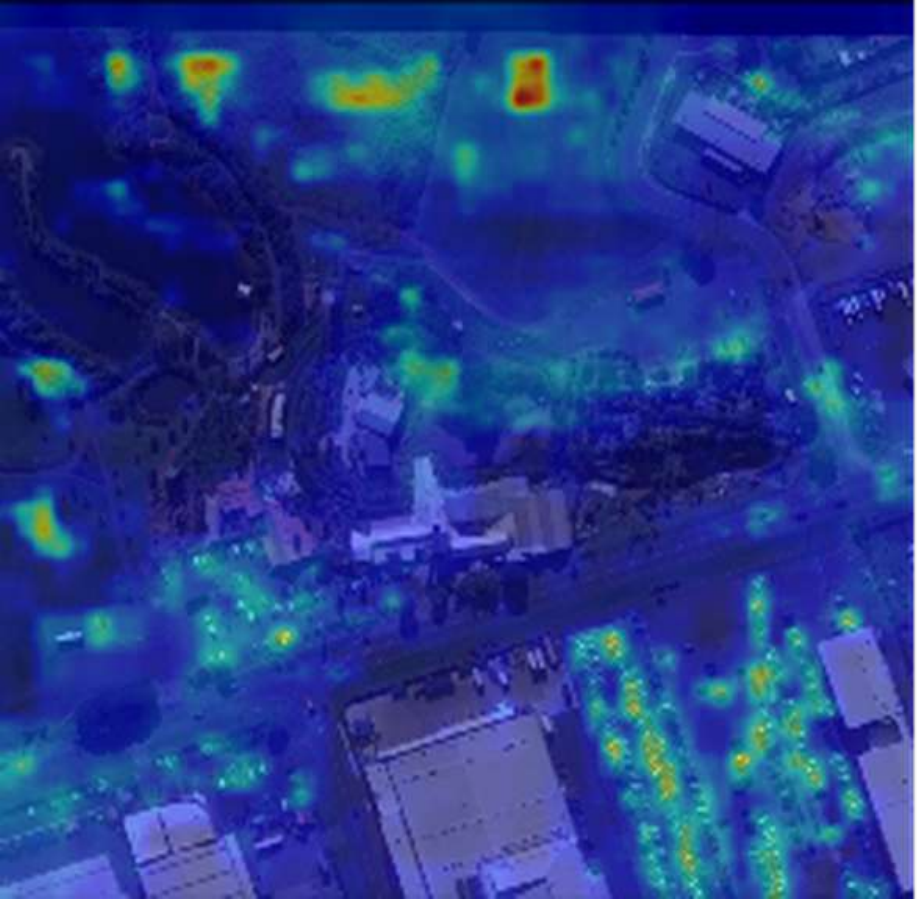} & 
        \includegraphics[width=\imgwidth]{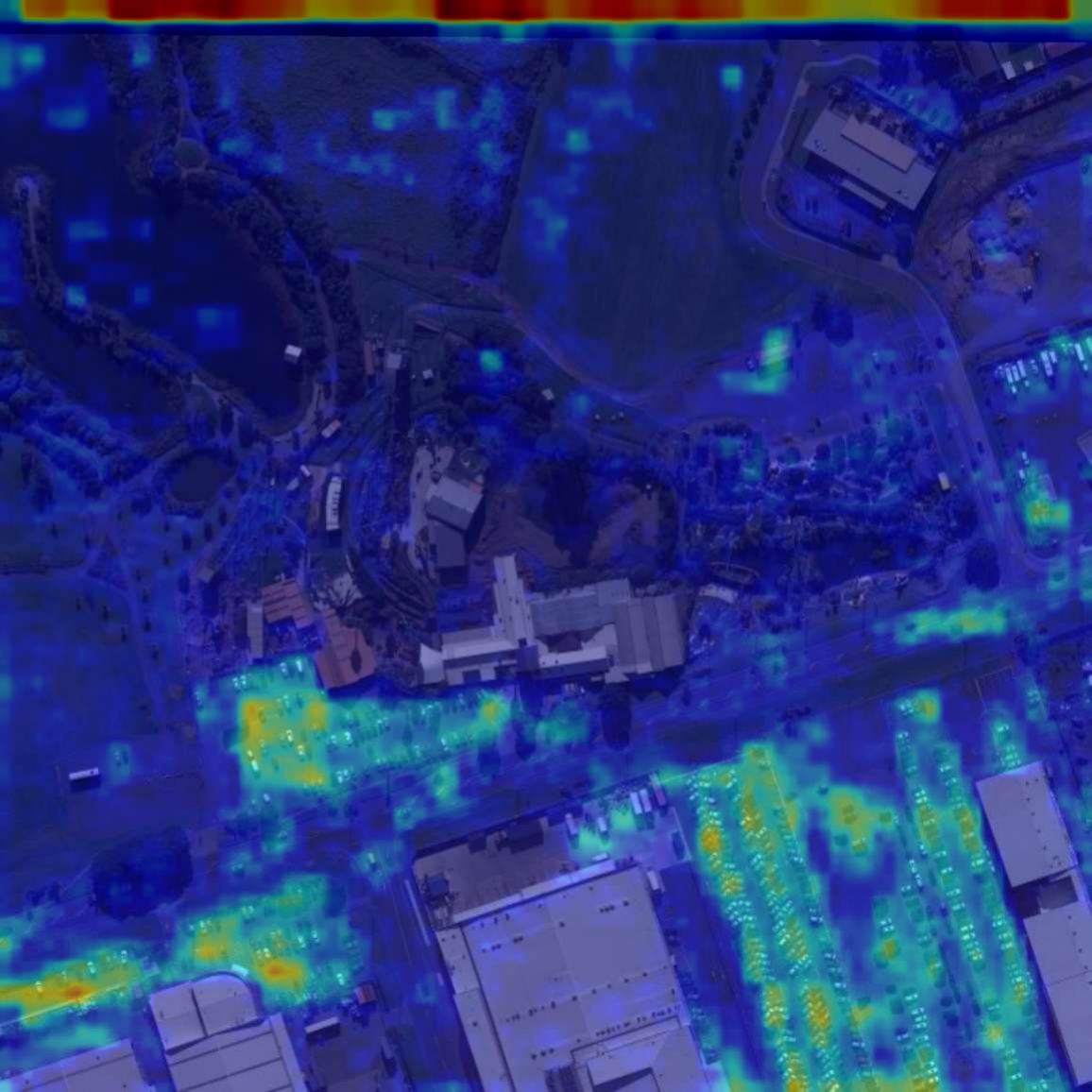} &
        \includegraphics[width=\imgwidth]{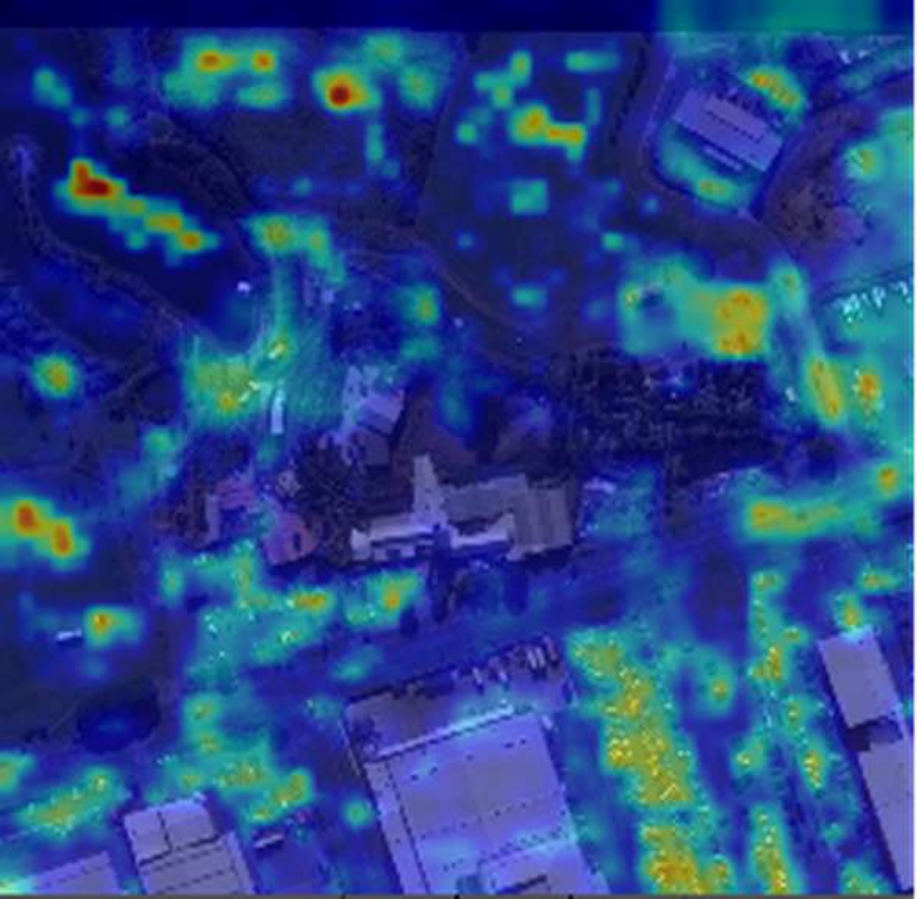} & 
        \includegraphics[width=\imgwidth]{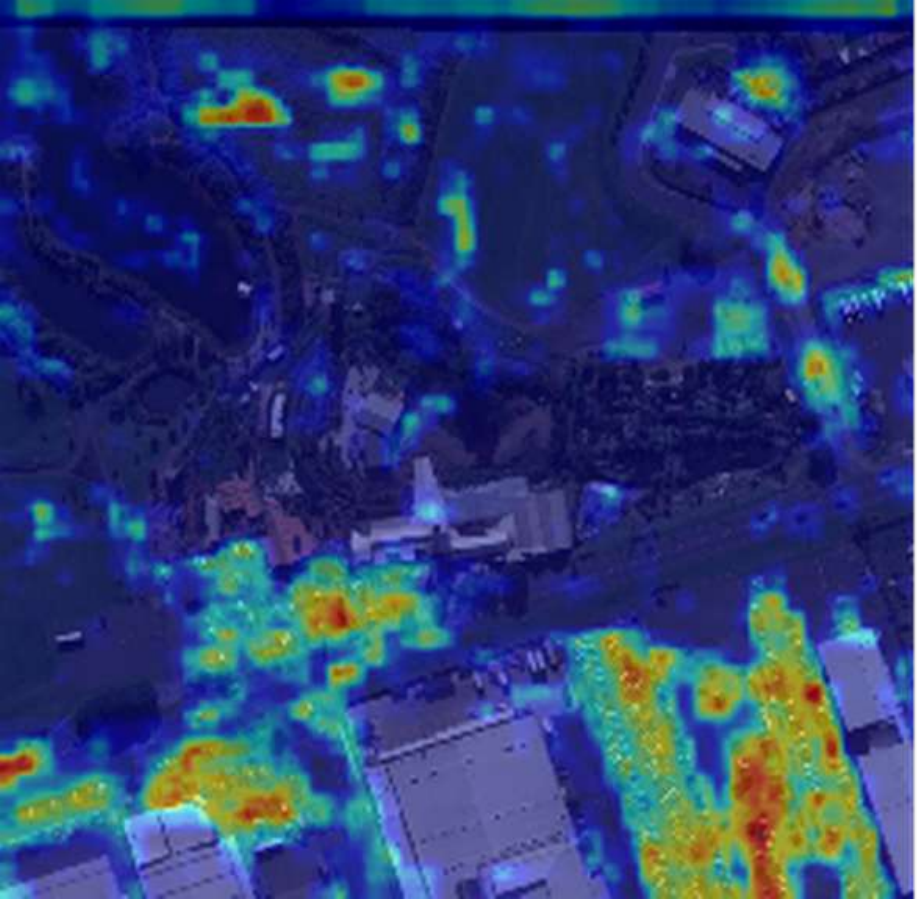} \\[-2pt]
        
        \includegraphics[width=\imgwidth]{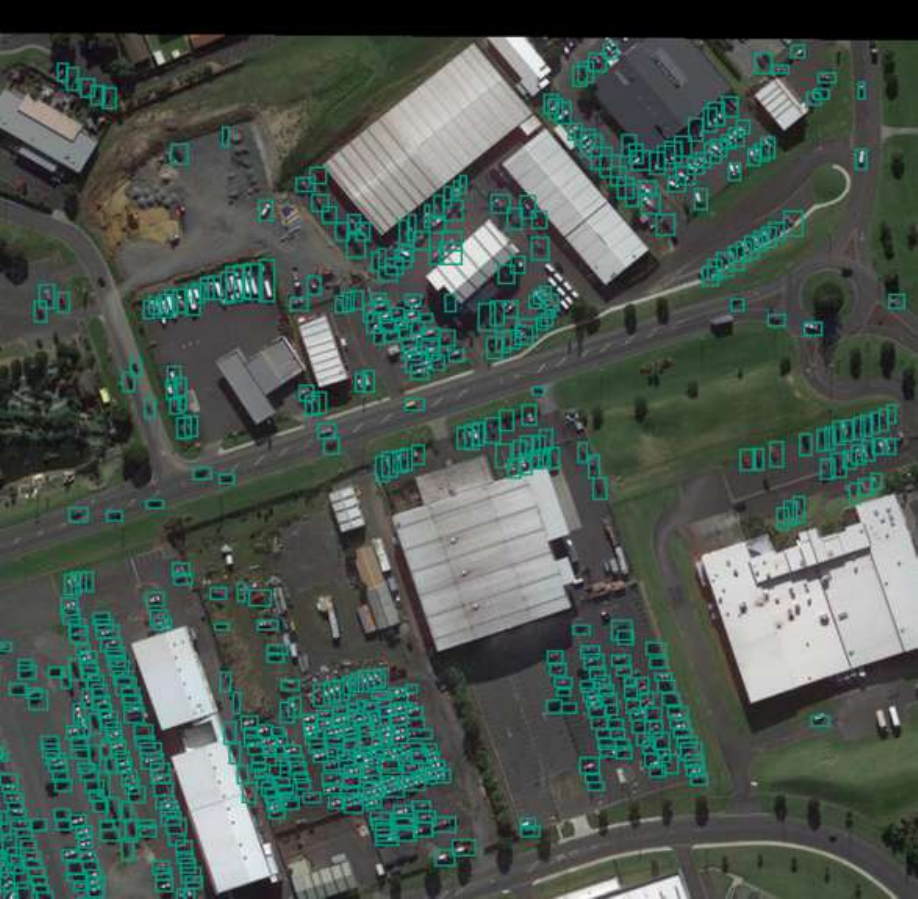} & 
        \includegraphics[width=\imgwidth]{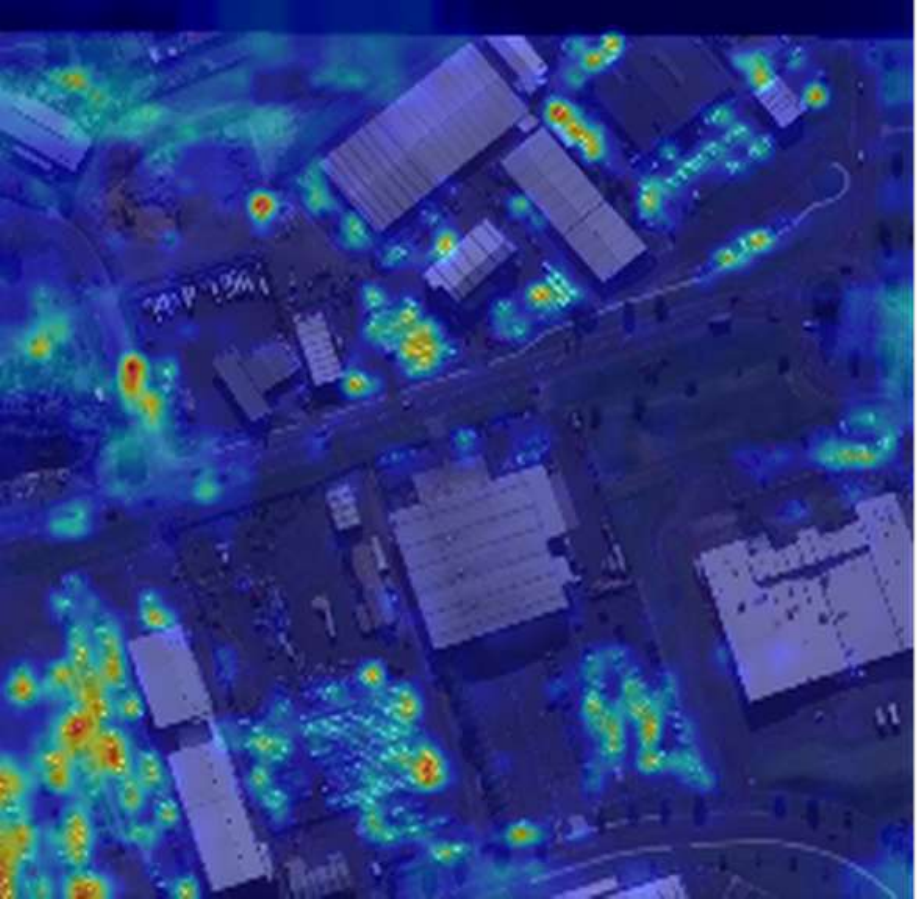} & 
        \includegraphics[width=\imgwidth]{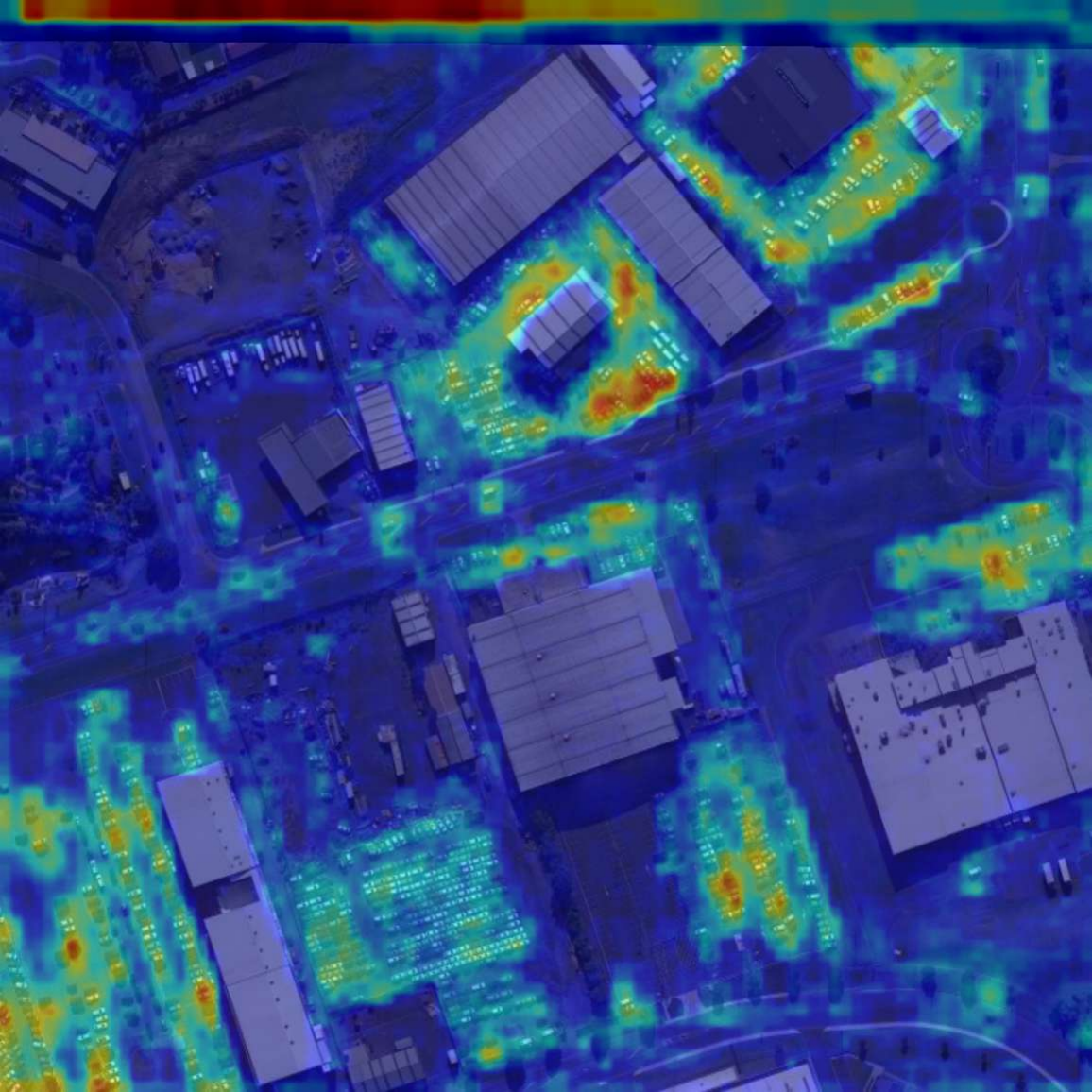} &
        \includegraphics[width=\imgwidth]{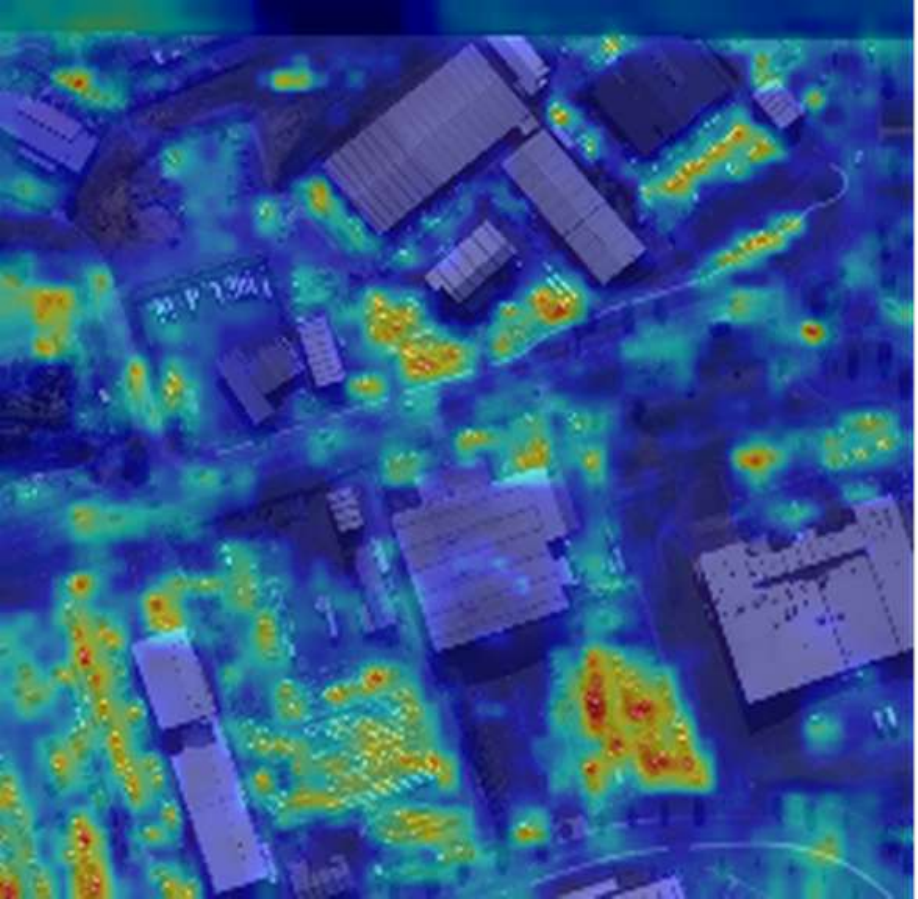} & 
        \includegraphics[width=\imgwidth]{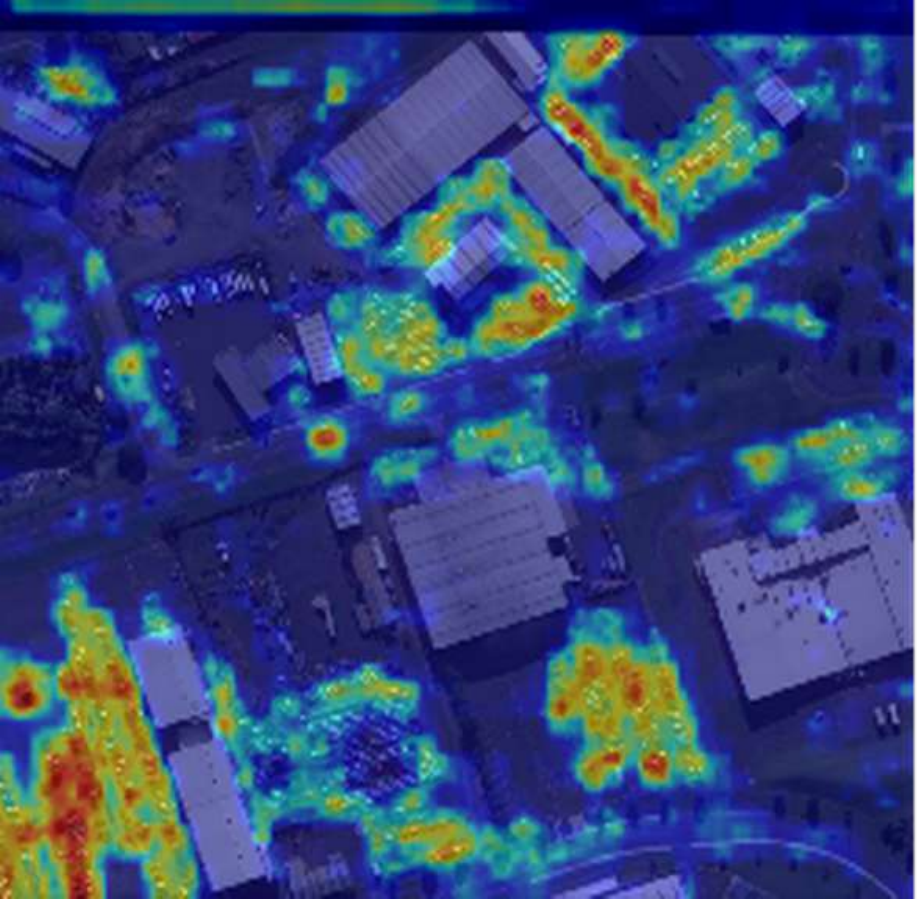} \\[-2pt]
        
        \includegraphics[width=\imgwidth]{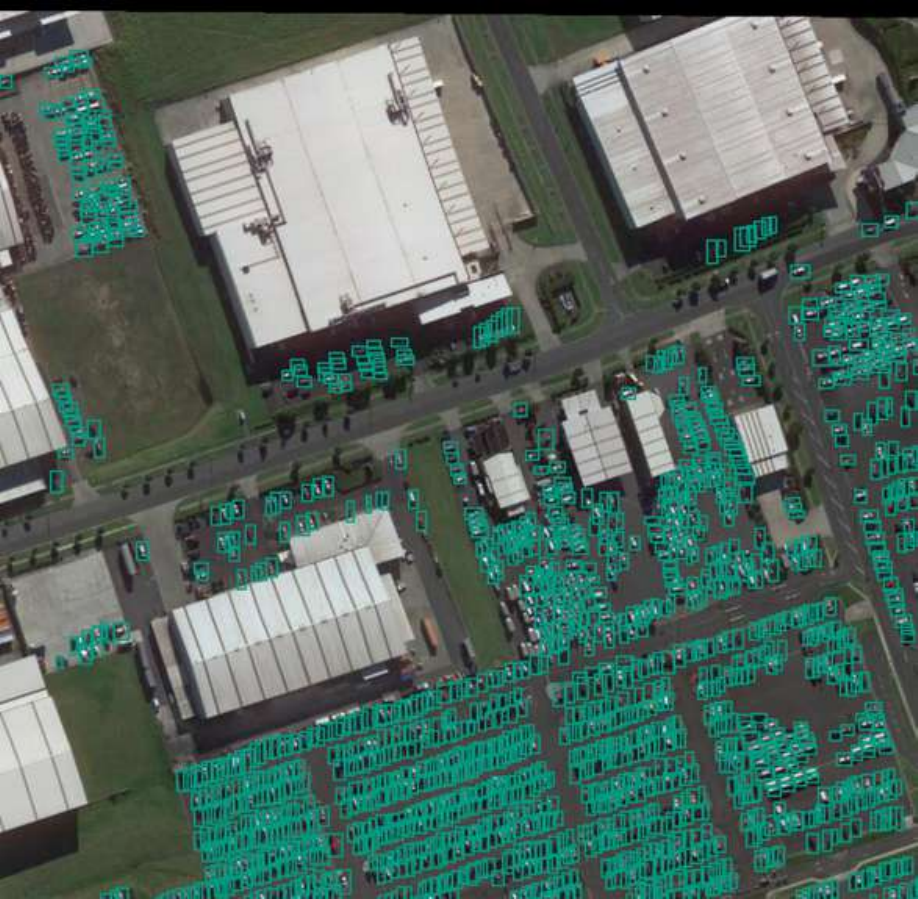} & 
        \includegraphics[width=\imgwidth]{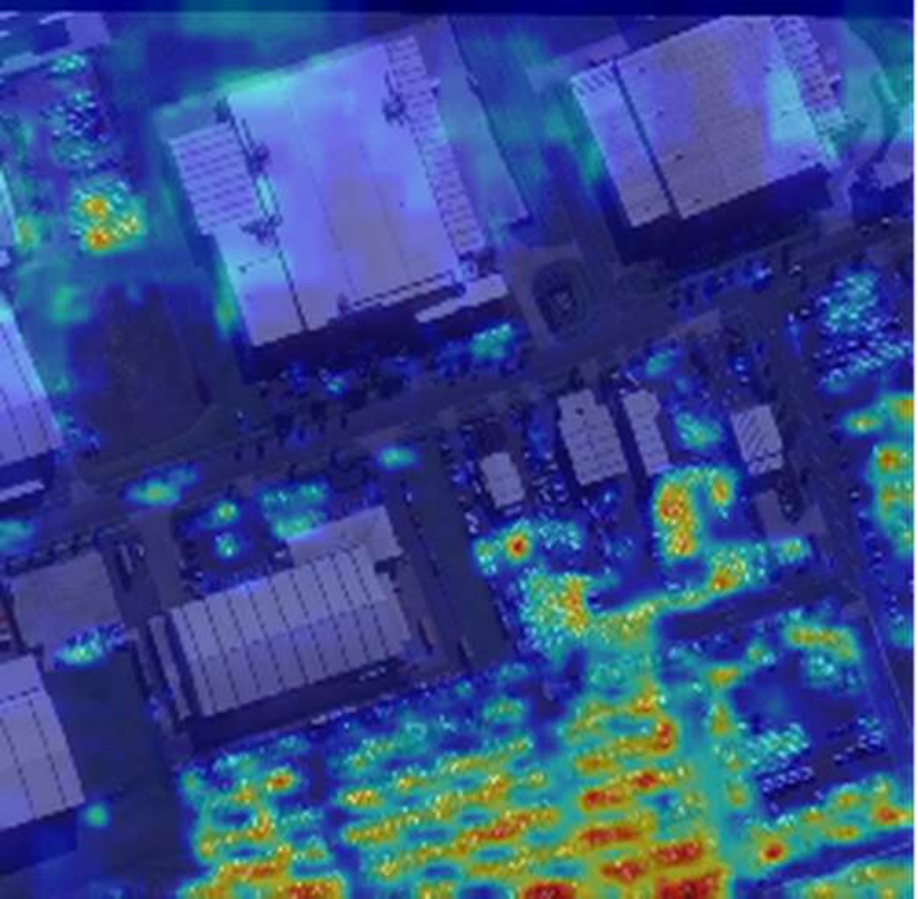} & 
        \includegraphics[width=\imgwidth]{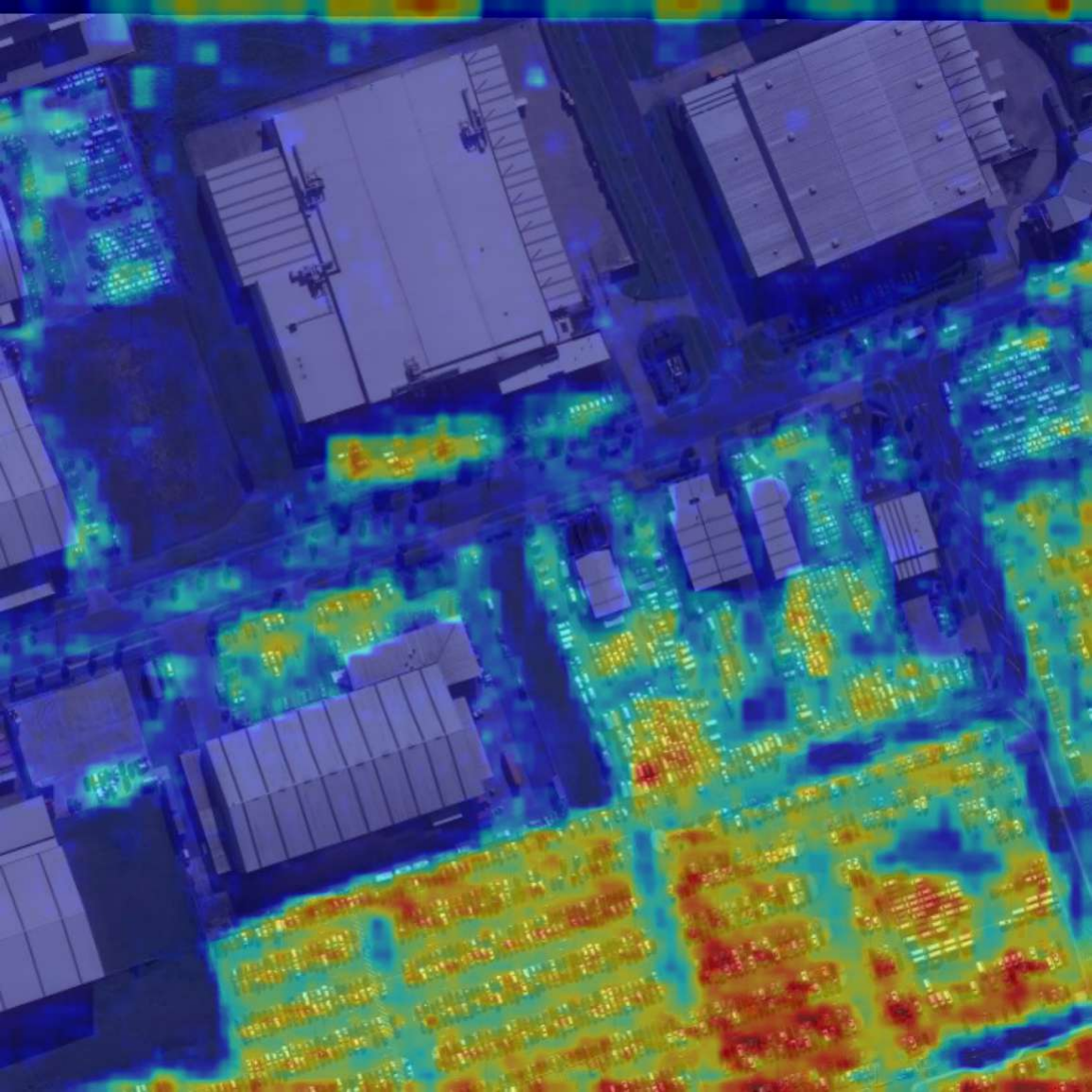} &
        \includegraphics[width=\imgwidth]{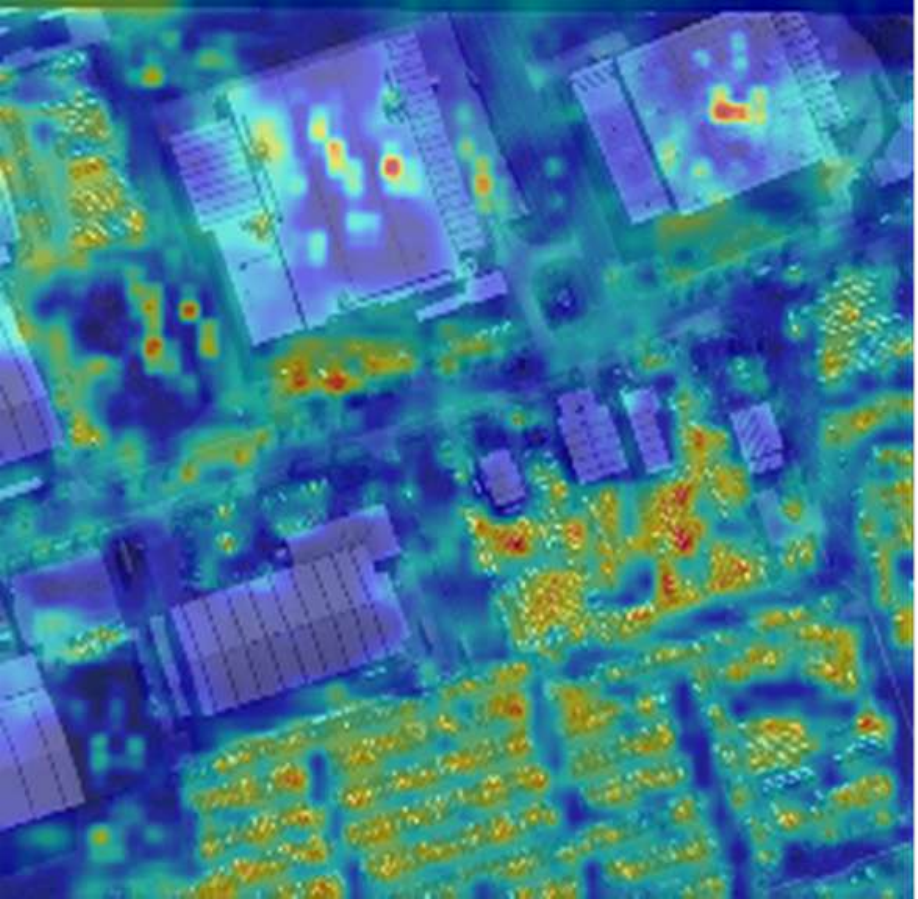} & 
        \includegraphics[width=\imgwidth]{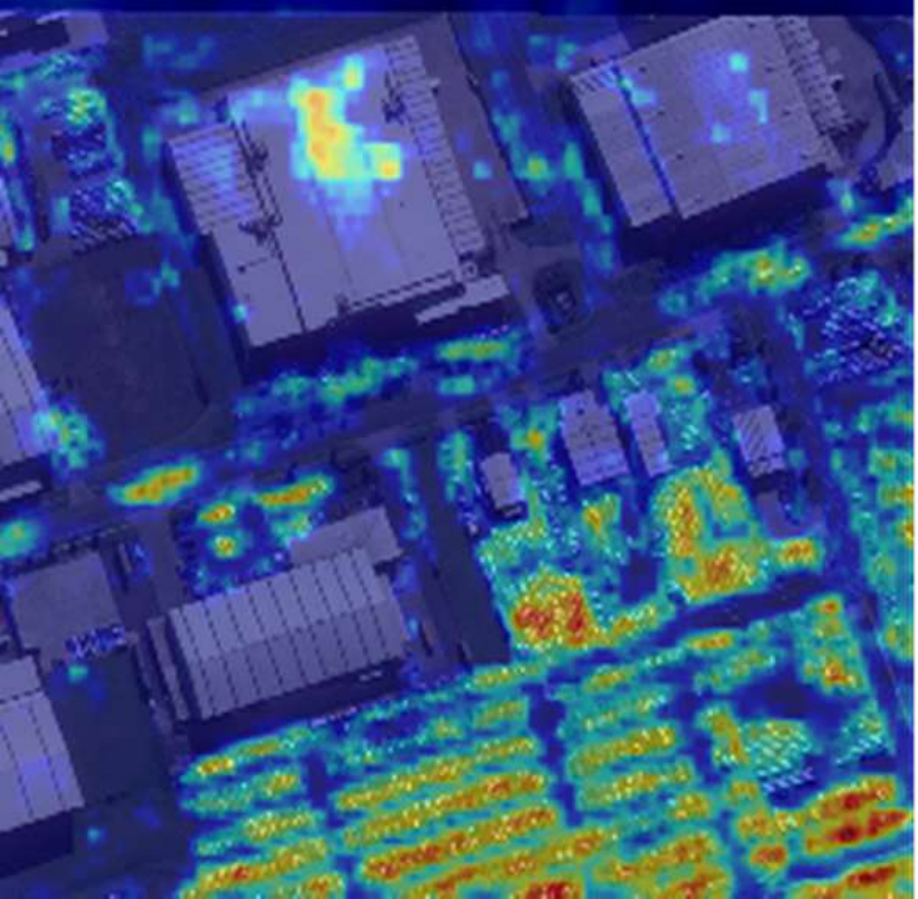} \\[-2pt]
        
        \small Input & \small YOLOv8 & \small RT-DETR & \small YOLOV11 & \small DRMNet(Ours) \\ 
    \end{tabular}
    \setlength{\tabcolsep}{6pt} 
    \vspace{2pt} 
    \caption{\small Visualize learning attention maps using GradCAM. The comparison of results from different methods shows that our approach can focus on object locations better and is less affected by background noise. The density map further enhances the model's ability to focus on object cluster areas.}
    \label{gram}
\end{figure*}

\subsubsection{Effectiveness of the DFFM}
We then assessed the DFFM module, which leverages the density map to guide feature fusion during the detection stage. When integrated alone, DFFM significantly improves performance over the baseline, boosting $AP_{50}$ by 2.1\% and $AP_{75}$ by 1.5\%. Notably, its impact is particularly pronounced on tiny objects, with $AP_{vt}$ and $AP_t$ increasing by 0.9\% and 2.8\%, respectively. These results indicate that DFFM effectively translates density guidance into more discriminative fused features, which helps the model generate predictions with higher quality and clearer boundaries.
Finally, we evaluated the synergy between DAFM and DFFM. Combining these two modules yields the best overall performance, improving $AP_{50}$ and $AP_{75}$ by 2.9\% and 2.1\%, respectively, compared to the baseline model. This confirms the high complementarity of these two modules: DAFM first refines region-specific features using density priors, then DFFM intelligently fuses foreground features at different scales. Their joint operation is crucial for maximizing the model's ability to perceive and detect dense, small objects.

\subsubsection{Main Detection Network}
The selection of a backbone network suitable for tiny object detection is crucial. We chose the YOLO architecture as a foundation due to its inherent advantages, such as cross-layer feature fusion and anchor-free dense sampling, which preserve high-resolution spatial information essential for detecting objects smaller than 8×8 pixels. To identify the optimal version for our task, we conducted a comparative experiment on several prominent YOLO variants, including the official YOLOv5 and YOLOv8, as well as other recent developments like YOLOv11 and YOLOv12. The results, visualized in the radar chart in Fig. \ref{fig:radar}, demonstrate that YOLOv8 (represented by the red line) consistently outperforms the other candidates across a range of detection metrics. Therefore, we selected YOLOv8 as the main detection network for our framework.

\renewcommand{\imgwidth}{3.5cm} 
\begin{figure*}[htbp]
    \centering
    \setlength{\tabcolsep}{2pt} 
    \begin{tabular}{>{\centering\arraybackslash}m{\imgwidth}
                    >{\centering\arraybackslash}m{\imgwidth}
                    >{\centering\arraybackslash}m{\imgwidth}
                    >{\centering\arraybackslash}m{\imgwidth}
                    >{\centering\arraybackslash}m{\imgwidth}} 
        \includegraphics[width=\imgwidth]{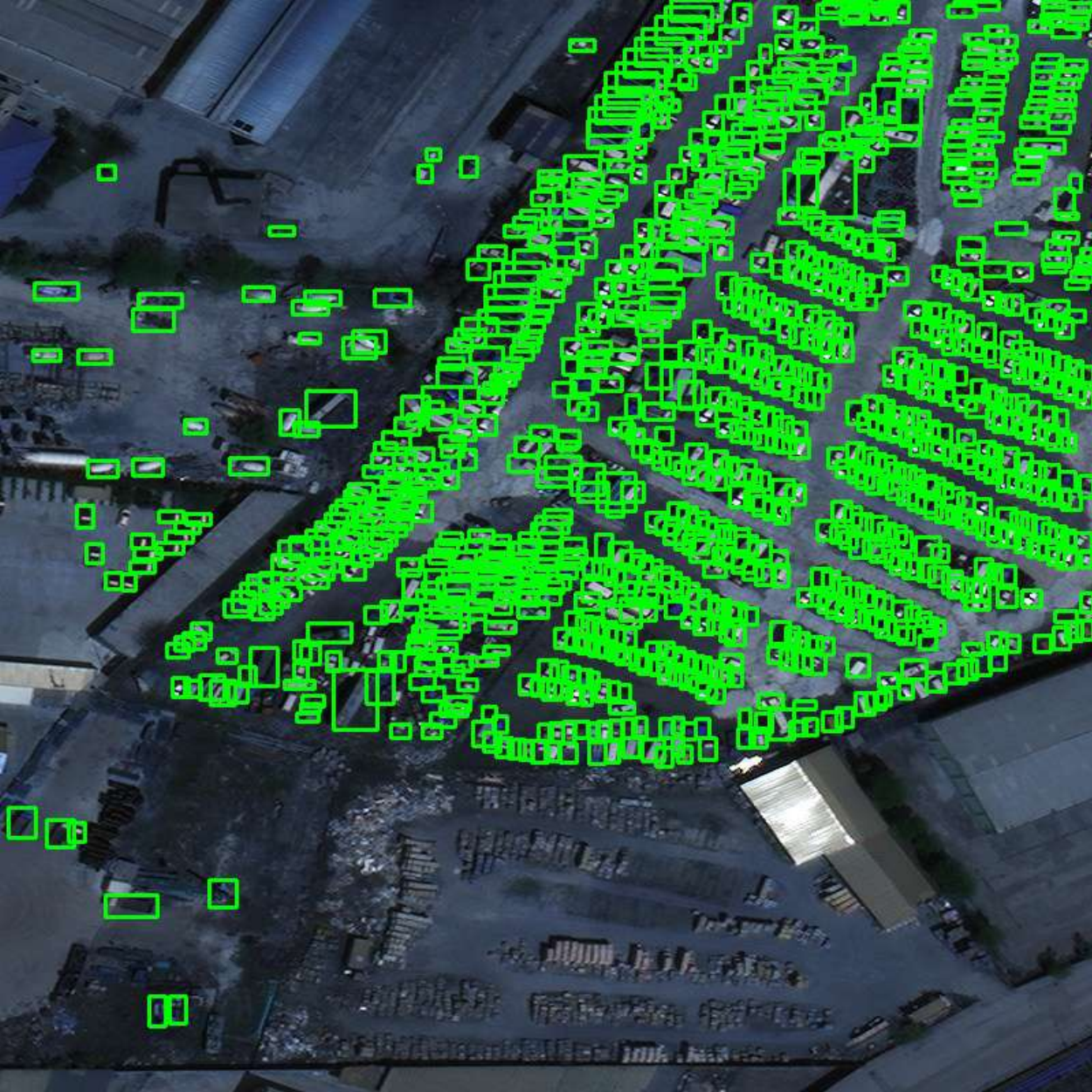} & 
        \includegraphics[width=\imgwidth]{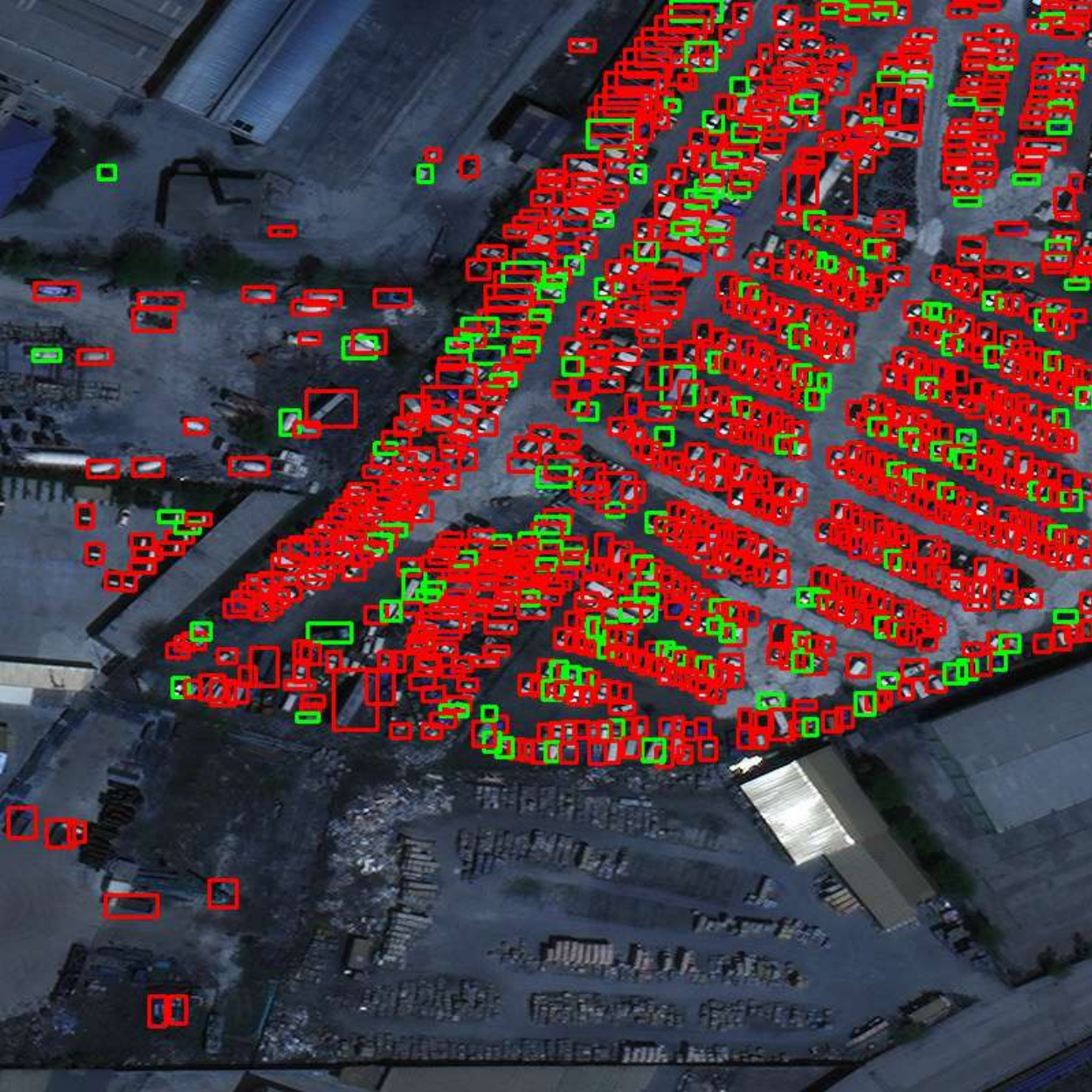} & 
        \includegraphics[width=\imgwidth]{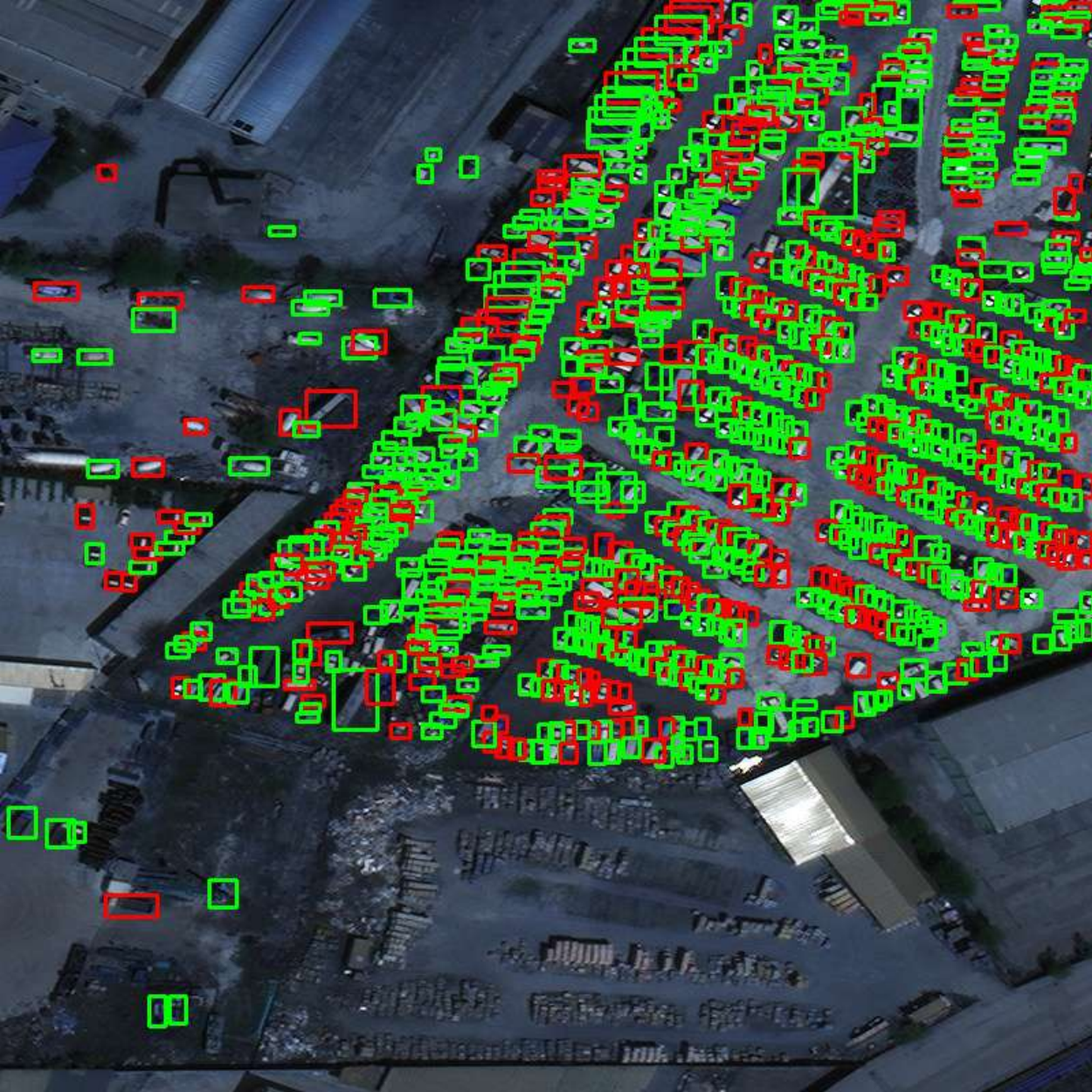} &
        \includegraphics[width=\imgwidth]{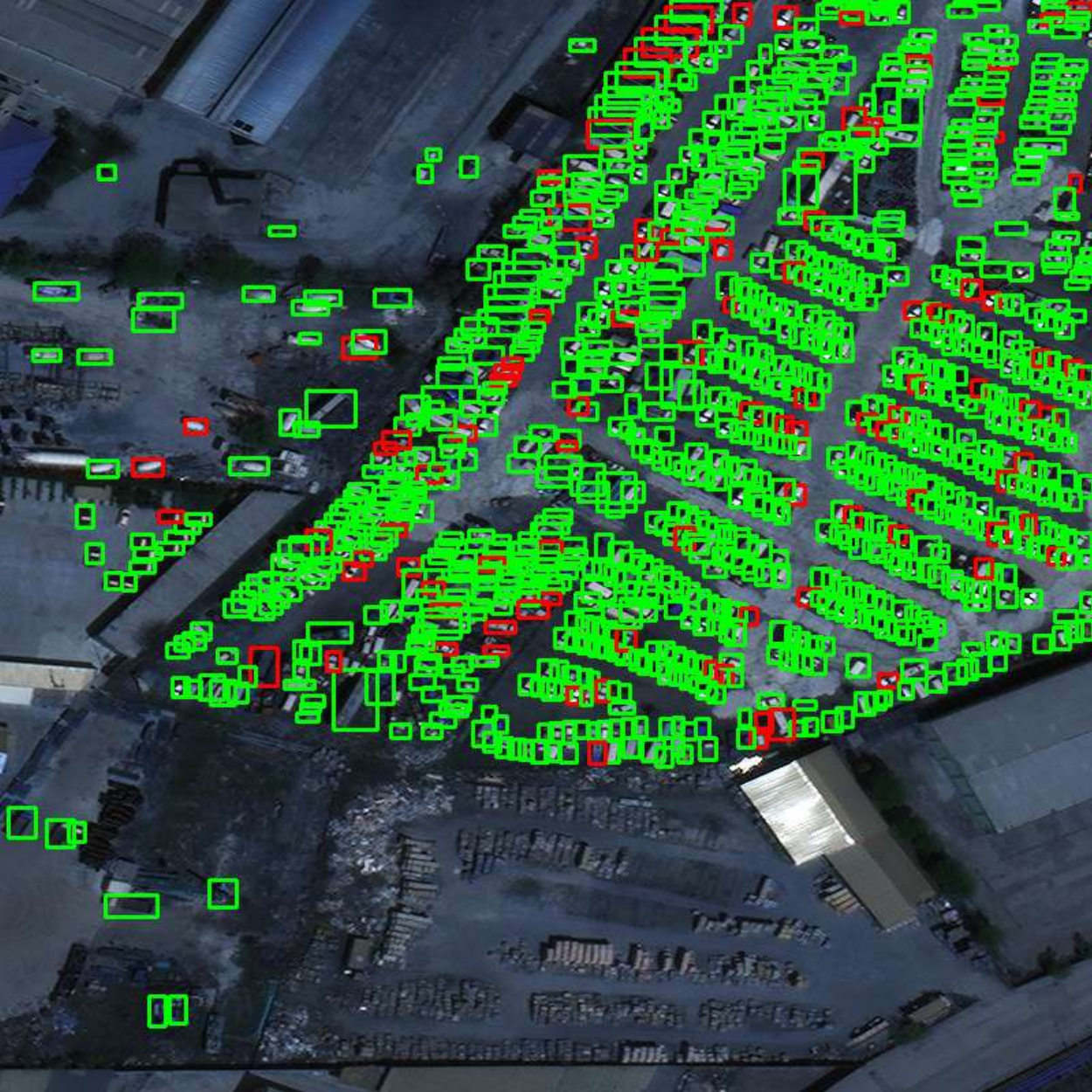} &
        \includegraphics[width=\imgwidth]{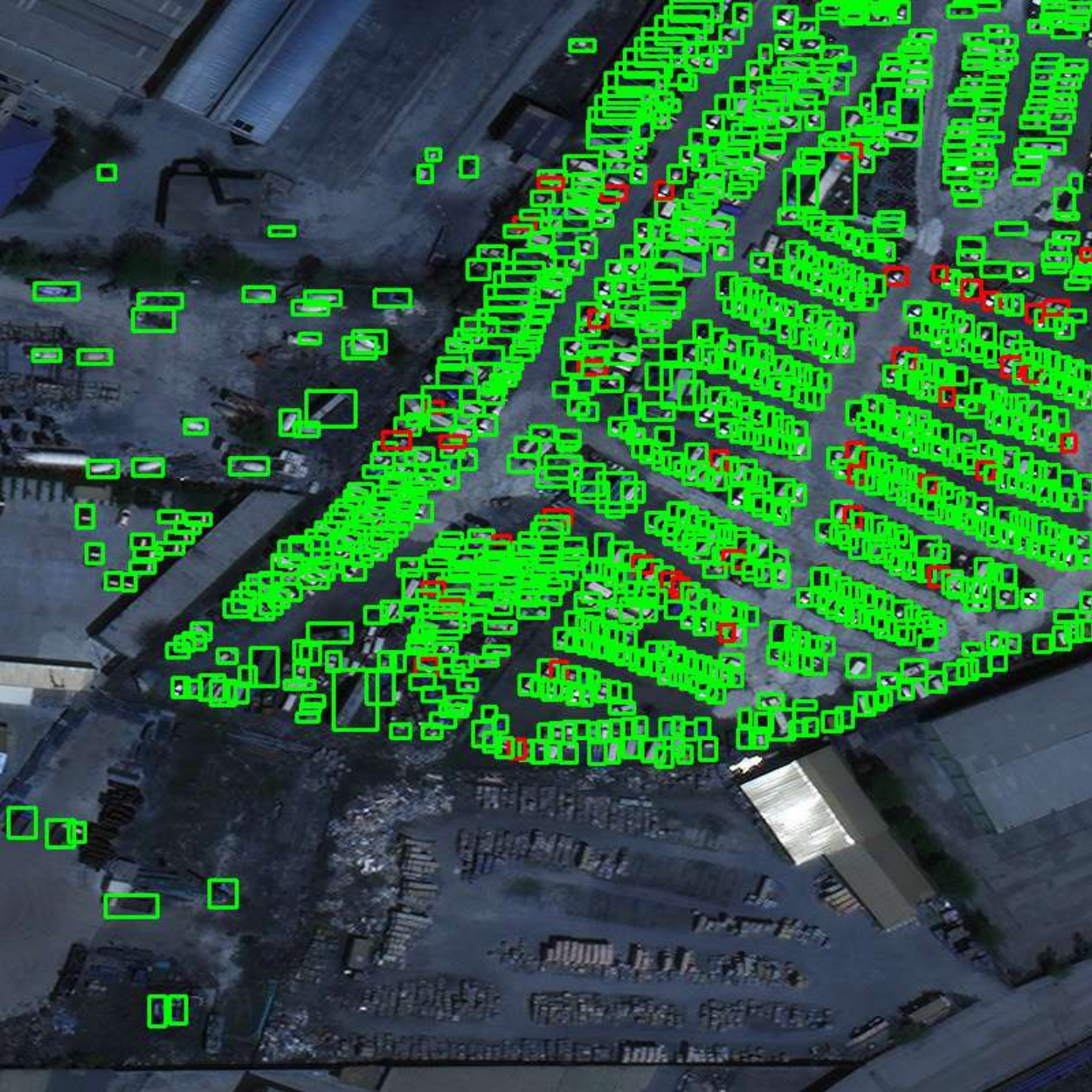} \\[-2pt] 
        
        \includegraphics[width=\imgwidth]{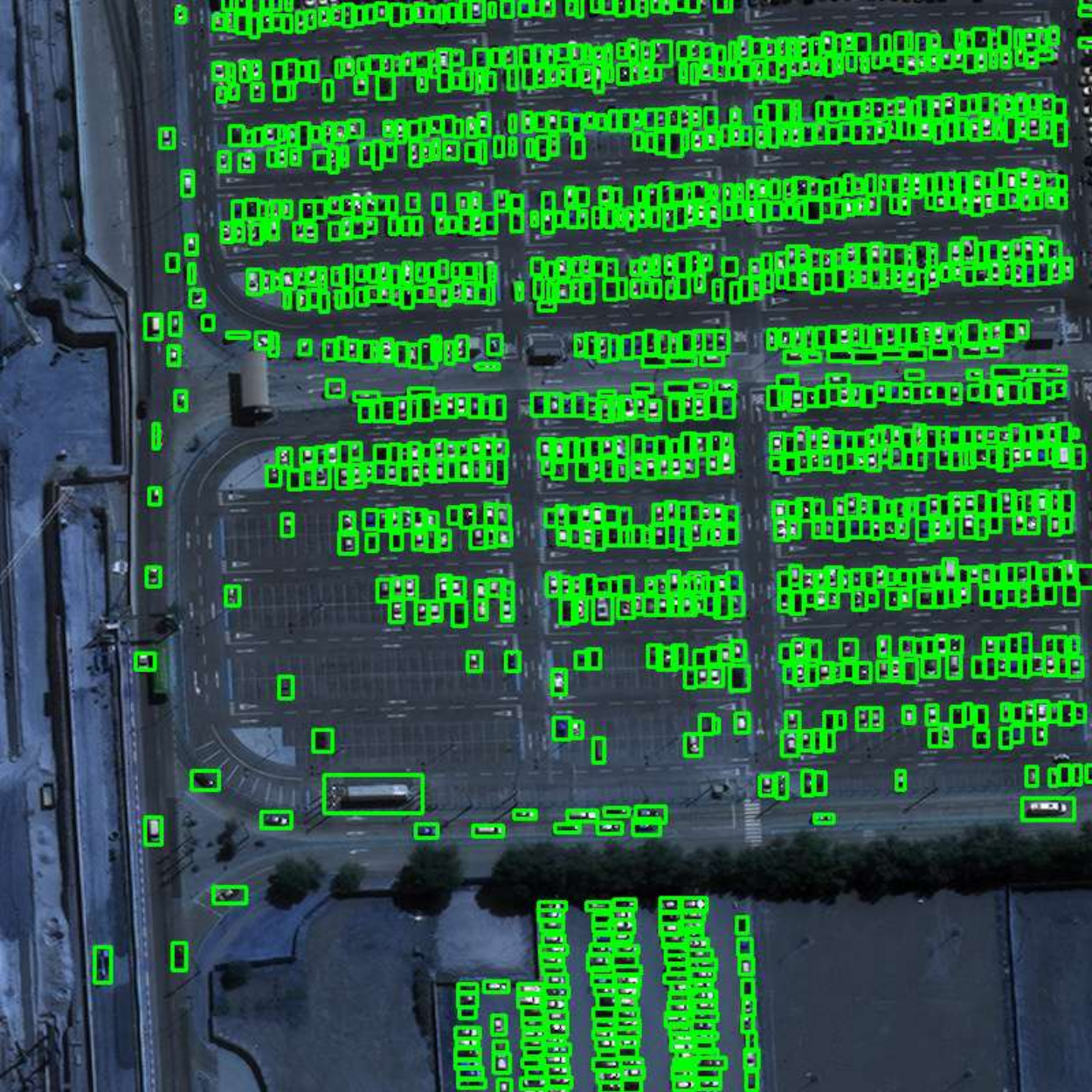} & 
        \includegraphics[width=\imgwidth]{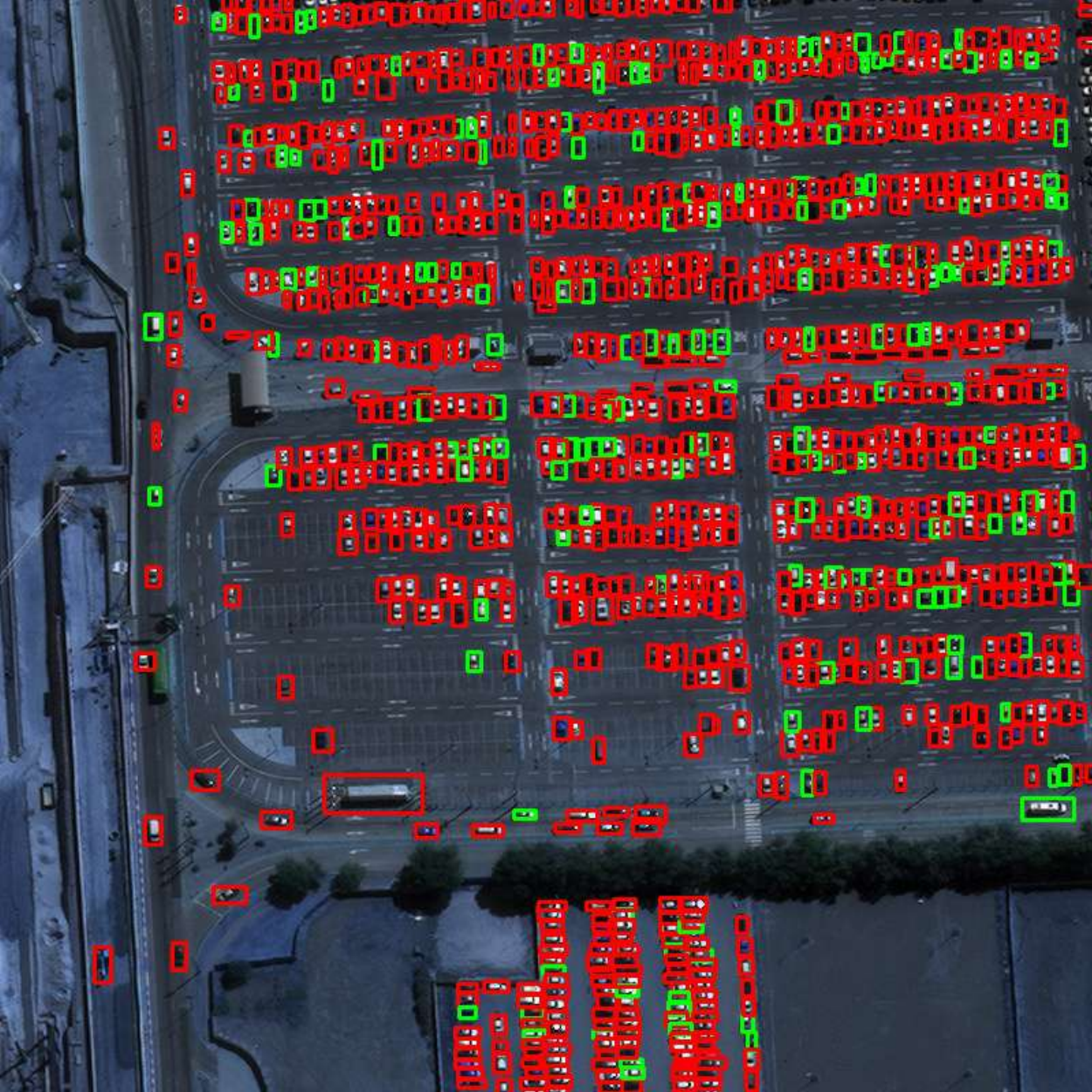} & 
        \includegraphics[width=\imgwidth]{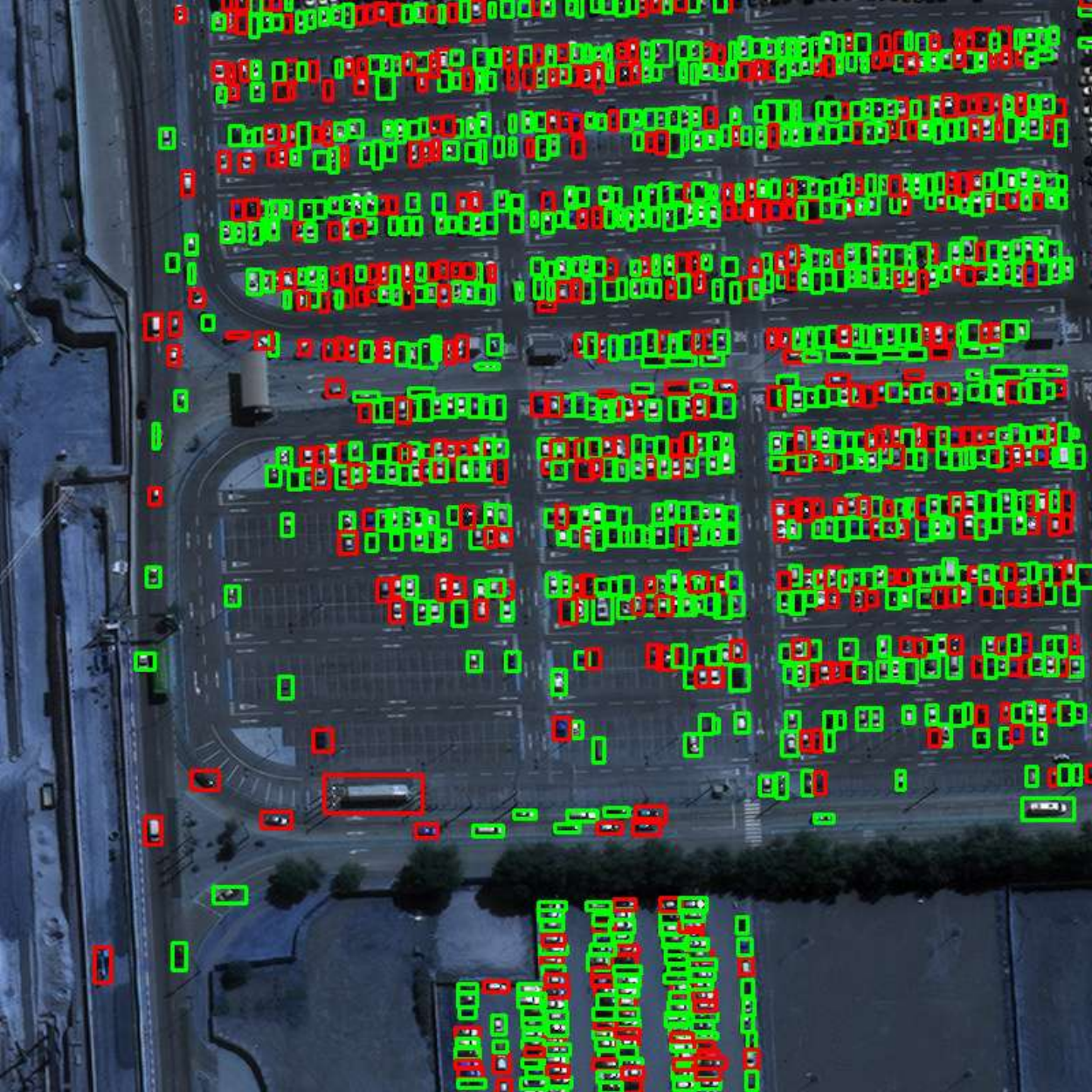} &
        \includegraphics[width=\imgwidth]{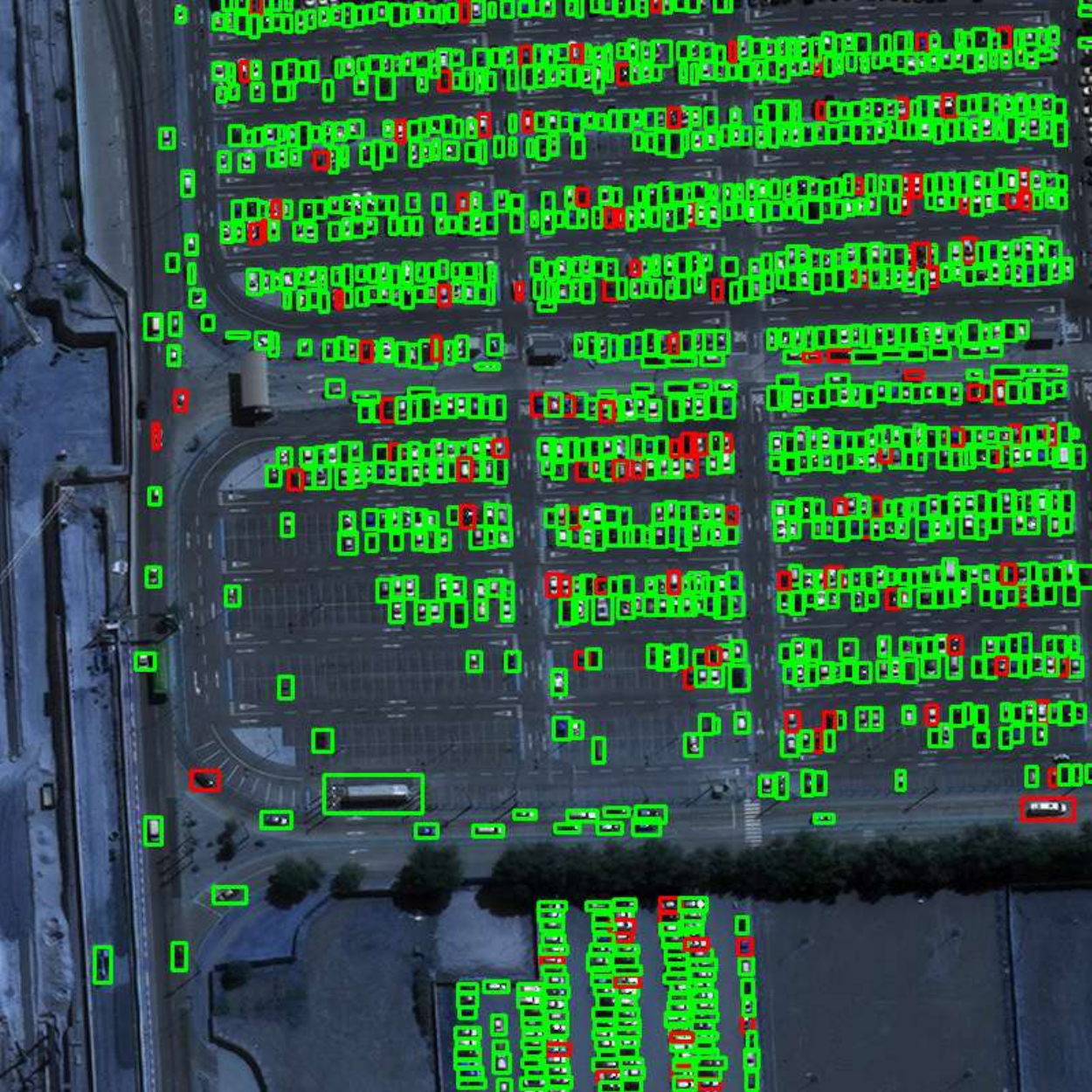} &
        \includegraphics[width=\imgwidth]{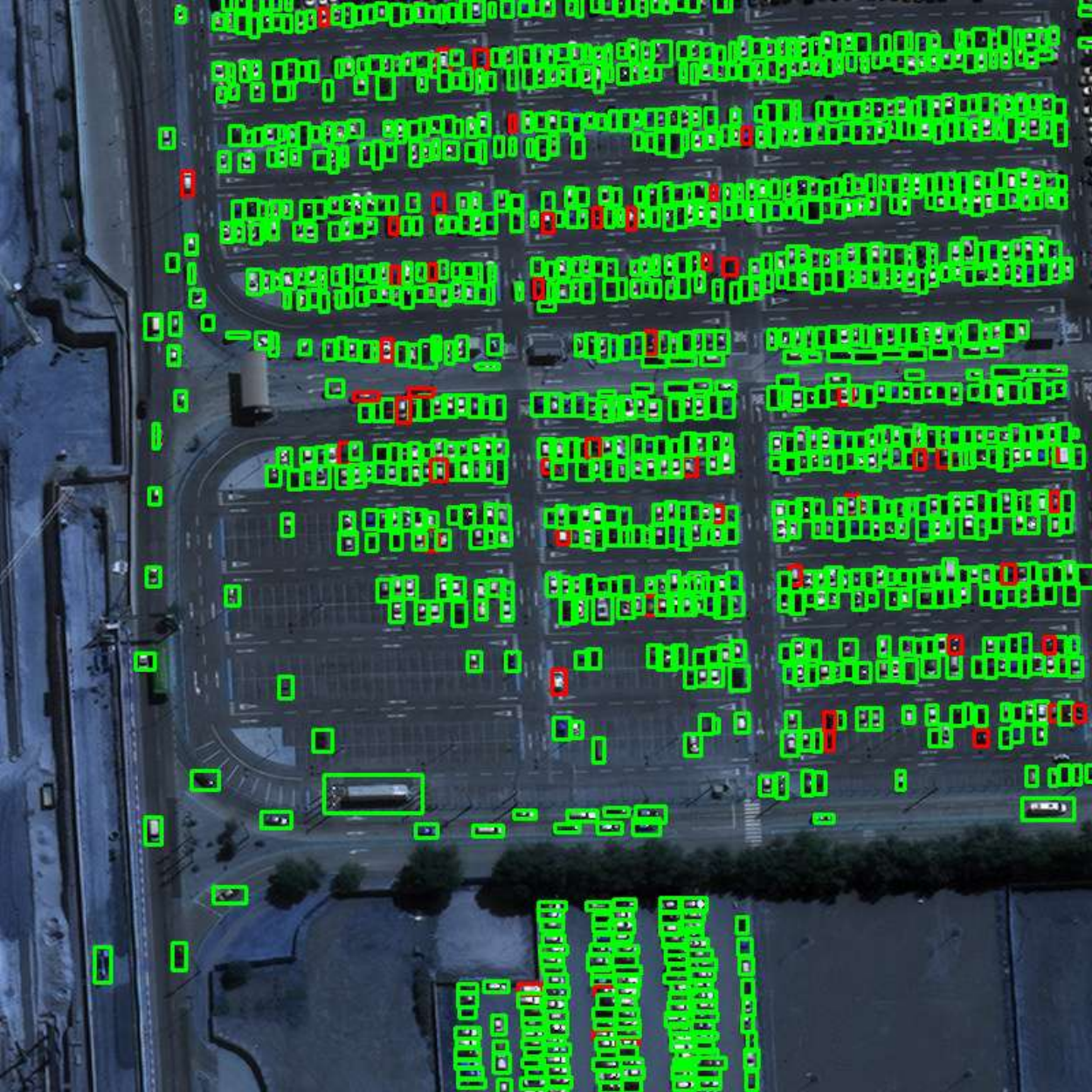} \\[-2pt]
        
        \includegraphics[width=\imgwidth]{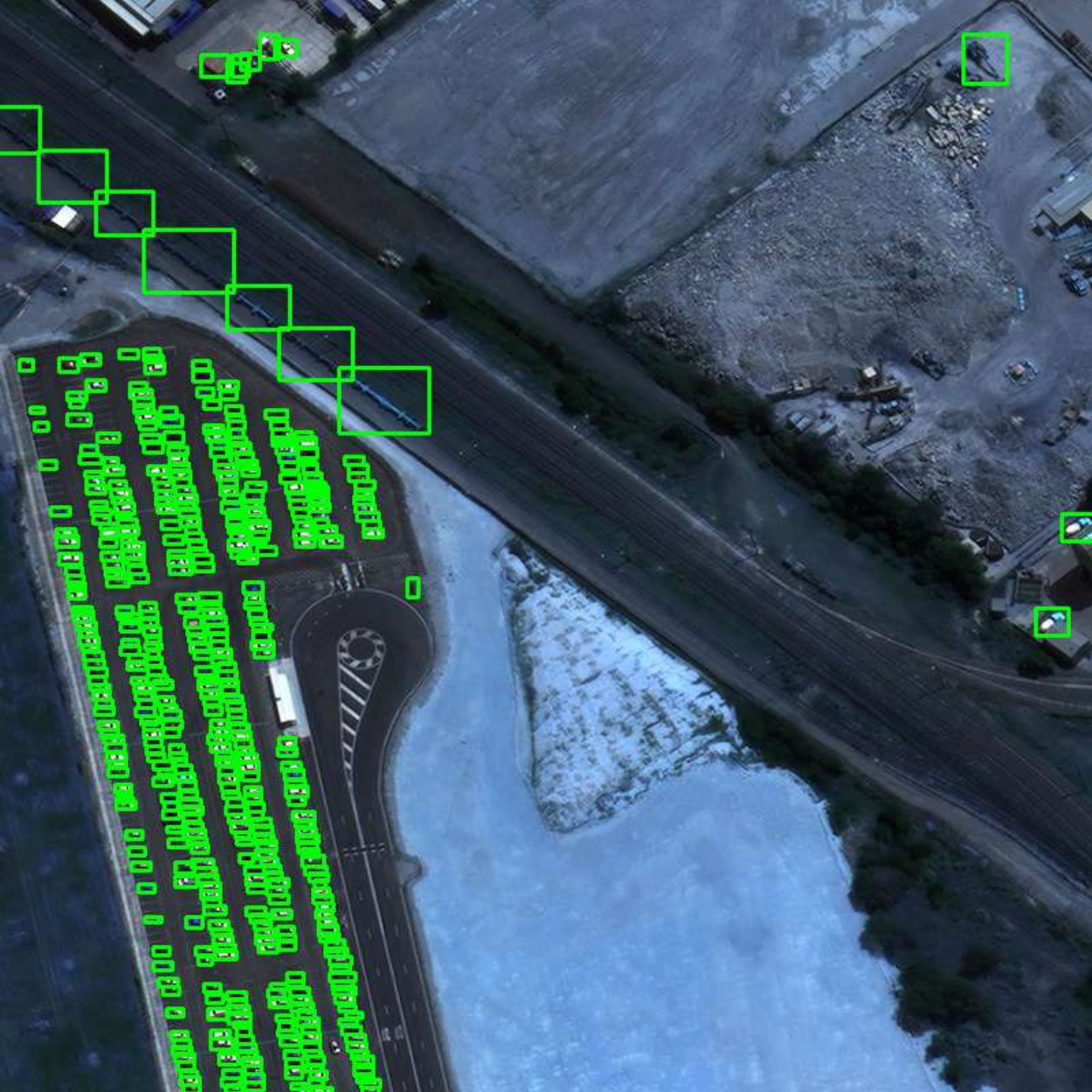} & 
        \includegraphics[width=\imgwidth]{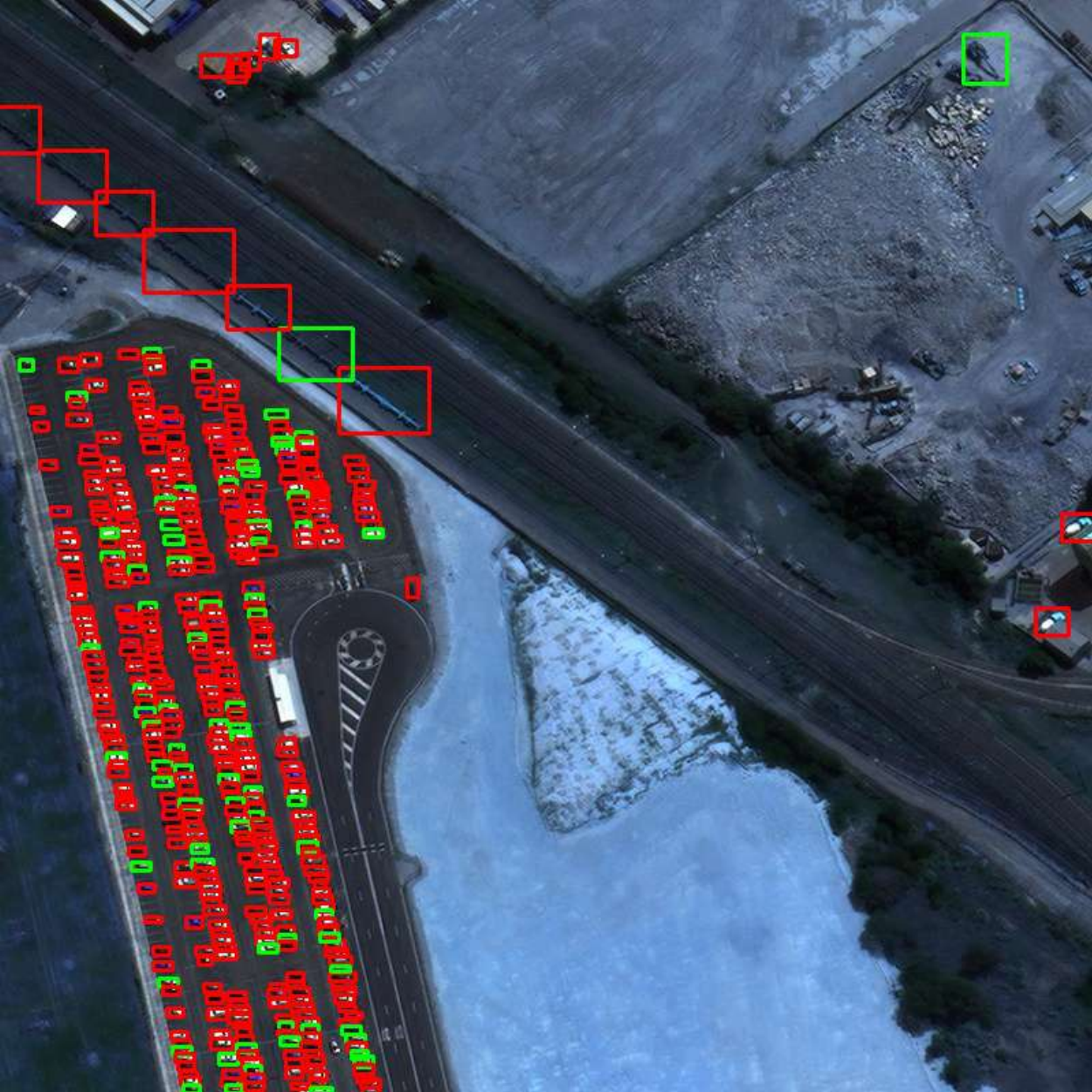} & 
        \includegraphics[width=\imgwidth]{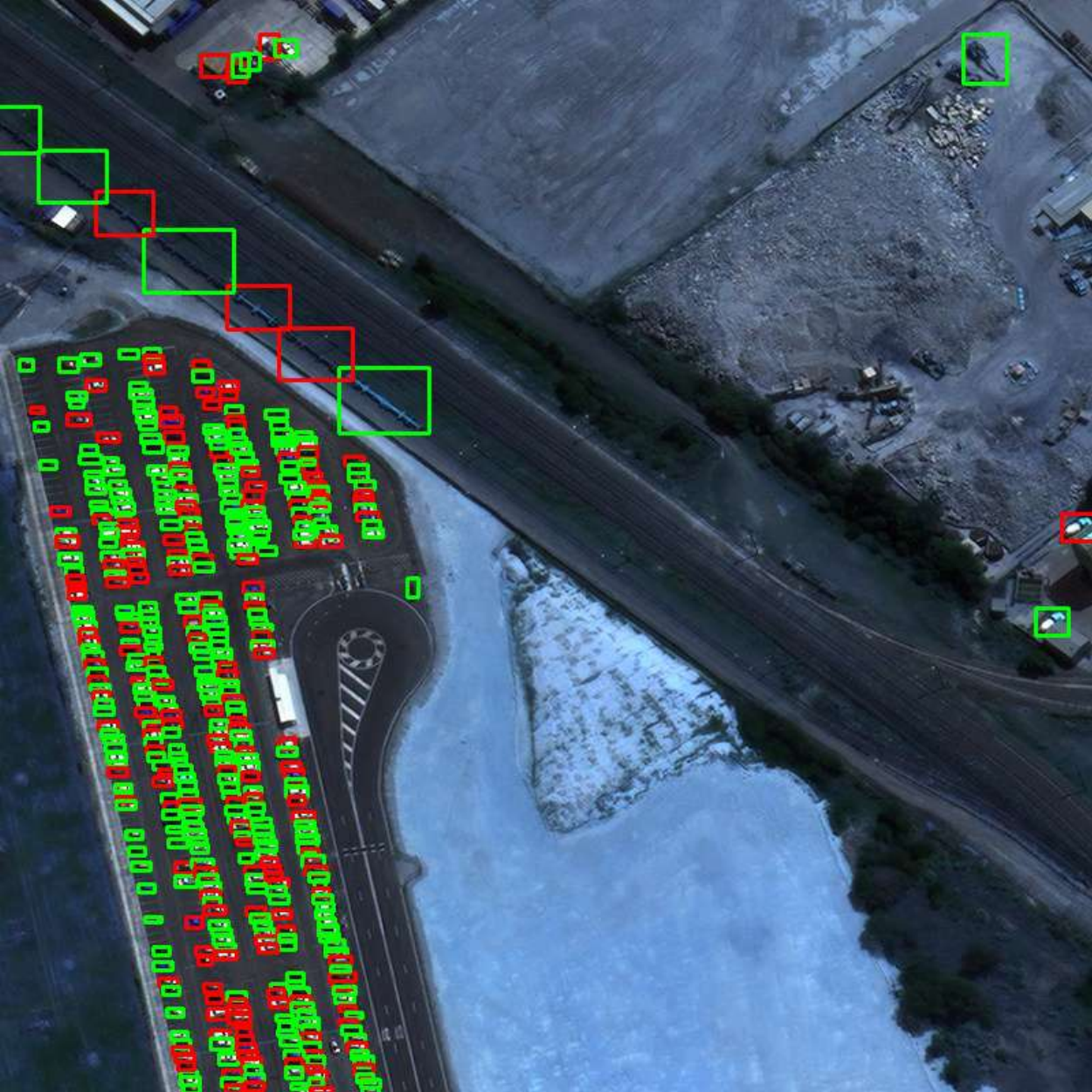} &
        \includegraphics[width=\imgwidth]{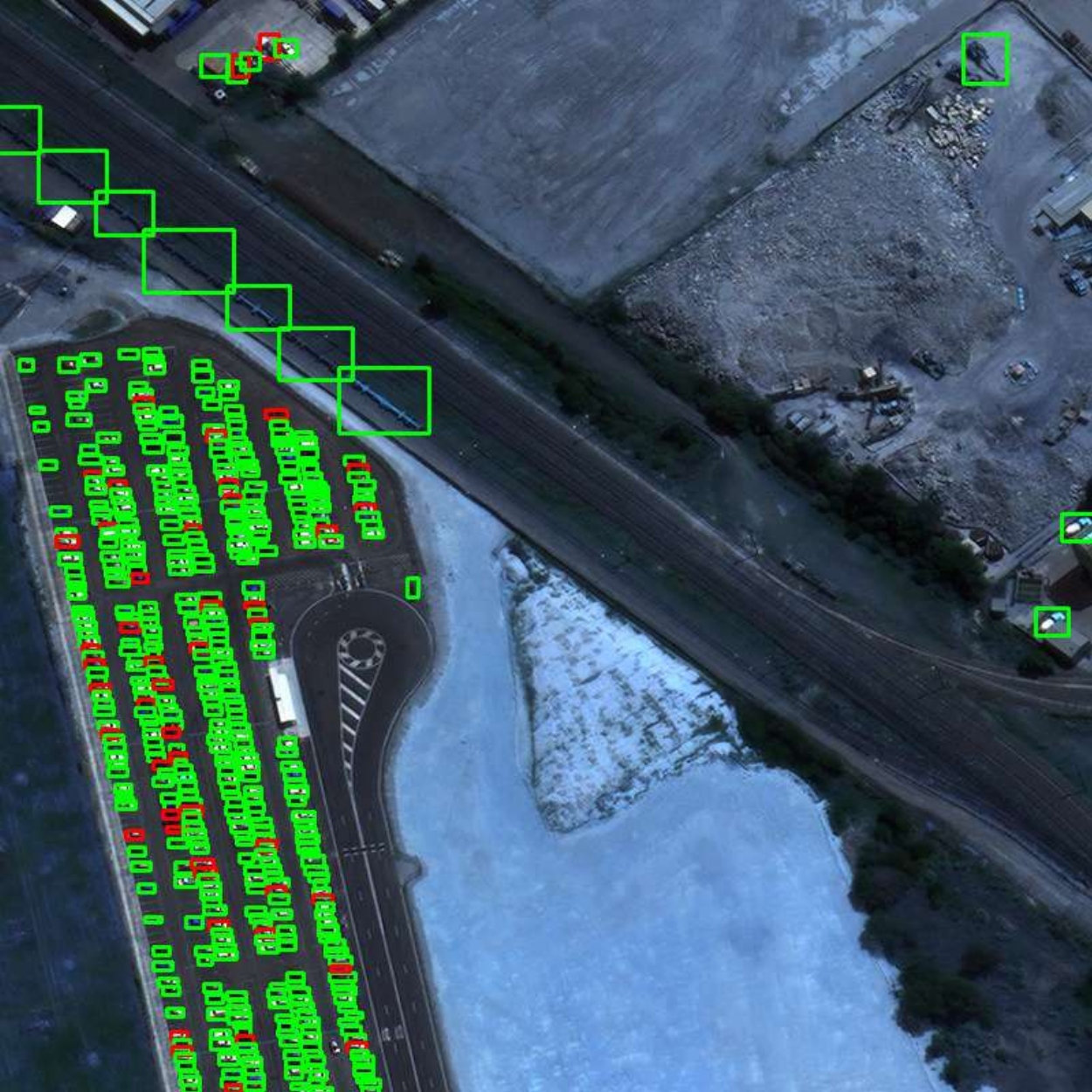} &
        \includegraphics[width=\imgwidth]{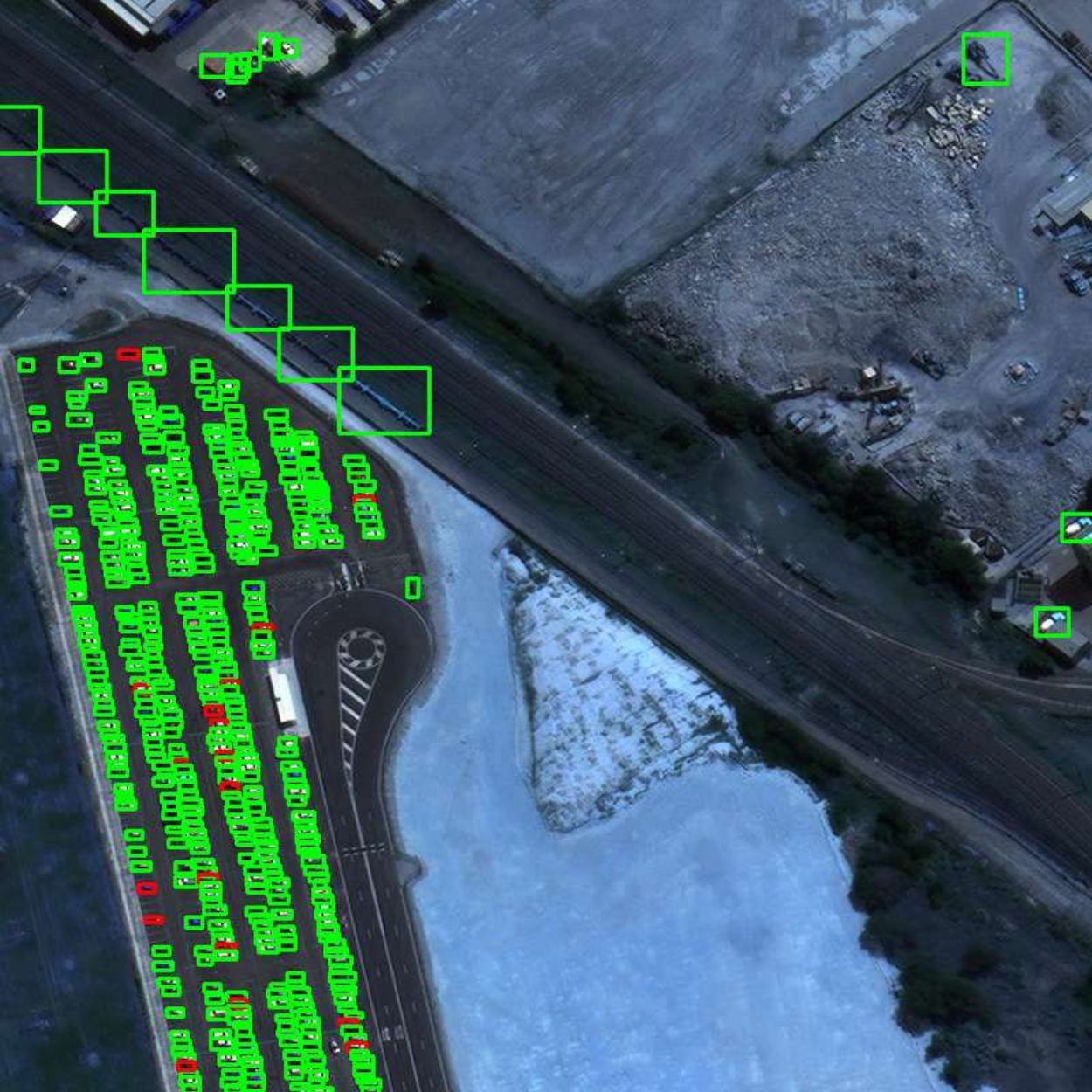} \\[-2pt]
        
        \includegraphics[width=\imgwidth]{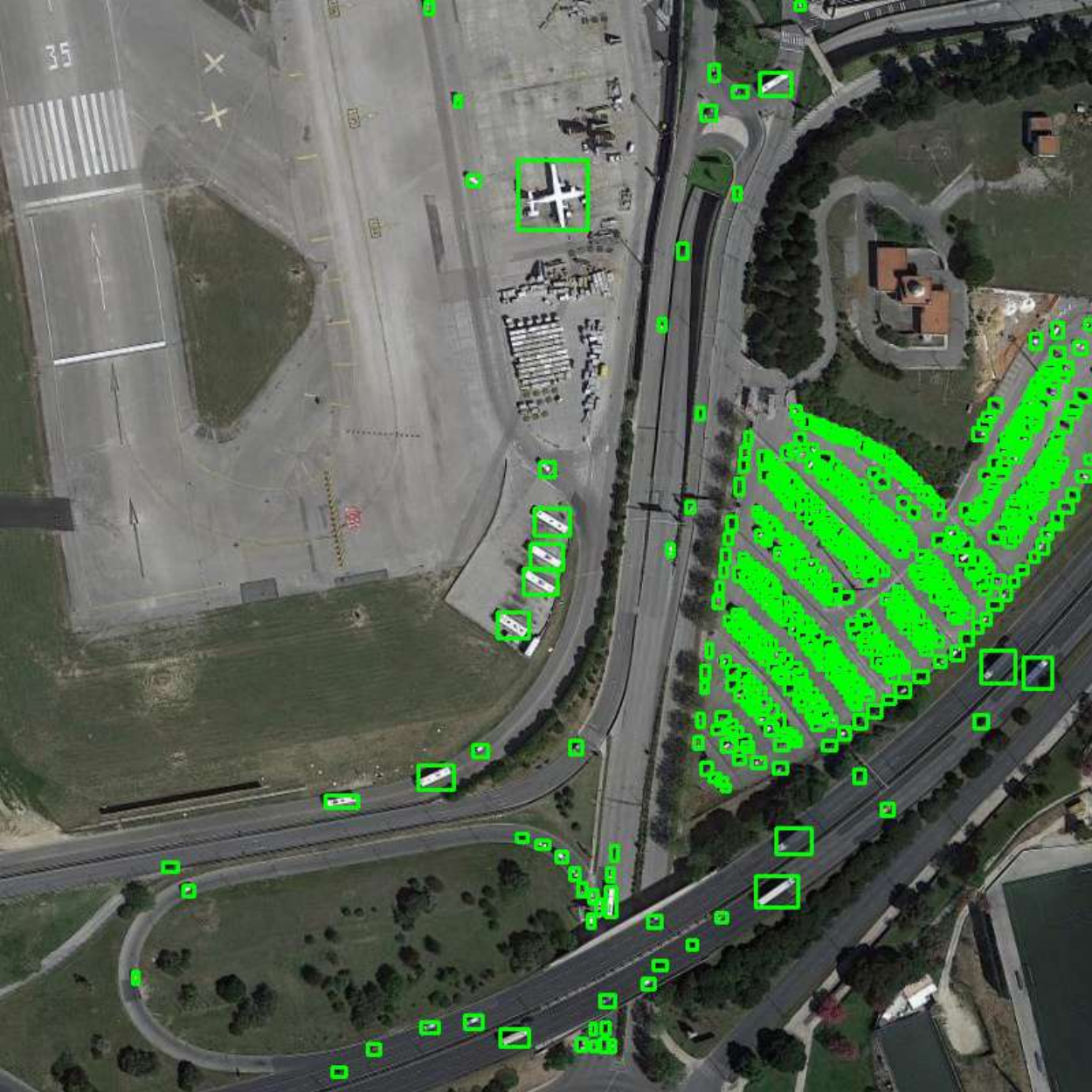} & 
        \includegraphics[width=\imgwidth]{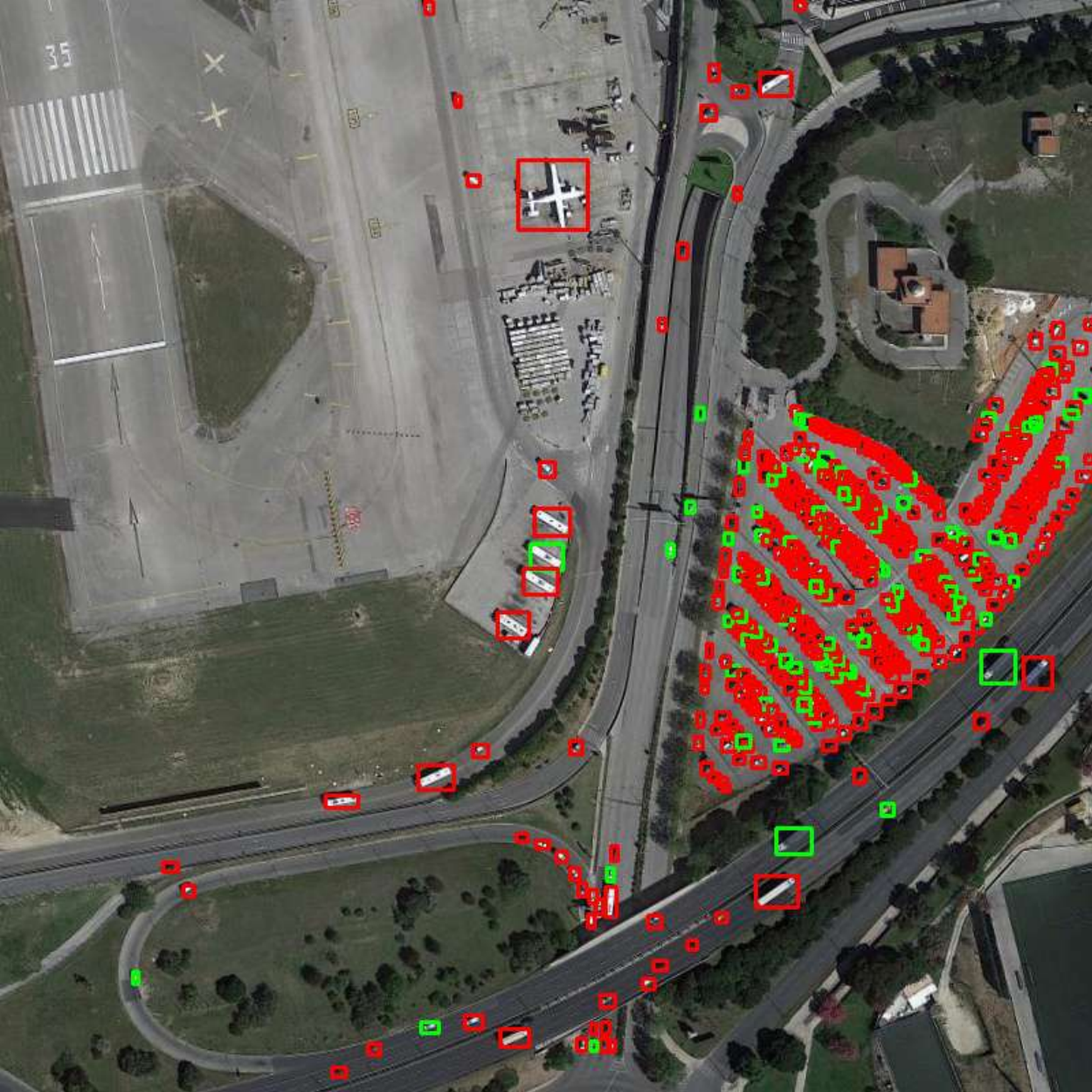} & 
        \includegraphics[width=\imgwidth]{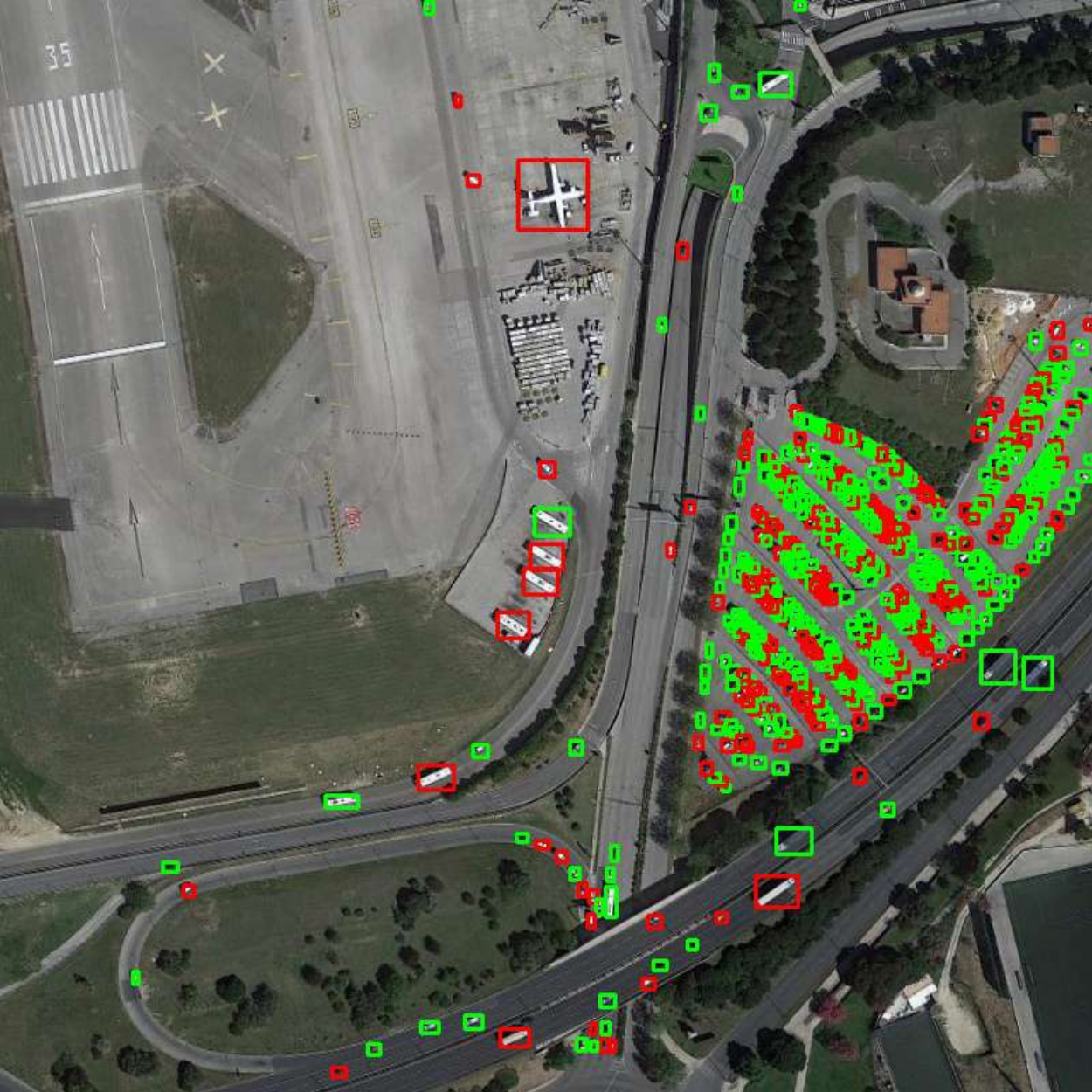} &
        \includegraphics[width=\imgwidth]{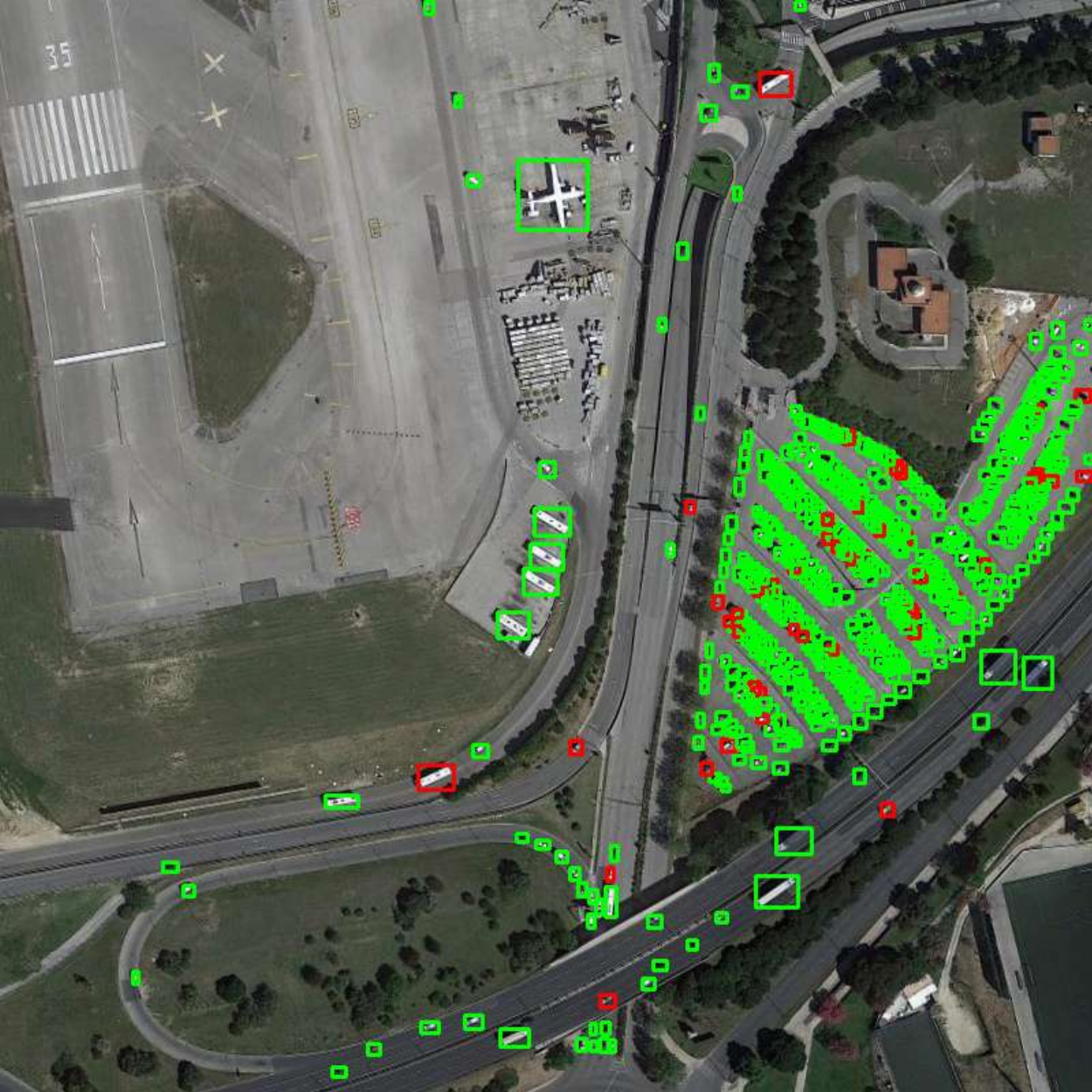} &
        \includegraphics[width=\imgwidth]{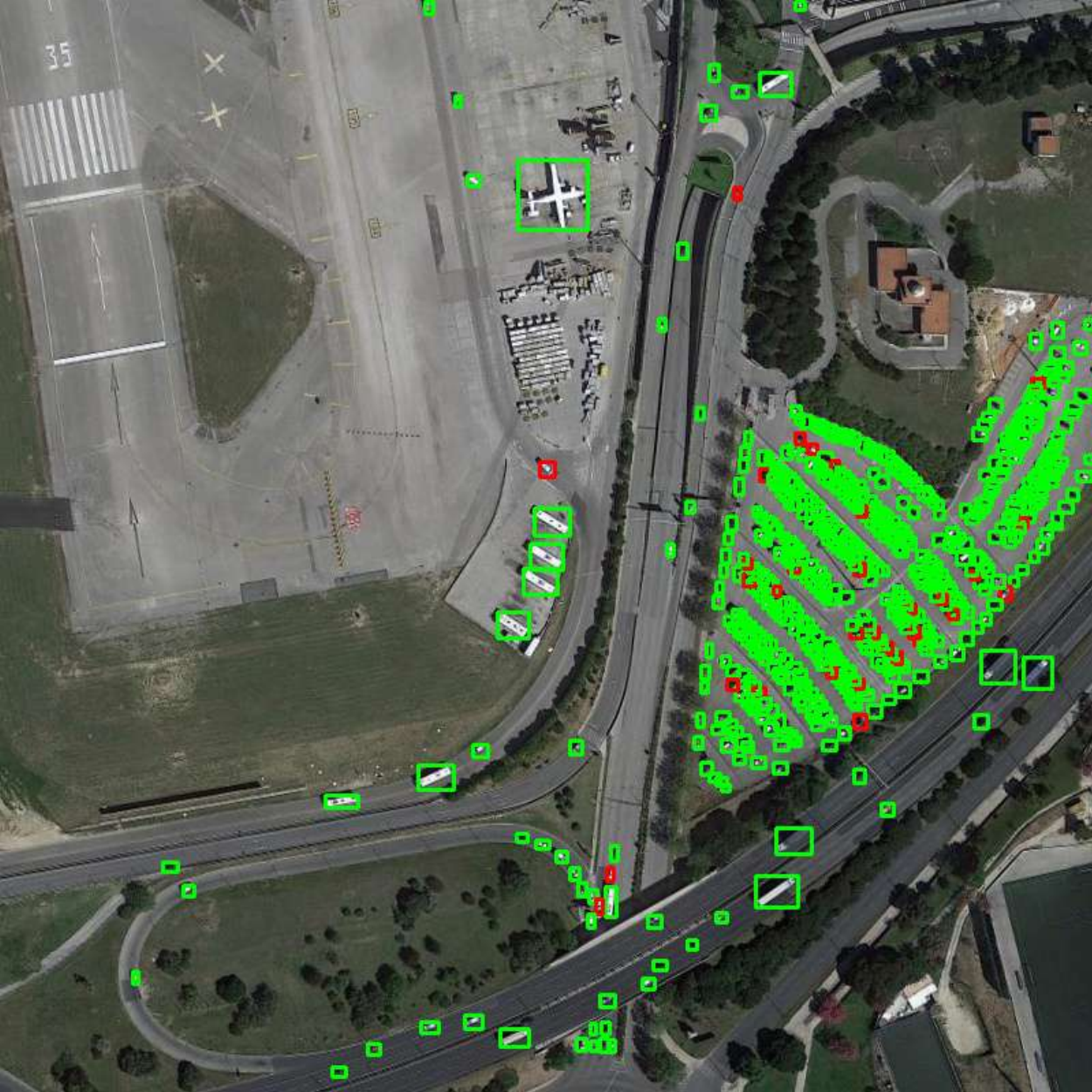} \\[-2pt]
        
        \includegraphics[width=\imgwidth]{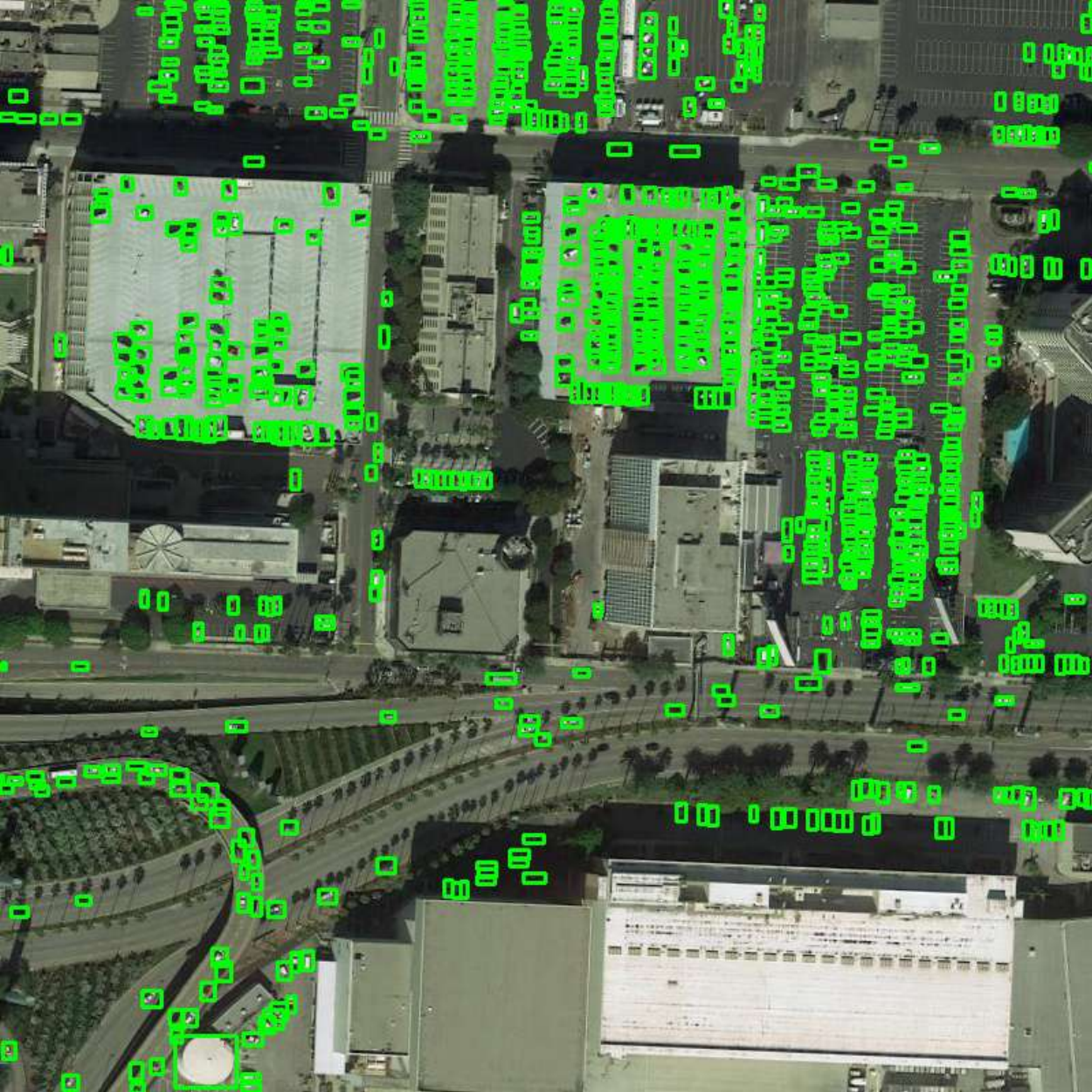} & 
        \includegraphics[width=\imgwidth]{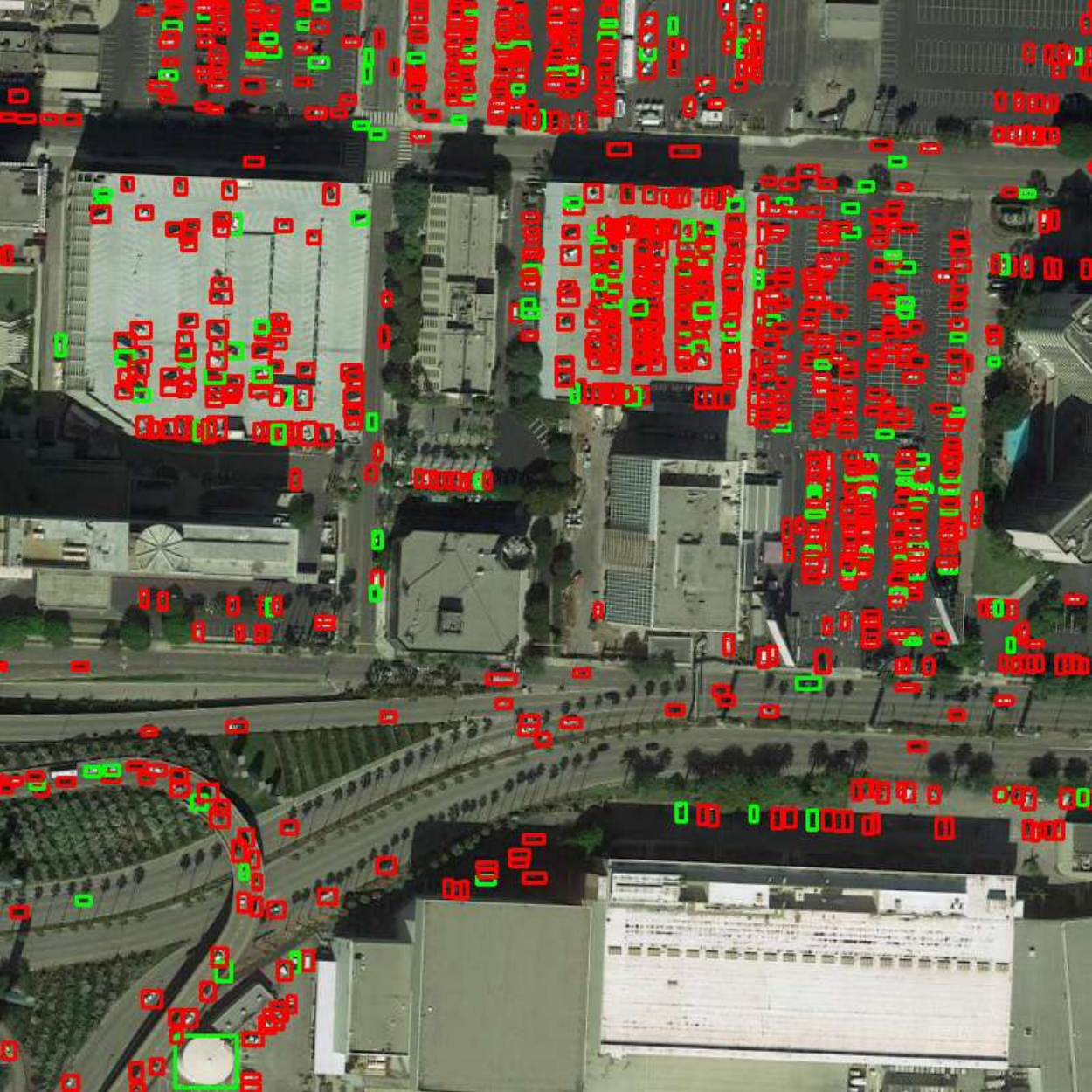} & 
        \includegraphics[width=\imgwidth]{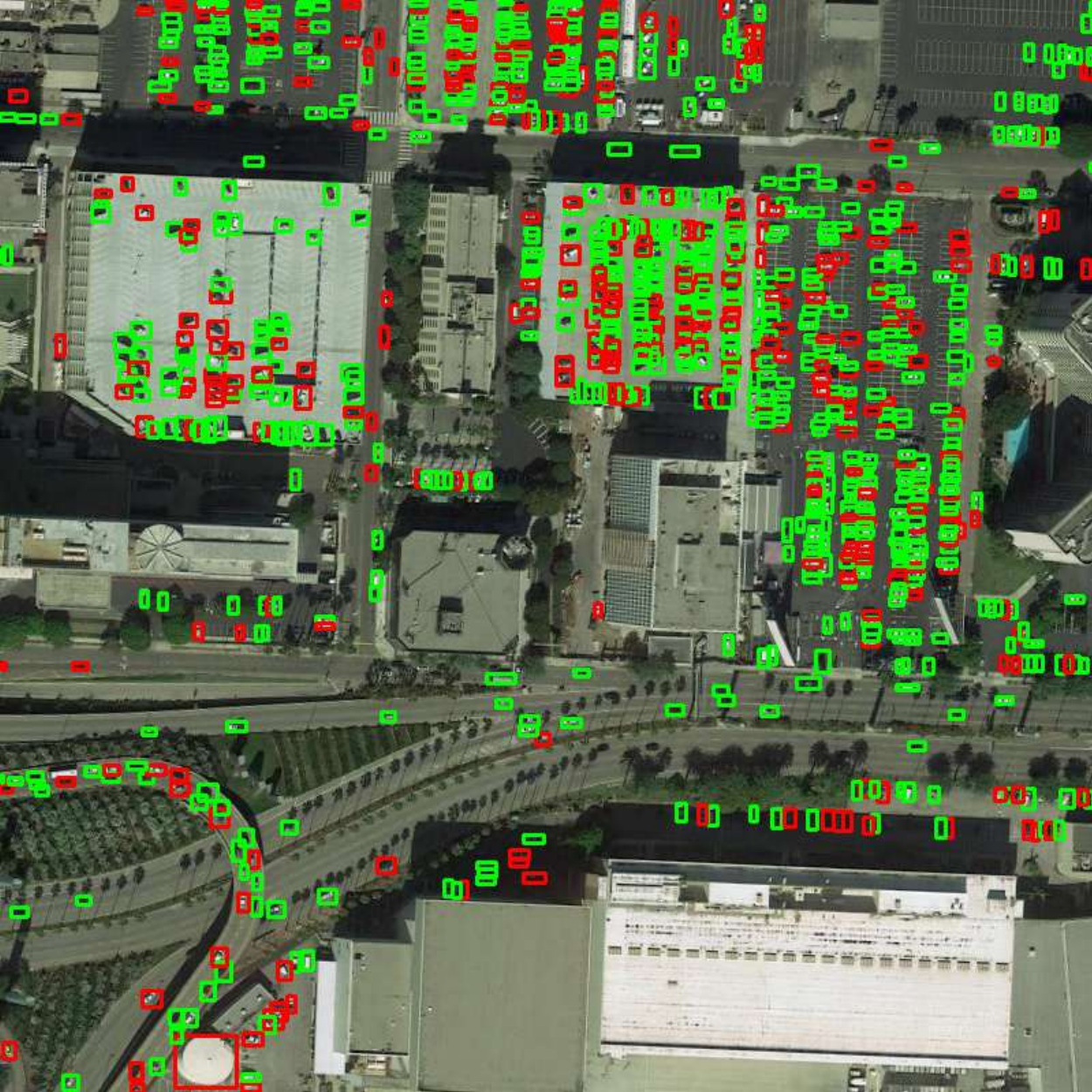} &
        \includegraphics[width=\imgwidth]{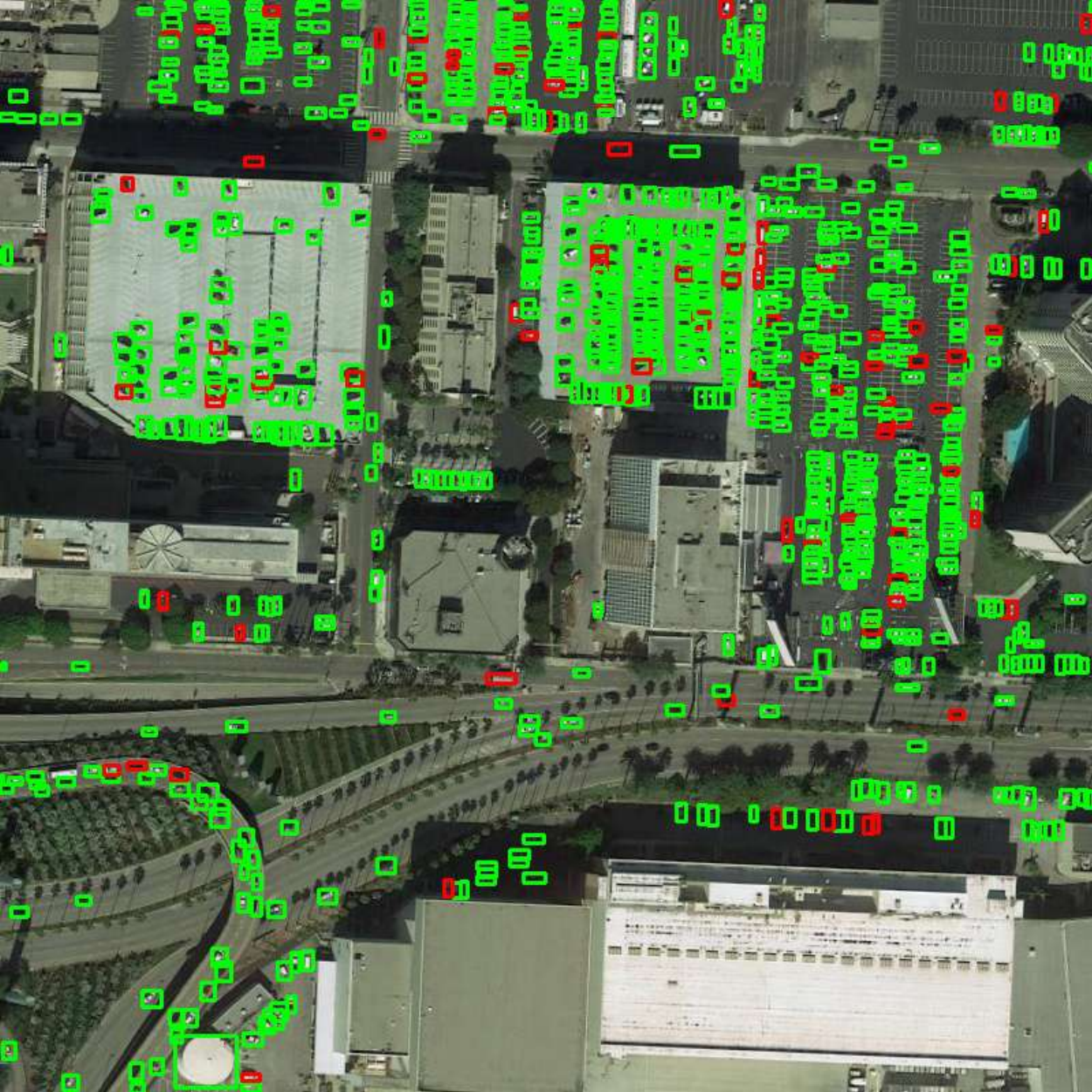} &
        \includegraphics[width=\imgwidth]{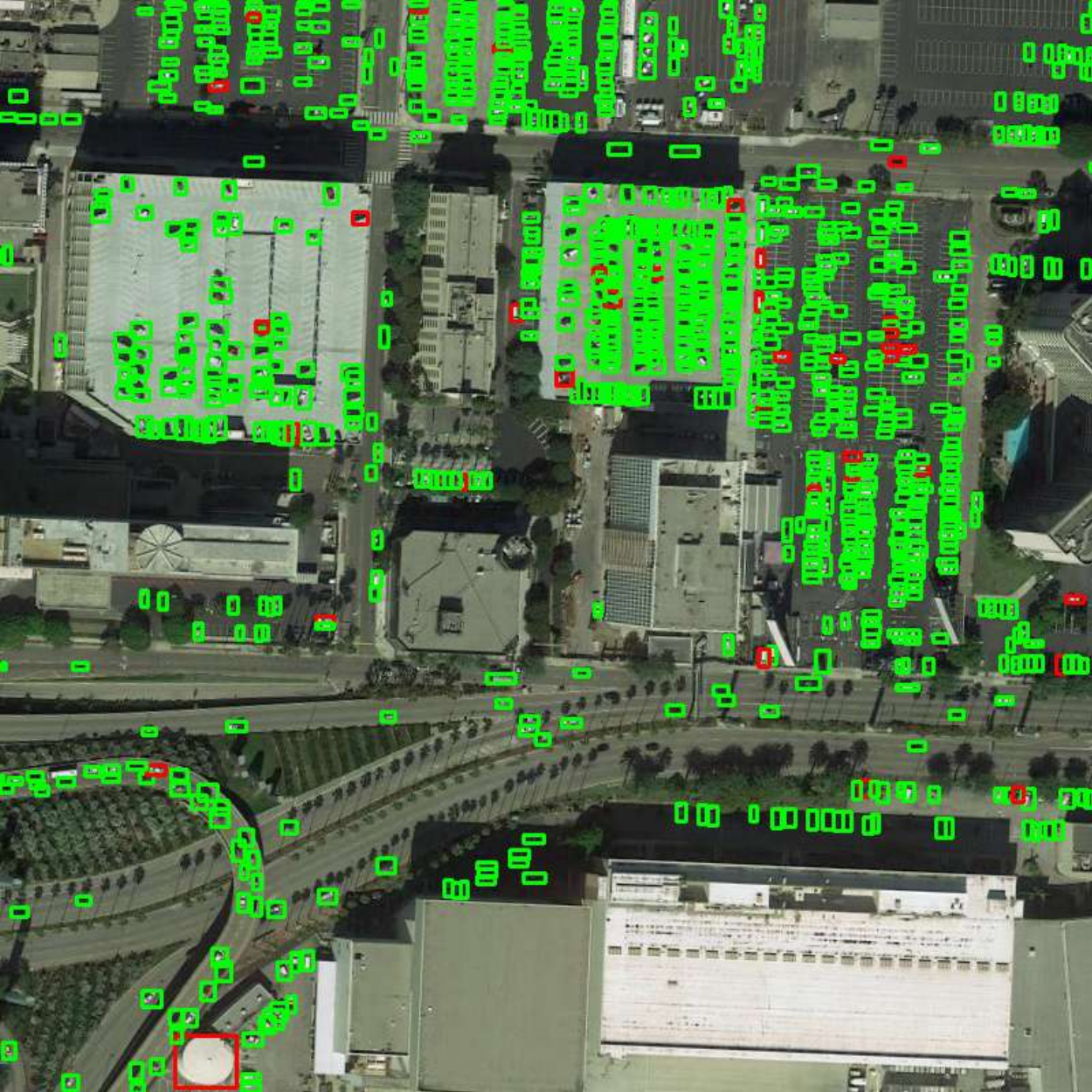} \\[-2pt]
        
        \small Label & \small R-CNN & \small KLDet & \small LTDNet & \small DRMNet(Ours) \\ 
    \end{tabular}
    \setlength{\tabcolsep}{6pt} 
    \vspace{2pt} 
    \caption{\small Visualization of detection results of different object detection models on AI-TOD dataset. The green boxes represent the TP predictions. The red boxes represent the FN predictions.}
    \label{object}
\end{figure*}

  
  

\subsubsection{Region Focusing Method}
We compare our Dense Attention Fusion Module (DAFM) against two representative fusion strategies: a naive Pooling Convolution Fusion (PCF) and the global Multi-Head Self-Attention (MSA). PCF, while simple, often degrades the fine-grained features crucial for small object detection. Conversely, global MSA, despite its powerful modeling capability, suffers from quadratic computational complexity and is prone to introducing background noise by attending to irrelevant pixels.
To address these trade-offs, DAFM introduces a surrogate attention mechanism. It first employs a small set of learnable queries to aggregate key foreground information into compact representations. Subsequently, the computationally intensive self-attention is performed only among these condensed representations. This design effectively filters out background noise and bypasses redundant computations.
As validated in Table \ref{tab:fuzadu}, DAFM achieves a 2.6\% $AP_{50}$ improvement over PCF. Furthermore, it matches the accuracy of global MSA while reducing computational costs by approximately 75\%. These results highlight DAFM's superior balance of accuracy and efficiency for dense object detection tasks.

\subsubsection{Effect of Different Multi-Scale Sampling Sizes }
Within the DFFM module, the EDH component features a multi-path architecture where each parallel path employs a different kernel size to capture features at various scales. We conducted an ablation study to determine the optimal number of paths and their respective kernel sizes, with results summarized in Table \ref{tab:kernel_performance}. We experimented with configurations of two and three paths, such as using kernel sizes of (5, 7) or (3, 6, 9).The results show that the three-path configuration with kernel sizes (3, 6, 9) achieves the best performance. Our analysis provides insights into this outcome. First, the inclusion of a small kernel (size 3) is crucial, as its small receptive field is adept at capturing the fine-grained details of tiny objects. Second, using kernel sizes with a larger interval (e.g., 3, 6, 9) creates more diverse receptive fields, ensuring complementary feature extraction. In contrast, simply increasing the number of paths or using very similar kernel sizes (e.g., (5, 6, 7)) leads to feature redundancy and diminishing returns. Finally, employing excessively large kernels can incorporate excessive background noise, which degrades detection accuracy.

\subsection{Visualization Analysis}
\subsubsection{Model Attention}
We visualized the heatmaps of the DRMNet and three representative baselines using Grad-CAM, as shown in Fig. \ref{gram}. Five images containing densely distributed objects were randomly selected from the dataset for qualitative evaluation. Although YOLOv8 is able to activate certain foreground objects, it fails to consistently attend to scattered tiny objects. RT-DETR \cite{rt-detr} shows limited sensitivity to densely clustered regions and exhibits inferior performance in scenes with dispersed object distributions. YOLOv11 further demonstrates insufficient attention allocation to dense object areas, leading to missed detections. In contrast, DRMNet focuses locally on object-rich areas and reduces the interference of external noise globally, thus providing clearer bounding boxes for objects compared to the former.

\subsubsection{Detection performance of dense objects}

Fig. \ref{object} illustrates qualitative comparisons between DRMNet and three representative detection frameworks, with the IoU threshold uniformly set to 0.3. For visualization consistency, correctly detected objects are marked with green bounding boxes, while missed targets are highlighted in red. The selected examples are taken from the AI-TOD dataset and feature scenes with densely clustered objects. As shown in Fig. \ref{object}, DRMNet exhibits superior detection capability in dense object aggregation areas, significantly reducing missed detections compared with the other methods. These qualitative results further confirm the effectiveness of DRMNet in addressing the challenges of dense tiny object detection in complex remote sensing scenes.

%








\section{CONCLUSION}
In this work, we propose a novel DRMNet for dense tiny object detection, which leverages density maps as explicit spatial priors to guide adaptive feature learning. DRMNet consists of three key components: a Density Generation Branch (DGB) for explicit spatial prior modeling, a Dense Area Focusing Module (DAFM) for adaptive region-aware feature interaction, and a Dual Filter Fusion Module (DFFM) for frequency-aware feature enhancement. The density map generated by DGB serves as strong guidance throughout the network, effectively steering the feature learning process. The DAFM utilizes density maps to identify and focus on dense object regions, enabling efficient local-global feature interaction. The DFFM performs density-guided cross-attention between the dual-stream frequency features to enhance their complementarity while suppressing background interference. Extensive experiments on the AI-TOD and DTOD datasets demonstrate that the proposed DRMNet achieves state-of-the-art performance, with remarkable advantages in tiny object detection tasks for high-resolution remote sensing imagery. In future work, we will further explore the generalization of the proposed framework to more complex scenarios and optimize the model’s efficiency for real-time detection applications.


\bibliographystyle{ieeetr}
\bibliography{references}

\ifCLASSOPTIONcaptionsoff
  \newpage
\fi
\end{document}